\documentclass[10pt,twoside,openright,english,italian]{book}

\usepackage[twoside=true]{geometry}

\usepackage{phdthesis}

\usepackage{fancyhdr}
\usepackage{color}
\usepackage{array}
\usepackage{mdwmath}
\usepackage{mdwtab}
\usepackage{amsmath,amssymb}
\usepackage{cite}
\usepackage{graphicx}
\usepackage{glossaries}
\usepackage{listings}
\usepackage{subfig}
\usepackage{booktabs}
\usepackage{latexsym}
\usepackage{color}
\usepackage{url}
\usepackage{bnf}
\usepackage{rotating}
\usepackage{multirow}
\usepackage{phdtitle}
\usepackage{paralist}
\usepackage{bibentry}
\usepackage[bookmarks=true,pdftex=false,bookmarksopen=true,pdfborder={0,0,0}]{hyperref}
\usepackage{lscape}

\usepackage{algpseudocode}

\usepackage{algorithm}
\usepackage{longtable}
\usepackage[T1]{fontenc} 
\usepackage[latin1]{inputenc}

\newcommand{\hide}[1]{}

\newcommand{\mat}[1]{\mathtt #1}

\newcommand{\ind}[1]{\mathbb 1_{#1}}
\newcommand{\vct}[1]{\mathbf #1}
\newcommand{\T}{\ensuremath{^\top}}
\newcommand{\argmax}{\operatornamewithlimits{\arg\,\max}}

\newcommand{\x}{{\bf x}}
\newcommand{\y}{{\bf y}}

\newcommand{\etal}{\textit{et al. }}

\newcommand{\dquote}[1]{``#1''}

\newtheorem{proposition}{Proposition}
\newtheorem{Definition}{Definition}

\DeclareMathAlphabet{\mathpzc}{OT1}{pzc}{m}{it}

\usepackage[greek,english,italian]{babel}

\usepackage{ulem}
\hypersetup{pdftitle={Title}, pdfauthor={Author}}

\nobibliography*

\department {Department Design and Planning in Complex Environments}
\phdprogram{Dottorato di Ricerca in Nuove tecnologie, informazione
territorio e ambiente, XXX ciclo}

\author{Yonatan Tariku Tesfaye}
\title{Applications of a Graph Theoretic Based Clustering Framework in Computer Vision and Pattern Recognition}

\supervisor{Andrea Prati}
\chair{Fabio Peron}
\titleimage{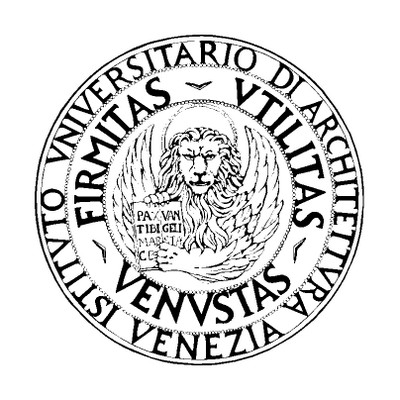}
\phdcycle{}

\begin{document}
\selectlanguage{english}

\maketitle

\pagestyle{empty}

\cleardoublepage
\newpage


\cleardoublepage
\newpage

\pagestyle{fancy}
\setcounter{page}{1}
\pagenumbering{Roman}

\chapter*{Abstract}
\lettrine{R}{ecently} 
, several clustering algorithms have been used to solve variety of problems from different discipline. This dissertation aims to address different challenging tasks in computer vision and pattern recognition by casting the problems as a clustering problem. We proposed novel approaches to solve multi-target tracking, visual geo-localization and outlier detection problems using a unified underlining clustering framework, i.e., dominant set clustering and its extensions, and presented a superior result over several state-of-the-art approaches.

Firstly, this dissertation will present a new framework for multi-target tracking in a single camera. We proposed a novel data association technique using Dominant set clustering (DCS) framework. We formulate the tracking task as finding dominant sets on the constructed undirected edge weighted graph. Unlike most techniques which are limited in temporal locality (i.e. few frames are considered), we utilized a pairwise relationships (in appearance and position) between different detections across the whole temporal span of the video for data association in a global manner. Meanwhile, temporal sliding window technique is utilized to find tracklets and perform further merging on them. Our robust tracklet merging step renders our tracker to long term occlusions with more robustness. DSC leads us to a more accurate approach to multi-object tracking by considering all the pairwise relationships in a batch of frames; however, it has some limitations. Firstly, it finds target trajectories(clusters) one-by-one (peel off strategy), causing change in the scale of the problem. Secondly, the algorithm used to enumerate dominant sets, i.e., replicator dynamics, have quadratic computational complexity, which makes it impractical on larger graphs. 

To address these problems and extend tracking problem to multiple non-overlapping cameras, we proposed a novel and unified three-layer hierarchical approach. Given a video and a set of detections (obtained by any person detector), we first solve {\em within-camera tracking} employing the first two layers of our framework and, then, in the third layer, we solve {\em across-camera tracking} by merging tracks of the same person in all cameras in a simultaneous fashion. To best serve our purpose, a constrained dominant set clustering (CDSC) technique, a parametrized version of standard quadratic optimization, is employed to solve both tracking tasks. The tracking problem is caste as finding constrained dominant sets from a graph. That is, given a constraint set and a graph, CDSC generates cluster (or clique), which forms a compact and coherent set that contains a subset of the constraint set. The approach is based on a parametrized family of quadratic programs that generalizes the standard quadratic optimization problem. In addition to having a unified framework that simultaneously solves within- and across-camera tracking, the third layer helps link broken tracks of the same person occurring during within-camera tracking.  A standard algorithm to extract constrained dominant set from a graph is given by the so-called replicator dynamics whose computational complexity is quadratic per step which makes it handicapped for large-scale applications. In this work, we propose a fast algorithm, based on dynamics from evolutionary game theory, which is efficient and salable to large-scale real-world applications. In test against several tracking datasets, we show that the proposed method outperforms competitive methods.

Another challenging task in computer vision is image geo-localization, here as well, we proposed a novel approach which cast geo-localization as a clustering problem of local image features. Akin to existing approaches to the problem, our framework builds on low-level features which allow local matching between images. We cluster features from reference images using Dominant Set clustering, which affords several advantages over existing approaches. First, it permits variable number of nodes in the cluster, which we use to dynamically select the number of nearest neighbors for each query feature based on its discrimination value. Second, this approach is several orders of magnitude faster than existing approaches. Thus, we use multiple weak solutions through constrained Dominant Set clustering on global image features, where we enforce the constraint that the query image must be included in the cluster. This second level of clustering also bypasses heuristic approaches to voting and selecting the reference image that matches to the query. We evaluated the proposed framework on an existing and new dataset and showed that it outperforms the state-of-the-art approaches by a large margin.

Finally, we present a unified approach for simultaneous clustering and outlier detection in data. We utilize some properties of a family of quadratic optimization problems related to dominant sets. Unlike most (all) of the previous techniques, in our framework the number of clusters arises intuitively and outliers are obliterated automatically. The resulting algorithm discovers both parameters (number of clusters and outliers) from the data. Experiments on real and large scale synthetic dataset demonstrate the effectiveness of our approach and the utility of carrying out both clustering and outlier detection in a concurrent manner.

\chapter *{Acknowledgments}

\begin{center}
	\textit{First and for most, I thank Jehovah God for everything he has done for me.} 
\end{center}	
I would like to express my gratitude to my supervisor Prof. Andrea Prati, for his scientific
guidance and support during my Ph.D. studies.

My sincere thanks also goes to my co-authors Prof. Marcello Pelillo for his insightful comments and  Dr. Mubarak Shah who provided me an opportunity to join his team, CRCV at University of Central Florida, as a research scholar for more than a year, without his precious support, great advice, and fruitful discussion it would not be possible to conduct most of my researches.

I thank my external reviewers: Prof. Lamberto Ballan, and Prof. Christian Micheloni for the time they spent on carefully reading the thesis and for their insightful comments and suggestions.

Also I thank my friends Josh, Leul, Sure, Tedo, Tinsu, Siyum  ... for all the fun time we had and the moments we spent at 'BAUM', Campo, coffee break and so on. In particular, I am grateful to Josh for stimulating discussions, for the sleepless nights we were working together before deadlines, and for all the fun we have had on the trips we had.

Last but not least, I would like to thank my families: my parents (Seni and Tare), my brother (Beck), and my aunty (Tiz) for supporting me spiritually throughout my Ph.D study and my life in general.  This is the fruit of your love, support, guidance and prayers.

\selectlanguage{english}
\chapter*{Preface}
\lettrine{T}his dissertation presents applications of a graph theoretic clustering framework, namely dominant sets and its variants, in solving problems in computer vision and pattern recognition. Chapter 3 presents the first application of dominant sets to solve tracking problem in a single camera. This led to the first publication \cite{YonEyaPelPraIET2016}. Chapter 4 introduces a new multi-target tracking approach in a multiple non-overlapping camera using constrained dominant sets. This work is under review at Transactions on Pattern Analysis and Machine Intelligence (TPAMI) [the arXiv version is in \cite{YonieyaPAMICDSC_tracker}] and has been developed in the last year of my PhD course, during my second visit to the Center for Research in Computer Vision (CRCV) at University of Central Florida, under the supervision of Dr. Mubarak Shah. Chapter 5 presents a Nobel approach to solve another yet very challenging computer vision problem, i.e., large scale image geo-localization using (constrained) dominant set clustering framework. The result has been published in TPAMI \cite{eyasuPAMIgeoloc}. This work is done during my first visit to CRCV at UCF. Finally, the last chapter of the thesis is devoted to outlier detections, we used some properties of dominant sets to simultaneously solve clustering and outlier detections from data. This work has been presented in \cite{zemene2016simultaneous}.

\chapter*{Introduction}

\lettrine{T}he clustering problem has been formulated in many contexts and by researchers in many disciplines; this reflects its broad appeal and usefulness as one of the steps in exploratory data analysis. Clustering has found applications in such diverse disciplines as biology, psychology, archaeology, geology, engineering, information retrieval, and remote sensing. 

Some interesting applications of clustering include clustering of job analytic data \cite{ZimJacJam82}, grouping homogenous industrial companies \cite{CheGnaKet74}, seventh and eighth centuries A.D.  sculptures clustering \cite{SirGovBag85}, clustering of collinear line segments in digital images \cite{SchShnRos82}, clustering is used to study mixture of human populations race \cite{rao1977cluster}, clustering for document classification and indexing \cite{garland1983experiment, salton1975vector} and clustering for investment portfolio performance comparison \cite{cohen1977methodological}.

Cluster analysis is used in numerous applications involving unsupervised learning, where there is no category label assigned to training patterns. As can be seen from the huge amount of literature several problems from different discipline used clustering approach. The use of clustering in computer science and engineering applications has been relatively recent. Cluster analysis plays an important role in solving many problems in natural language processing, pattern recognition and image processing. Recently, its use in the computer vision and pattern recognition community is growing exponentially. Clustering is used in speech and speaker recognition \cite{rabiner1979considerations}, clustering in multi-target tracking is used to solve data association problem \cite{gmmcp, AmiShaECCV12, sutanto1997mean,chen2004dynamic,medeiros2008distributed}, group detection in a video is also casted as a clustering problem in \cite{vascon2014game,hung2011detecting,alletto2014ego}, \cite{solera2017tracking} used clustering to track group, in image geo-localization clustering is used to solve matching between query and reference images \cite{amirshahpami2014}, image retrieval has been effectively casted as a clustering problem \cite{zemene2016constrained}, outlier detection \cite{zemene2016simultaneous}, image registration \cite{stockman1982matching}, and image segmentation is one of the most studied application of clustering in computer vision \cite{zemene2017dominant,ZemPelECCV16, wu1993optimal,chuang2006fuzzy,shi2000normalized,pappas1992adaptive,cai2007fast}.

The availability of a vast collection of clustering algorithms in the literature can easily confound a user attempting to select an algorithm suitable for the problem at hand \cite{JaiMurFlyACM99}. A list of admissibility criteria has been suggested by \cite{FisNesbio71} to make a comparison between clustering algorithms. The criteria used in \cite{FisNesbio71} are mainly based on how the cluster is formed, sensitivity of the clustering technique to changes that do not affect the structure of data, and the way the data was structured. 

As there does not exist a universal clustering approach which can be applied for all kinds of modalities of problems, we need to find the right clustering approach based of the problem at hand. The focus of this dissertation is to propose a unified, novel and robust approaches for problems in multi-target tracking, geo-localization and outlier detection. In all the proposed frameworks, we used the same graph theoretic clustering approach, namely, dominant set clustering (DSC), and its extension, constrained dominant set clustering framework. The intrinsic properties of these approaches make them very suitable for the above-mentioned problems. 

Dominant set clustering framework is first introduced by Pavan and Pelillo \cite{PavPel03,PavPelPAMI07}, and showed its effectiveness in the areas of segmentation, image retrieval and group detection. Dominant set clustering is a graph-theoretic approach for pairwise data clustering which is motivated by the analogies between the intuitive concept of a cluster and that of a dominant set of vertices. It generalizes a maximal clique problem to an edge-weighted graphs. Its correspondence with a linearly constrained quadratic optimization program under the standard simplex, allowed us to use of straightforward and easily implementable continuous optimization techniques from evolutionary game theory. We follow a pill-off strategy to enumerate all possible clusters, that is, at each iteration we remove the cluster from the graph. Even though, it has several advantages over other clustering approaches, the iterative approach we follow to extract clusters cause change in the scale of the problem. Meaning, we do not have a theoretical guaranty that clusters found at the consecutive iterations are the local solutions of the original graph.

Constrained dominant set clustering framework is first proposed in \cite{ZemPelECCV16}. It is a parametrized version of standard quadratic optimization. By properly controlling a regularization parameter which determines the structure and the scale of the underlying problem, we are able to extract groups of dominant-set clusters which are constrained to contain user-selected elements. A standard algorithm to extract constrained dominant sets from a graph is given by the so-called replicator dynamics, whose computational complexity is quadratic per step which makes it handicapped for large-scale applications. In this work, we propose noble fast algorithm, based on dynamics from evolutionary game theory, which is efficient and salable to large-scale real-world applications.

In this thesis, we aim to address problems in multi-target tracking in single and multiple cameras, large scale image geo-localization, and outlier detections by proposing several new algorithms, which utilize the above-mentioned clustering frameworks. 

Firstly, we proposed a novel and efficient single camera multi-target tracking approach, which formulates the tracking task as finding dominant sets in an auxiliary undirected edge-weighted graph. The nodes in the graph represent detection responses from consecutive frames and edge weights depict the similarity between detection responses. In this formulation, the extracted cluster (dominant set) represents a trajectory of a target across consecutive frames. Unlike most techniques, which are limited in temporal locality (i.e. few frames are considered), we utilized a pairwise relationships (in appearance and position) between different detections across the whole temporal span of the video for data association in a \textit{global} manner. Meanwhile, temporal sliding window technique is utilized to find tracklets and perform further merging on them. Our robust tracklet merging step renders our tracker to long term occlusions with more robustness. 

Even though, the approach showed competitive result against several state-of-the-art approaches, it suffers the short comings of the underlined clustering framework, i.e. dominant sets. To solve the limitations of the above approach and extend multi-target tracking across multiple non-overlapping cameras, we proposed a robust and unified three-layered hierarchical approach. We first determine tracks within each camera, by solving data association, and later we associate tracks of the same person in different cameras in a unified approach, hence solving the across-camera tracking. Since appearance and motion cues of a target tend to be consistent in a short temporal window in a single camera tracking, tracklets are generated within short temporal window first and later they are merged to form full tracks (or trajectories). To best serve our purpose, a constrained dominant set clustering (CDSC) technique is employed to solve both tracking tasks. The tracking problem is caste as finding constrained dominant sets from a graph. That is, given a constraint set and a graph, CDSC generates cluster (or clique), which forms a compact and coherent set that contains all or part of the constraint set. \textit{Clusters} represent tracklets and tracks in the first and second layers, respectively. The proposed within-camera tracker can robustly handle long-term occlusions, does not change the scale of original problem as it does not remove nodes from the graph during the extraction of compact clusters and is several orders of magnitude faster (close to real time) than existing methods. Also, the proposed across-camera tracking method using CDSC and later followed by refinement step offers several advantages. More specifically, CDSC not only considers the affinity (relationship) between tracks, observed in different cameras, but also considers the affinity among tracks from the same camera. Consequently, the proposed approach not only accurately associates tracks from different cameras but also makes it possible to link multiple short broken tracks obtained during within-camera tracking, which may belong to a single target track. 

Next, we present a novel approach for a challenging problem of large scale image geo-localization using image matching, in a structured database of city-wide reference images with known GPS coordinates. This is done by finding correspondences between local features of the query and reference images. We first introduce automatic Nearest Neighbors (NN) selection into our framework, by exploiting the discriminative power of each NN feature and employing different number of NN for each query feature. That is, if the distance between query and reference NNs is similar, then we use several NNs since they are ambiguous, and the optimization is afforded with more choices to select the correct match. On the other hand, if a query feature has very few low-distance reference NNs, then we use fewer NNs to save the computation cost. Thus, for some cases we use fewer NNs, while for others we use more requiring on the average approximately the same amount of computation power, but improving the performance, nonetheless. This also bypasses the manual tuning of the number of NNs to be considered, which can vary between datasets and is not straightforward.  Next, we cluster features from reference images using Dominant Set clustering, which possesses several advantages over existing approaches. First, it permits variable number of nodes in the cluster. Second, this approach is several orders of magnitude faster than existing approaches. Finally, we use multiple weak solutions through constrained Dominant Set clustering on global image features, where we enforce the constraint that the query image must be included in the cluster. This second level of clustering also bypasses heuristic approaches to voting and selecting the reference image that matches to the query. 

Finally, we present a modified dominant set clustering approach for simultaneous clustering and outlier detection from data (SCOD). Unlike most approaches our method requires no prior knowledge on both the number of clusters and outliers, which makes our approach more convenient for real applications. A naive approach to apply dominant set clustering is to set a threshold, say cluster size, and label clusters with smaller cluster size than the threshold as outliers. However, in the presence of many cluttered noises (outliers) with a uniformly distributed similarity (with very small internal coherency), the dominant set framework extracts the set as one big cluster. That is, cluster size threshold approaches are handicapped in dealing with such cases. Thus, what is required is a more robust technique that can gracefully handle outlier clusters of different size and cohesiveness. Dominant set framework naturally provides a principled measure of a cluster's cohesiveness as well as a measure of vertex participation to each group (cluster). On the virtue of this nice feature of the framework, we propose a technique which simultaneously discover clusters and outlier in a computationally efficient manner.


In summary, this dissertation makes the following important contributions:
\begin{itemize}
	\item In the proposed Dominant set clustering based single camera multi-target tracker, we formalize the tracking problem as finding DSC from the graph and each cluster represent a trajectory of a single target in different consecutive frames and we extract clusters one-by-one iteratively, which cause change in the scale of the problem. 
	
	\item To improve the limitations of the above approach and extend the approach to a multiple non-overlapping cameras we proposed a new technique which is based on Constrained Dominant Set Clustering, this approach not only address the limitations of our previous approach but also handles tracking in a multiple non-overlapping cameras in a unified manner. 
	
	\item The dissertation also included a novel approach to solve another very challenging task in computer vision, large scale image geo-localization, in this framework we proposed a new approach, which is also based on Dominant Set clustering and its extensions, the approach is based on image matching, we first collect candidate matches based of their local feature matching and later we use Constrained Dominant Set Clustering to decide the best match from the short listed candidates, Unlike most previous approach which use a simple voting scheme.
	
	\item	Finally, the dissertation proposed a novel approach using some properties of dominant sets clustering approach, for simultaneous clustering and outlier detection. Unlike most (all) previous approaches the proposed method does not require any prior knowledge of neither the number of clusters nor outliers.
	
\end{itemize}

The rest of the dissertation is structured as follows: In chapter \ref{DSC_and_CDSC}, we briefly introduce dominant set and constrained dominant set clustering frameworks and present the newly proposed fast approach to extract constrained dominant sets. In chapter \ref{DSC_tracker}, we present new approach for multi-target tracking using Dominant Sets in a single camera scenario. In Chapter \ref{CDSC_tracker}, we present a robust multi-target multi-camera tracking approach using constrained dominant sets. A new approach for large scale image geo-localization is presented in chapter \ref{Geo-localization}. Finally, in chapter \ref{SCOD}, a novel method is proposed for detecting outliers while extracting compact clusters.
\selectlanguage{english}

\tableofcontents

\cleardoublepage
\listoffigures
\listoftables

\cleardoublepage
\newpage

\setcounter{page}{1}
\pagenumbering{arabic}

\cleardoublepage
\chapter{Graph Theoretic Based Clustering Framework}

\label{DSC_and_CDSC}
The problem of clustering consists in organizing a set of objects into clusters (groups, subsets, or categories), in a way that objects in the same cluster should be similar (internal homogeneity), and dissimilar ones organized into different cluster (external Inhomogeneity). 

In particular clustering algorithms are distinguished in two main classes:  feature and similarity (pairwise) -based clustering. In feature-based clustering, objects are represented as points in a metric space with distances reflecting the dissimilarity relations. While in similarity-based clustering, objects are described indirectly by their respective similarity relations. In numerous real-world application domains, it is not possible to find satisfactory features, but it is more natural to provide a measure of similarity. For example, when features consist of both continuous and categorical variables or when the objects to be classified are represented in terms of graphs or structural representations.

A classical approach to pairwise clustering uses concepts and algorithms from graph theory \cite{jain1988algorithms,duda2000pattern}. Certainly, it is natural to map the data to be clustered to the nodes of a weighted graph with edge weights representing similarity relations. These methods are of significant interest since they cast clustering as pure graph-theoretic problems for which a solid theory and powerful algorithms have been developed. 

Graph-theoretic algorithms basically consist of searching for certain combinatorial structures in the similarity graph, such as a minimum spanning tree \cite{zahn1971graph} or a minimum cut \cite{wu1993optimal} and, a complete subgraph (clique) \cite{jain1988algorithms}. Authors in \cite{augston1970analysis,raghavan1981comparison}, argue that the maximal clique is the strictest definition of a cluster. Authors in \cite{PavPel03,PavPelPAMI07}, proposed dominant set clustering framework which generalizes the maximal clique problem to an edge weighted graphs and later authors in \cite{ZemPelECCV16}, proposed an approach which generalizes dominant set framework. 

In this dissertation, we used Dominant set clustering and its extension, Constrained dominant set clustering, approach to solve different problems in computer vision and patter recognition. In the rest of this chapter, we will briefly introduce these clustering frameworks and present newly proposed faster approach to extract constrained dominant sets from the graph, which is linear in complexity.

%

\section{Dominant set clusters and their properties}
\label{subsect:DominantSetClustering}

The theoretical formulation of dominant set clustering has been introduced in \cite{PavPel03,PavPelPAMI07}. It is a combinatorial concept in graph theory that generalizes the notion of the maximal clique to edge-weighted graphs. We can represent the data to be clustered as an undirected edge-weighted graph $G = (V, E,\omega)$ with no self-loops,
where $V = \{1, . . . , n\}$ is the set of vertex which corresponds to data points, $E \subseteq V \times V$ is the edge set representing neighborhood relationships, and $\omega : E \rightarrow \mathbb{R}_+^*$ is the (positive) weight function which quantifies the similarity of the linked objects.  As customary, we represent the graph $G$ with the corresponding weighted adjacency (or similarity) matrix, which is the $n \times n$ nonnegative, symmetric matrix 
\begin{equation}
\mat{A} = (a_{ij})  
\end{equation}
where:
$$
a_{ij} =
\begin{cases}
\omega(i,j), & \text{ if } i  \neq  j \text{ and } (i,j)\in E \\
0, & \text{  if } i=j
\end{cases}
\label{eqn:AffinityMatrix}
$$
Since there are no self-loop in graph $G$, all entries on the main diagonal of $\mat{A}$ are zero.
In an attempt to formally capture this notion, we need some notations and definitions. 

Let $S \subseteq V$ be a non-empty subset of vertices and $ i  \in V$. The (average) weighted degree of $i$ w.r.t. $S$ is defined as:


\begin{equation}
\label{eq:AWDeg}
AWDeg_S(i)=\frac{1}{|S|}\sum\limits_{j\in S} a_{i,j} 
\end{equation}

\noindent Moreover, when $j \notin S$, we can measure similarity between nodes $j$ and $i$, with respect to the average similarity between node $i$ and its neighbors in $S$ as:

\begin{equation}
\phi_S(i,j)=a_{i,j} -  AWDeg_S(i).
\end{equation}

We can compute the weight of $i \in S$ w.r.t. $S$ as:

\begin{flushleft}
	\begin{equation}
	w_S(i)  =
	\begin{cases}
	1 &  \text{if $|S|=1$}\\
	\sum\limits_{j\in S\backslash\{i\}}\phi_{S\backslash\{i\}}(j,i) w_{S\backslash\{i\}}(j) &\text{$otherwise$}
	\end{cases}
	\label{eqn:recursiveDefinition}
	\end{equation}
\end{flushleft}

\noindent Here, we can compute the total weight of the set $S$, $W(S)$, by summing up each weights $w_s(i)$. From this recursive characterization of the weights we can obtain a measure of the overall similarity between a vertex $i$ and $S\setminus\{i\}$ w.r.t the overall
similarity among $S\setminus\{i\}$, and in \cite{PavPel03,PavPelPAMI07} characterizes a set $S$ as dominant set if it satisfies the following two conditions:

\begin{enumerate} 
	\item $w_S(i)>0$, for all $i \in S$.
	\item $w_{S \bigcup {i}}(i)<0$, for all $i \notin S$.
\end{enumerate}

It is evident from the definition that a dominant set satisfies the two basic properties of a cluster: internal coherence and external incoherence. Condition 1 indicates that a dominant set is internally coherent, while condition 2 implies that
this coherence will be destroyed by the addition of any vertex from outside. In other words, a dominant set is a maximally coherent set of data.

\begin{figure}[H]
	\centering
	\includegraphics[width=1\columnwidth,trim=0cm 0.5cm 1cm 0cm, clip]{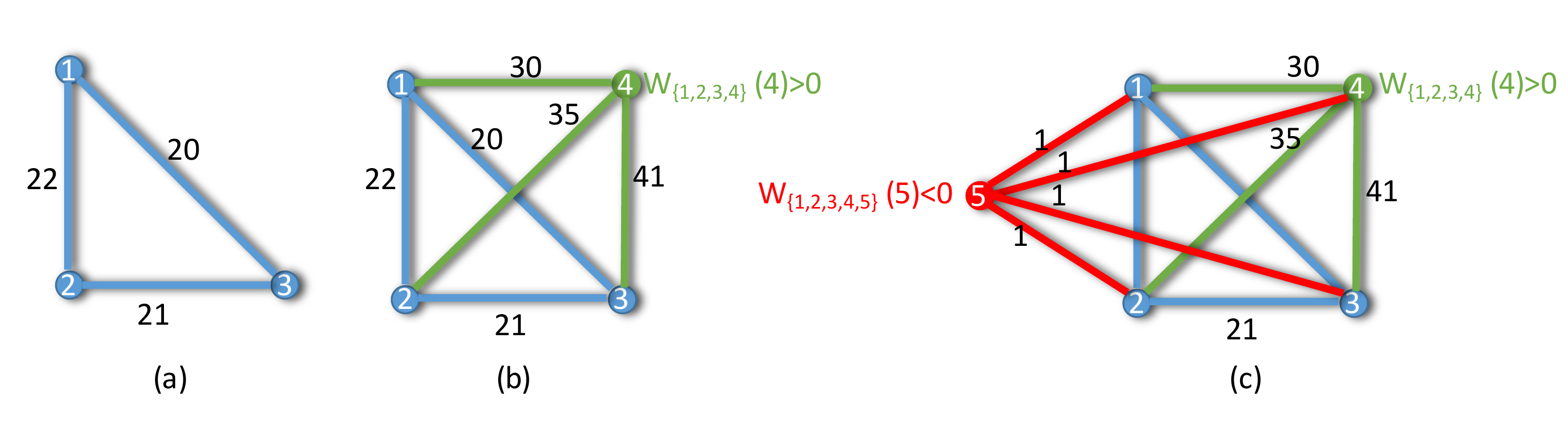}
	\caption{Dominant set example: (a) shows a compact set (dominant set), (b) node 4 is added which is highly similar to the set \{1,2,3\} forming a new compact set. (c) Node 5 is added to the set which has very low similarity with the rest of the nodes and this is reflected in the value $W_{\{1,2,3,4,5\}}(5)$.}
	\label{fig:exemplar}
\end{figure}

\textbf{Example:} Let us consider a graph with nodes $\{1,2,3\}$, which forms a coherent group (dominant set) with edge weights 20, 21 and 22 as shown in Fig. \ref{fig:exemplar}(a). Now, let us try to add a node $\{4\}$ to the graph which is highly similar to the set \{1,2,3\} with edge weights of  30, 35 and 41 (see Fig. \ref{fig:exemplar}(b)). Here, we can see that adding node \{4\} to the set increases the overall similarity of the new set \{1,2,3,4\}, that can be seen from the fact that the weight associated to the node \{4\} with respect to the set \{1,2,3,4\} is positive, $(W_{\{1,2,3,4\} }(4)>0)$. On the contrary, when adding node \{5\} which is less similar to the set \{1,2,3,4\} (edge weight of 1 - Fig. \ref{fig:exemplar}(c)) the overall similarity of the new set \{1,2,3,4,5\}  decreases, since we are adding to the set something less similar with respect to the internal similarity. This is reflected by the fact that the weight associated  to node \{5\} with respect to the set \{1,2,3,4,5\} is less than zero $(W_{\{1,2,3,4,5\}} (5)<0)$.

From the definition of a dominant set the set \{1,2,3,4\} (Fig. \ref{fig:exemplar} (b)) forms a dominant set, as it satisfies both criteria (internal coherence and external incoherence). 
While the weight associated to the node out side of the set (dominant set) is less than zero, $W_{\{1,2,3,4,5\}} (5)<0$. 

The main result presented in \cite{PavPel03,PavPelPAMI07} establishes an intriguing one-to-one correspondence between dominant sets and strict local maximizers of the problem:

\begin{equation}
\label{eq2}
\text{max} \quad \vct{x}^T \mat{A} \vct{x} \quad \text{s.t.} \quad \vct{x}\in \Delta
\end{equation}
where $\Delta=\{ \vct{x} \in \mathbb{R}^n~:~ \sum_i \vct{x}_i = 1, \text{ and } \vct{x}_i \geq 0 \text{ for all } i=1 \ldots n\}$ is the standard simplex of $\mathbb{R}^n$. 
Moreover, in \cite{PavPel03,PavPelPAMI07} it is proven that if $S$ is a dominant set, then its weighted characteristic vector $X^S$, defined below, is a strict local solution of the problem \eqref{eq2}:
\begin{displaymath}
\vct{x}_i^S=
\begin{cases} \frac{w_S(i)}{W(S)},&\text{if\quad $i \in S$},\\ 0,&\text{otherwise}
\end{cases}
\end{displaymath}
\noindent Conversely, under mild conditions, it turns out that if $\vct{x}$ is a strict local solution of problem \eqref{eq2}, then its "support"  $S = \{i \in V~:~x_i > 0\}$ is a dominant set.
By virtue of this result, we can find a dominant set by first localizing a solution of problem \eqref{eq2} with an appropriate continuous optimization technique, and then picking up the support set of the found solution. In this sense, we indirectly perform combinatorial optimization via continuous optimization.



The standard approach for finding the local maxima of problem \eqref{eq2}, as used in \cite{PavPelPAMI07}, is to use replicator dynamics a well-known family of algorithms from evolutionary game theory inspired by Darwinian selection processes. 
Let $\mat{A}$ be a non-negative real-valued $n \times n$ matrix, the discrete version of the dynamics is defined as follows:

\begin{equation}
\vct{x}_i^{(t+1)}=\vct{x}_i^{(t)} \frac{(\mat{A}\vct{x}^{(t)})_i}{\vct{x}^{(t)'}\mat{A}\vct{x}^{(t)}}
\label{replicatore}
\end{equation}
for $i=1,\ldots,n$.

Variety of StQP applications \cite{BomGO97,KarJanNeuCraGabJMRI2007,PavPelPAMI07,PelEO2009,SinNarIJCV2008} have proven the effectiveness of replicator dynamics. However, its computational complexity, which is $O(n^2)$ for problems involving $n$ variables, prevents it from being used in large-scale applications. Its exponential variant, though it reduces the number of iterations needed for the algorithm to find a solution, suffers from a per-step quadratic complexity.

In this work, we adopt a new class of evolutionary game dynamics called infection-immunization dynamics (InImDyn), which have been shown to have a linear time/space complexity for solving standard quadratic programs. In \cite{BulPelBomCVIU11}, it has been shown that InImDyn is orders of magnitude faster but as accurate as the replicator dynamics.

\begin{algorithm}
	\caption{Find Equilibrium($\mat{B}$,$\vct x$,$\tau$)}\label{alg:inimdyn}
	\textbf{Input}:  $n\times n$ payoff matrix $\mat{B}$, initial distribution $\vct x\in\Delta$ and tolerance $\tau$.\\
	\textbf{Output}: Fixed point $\vct x$
	\begin{algorithmic}[1]
		\While {$\epsilon(\vct x)>\tau$} \\
		$\vct y\gets \mathcal S(\vct x)$ \\
		$\delta\gets 1$
		\If{${(\vct y-\vct x)\T\mat{B}(\vct y-\vct x)}<0$}
		\State $\delta\gets\min\left\{\frac{(\vct x - \vct y)\T\mat{B}\vct x}{{(\vct y - \vct x)\T\mat{B}(\vct y-\vct x)}},1\right\}$
		\EndIf
		\State $\vct x\gets \delta(\vct y-\vct x)+\vct x$
		\EndWhile\\
		\Return $\vct x$
	\end{algorithmic}
	\vspace{-.03in}
\end{algorithm}

The dynamics, inspired by infection and immunization processes summarized in Algorithm (\ref{alg:inimdyn}), finds the optimal solution by iteratively refining an initial distribution $\vct{x} \in \Delta$. The process allows for invasion of an infective distribution $\vct{y}\in \Delta$ that satisfies the inequality $ (\vct y - \vct x)\T \mat{A}\vct x > 0$, and combines linearly $\vct{x}$ and $\vct{y}$ (line 7 of Algorithm (\ref{alg:inimdyn})), thereby engendering a new population $\vct{z}$ which is immune to $\vct{y}$ and guarantees a maximum increase in the expected payoff.

A selective function, $\mathcal{S}(\vct{x})$, returns an infective strategy for distribution $\vct{x}$ if it exists, or $\vct{x}$ otherwise (line 2 of Algorithm (\ref{alg:inimdyn})).
Selecting a strategy $\vct{y}$ which is infective for the current population $\vct{x}$, the extent of the infection, $\delta_{\vct y}(\vct x)$, is then computed in lines 3 to 6 of Algorithm (\ref{alg:inimdyn}).

By reiterating this process of infection and immunization the dynamics drives the population to a state that cannot be infected by any other strategy. If this is the case then $\vct{x}$ is an equilibrium or fixed point under the dynamics. The refinement loop of Algorithm (\ref{alg:inimdyn}) controls the number of iterations allowing them to continue until $\vct{x}$ is with in the range of the tolerance $\tau$ and we emperically set $\tau$ to $10^{-7}$. The range $\epsilon(\vct{x})$ is computed as  $\epsilon(\vct{x})$ = $\sum\limits_{i\in J}\textrm{min}\left\{x_i,( \mat{A}\vct{x})_i-\vct{x}\T \mat{A}\vct{x}\right\}^2$.

\section{Constrained Dominant Set clustering. } \label{CDSC}
As introduced in \cite{ZemPelECCV16}, constrained dominant set clustering, a constrained quadratic optimization program, is an efficient and accurate approach, which has been applied for interactive image segmentation, image geo-localization \cite{eyasuPAMIgeoloc} and multi-target tracking problems \cite{YonieyaPAMICDSC_tracker,YonEyaPelPraIET2016,tesfaye2014multi}. The approach generalizes dominant set framework \cite{PavPelPAMI07}, which is a well known generalization of the maximal clique problem to edge weighted graphs. 

Given an edge weighted graph $G(V,E,\omega)$ and a constraint set $\mathcal{Q} \subseteq V$, where $V, E$ and $\omega$, respectively, denote the set of nodes (of cardinality $n$), edges and edge weights. The objective is to find the sub-graph that contains all or some of elements of the constraint set, which forms a coherent and compact set. 

Consider a graph, $G$, with $n$ vertices (set $ V $), and its weighted adjacency matrix $\mat{A}$. Given a parameter $ \alpha > 0$, let us define the following parametrized quadratic program:
\begin{equation}
\label{eqn:parQP}
\begin{array}{ll}
\text{maximize }  &  f_{\mathcal{Q}}^\alpha(\x) = \x\T ({\mat{A}} - \alpha I_{\mathcal{Q}}) \x, \\
\text{subject to} &  \mathbf{x} \in \Delta,
\end{array}
\end{equation}
\noindent where $\Delta=\{ \x \in \mathbb{R}^n: \sum_i x_i = 1, \text{ and } x_i \geq 0 \text{ for all } i=1 \ldots n\}$, $\mathbf{x}$ contains a membership score for each node and  $I_{\mathcal{Q}}$ is the $n \times n$ diagonal matrix whose diagonal elements are set to 1 in correspondence to the vertices contained in $ V \setminus \mathcal{Q}$ (a set $V$ without the element $\mathcal{Q}$) and to zero otherwise.

Let $\mathcal{Q} \subseteq  V$, with $\mathcal{Q} \neq \emptyset$ and let $\alpha > \lambda_{\max}( {\mat{A}}_{ V \setminus \mathcal{Q}}) $, where $\lambda_{\max}( {\mat{A}}_{ V \setminus \mathcal{Q}})$ is the largest eigenvalue of the principal submatrix of ${\mat{A}}$ indexed by the elements of $ V \setminus \mathcal{Q}$.
If $\x$ is a local maximizer of $f_{\mathcal{Q}}^\alpha$ in $\Delta$, then
$\sigma(\x) \cap \mathcal{Q} \neq \emptyset$, where,
$
\sigma(\x) = \{i \in V~:~x_i > 0\}
$ .

The above result provides us with a simple technique to determine dominant set clusters containing user-specified query vertices, $\mathcal{Q}$. Indeed, if $\mathcal{Q}$ is a vertex selected by the user, by setting
\begin{equation}
\label{alphabound}
\alpha > \lambda_{\max}({\mat{A}}_{ V \setminus \mathcal{Q}}),
\end{equation}
we are guaranteed that all local solutions of (\ref{eqn:parQP}) will have a support
that necessarily contains elements of $\mathcal{Q}$.

\section{Fast Approach for Solving Constrained Dominant Set Clustering } \label{speedin up data association }

Constrained quadratic optimization program (\ref{eqn:parQP}), can be solved using dynamics from evolutionary game theory. The well-known standard game dynamics to equilibrium selection, replicator dynamics, though efficient, poses serious efficiency problems, since the time complexity for each iteration of the replicator dynamics is $\mathcal{O}(n^2)$, which makes it not efficient for large scale data sets \cite{ZemPelECCV16}. Rota Bul{\`{o}} \etal \cite{BulPelBomCVIU11} proposed a new class of evolutionary game dynamics, called Infection and Immunization Dynamics (InfImDyn).  InfImDyn solves the problem in linear time. However, it needs the whole affinity matrix  to extract a dominant set which, more often than not, exists in local range of the whole graph.
Dominant Set Clustering (DSC) \cite{PavPelPAMI07} is an iterative method which, at each iteration, peels off a cluster by performing a replicator dynamics until its convergence. Efficient out-of-sample \cite{PavPelNIPS04}, extension of dominant sets, is the other approach which is used to reduce the computational cost by sampling the nodes of the graph using some given sampling rate that affects the framework efficacy. Liu \etal \cite{LiuLatYanPAMI13} proposed an iterative clustering algorithm, which operates in two steps: Shrink and Expansion. These steps help reduce the runtime of replicator dynamics on the whole data, which might be slow. The approach has many limitations such as its preference of  sparse graph with many small clusters and the results are sensitive to some additional parameters. 
Another approach which tries to reduce the computational complexity of the standard quadratic program (StQP \cite{Bom02}) is proposed by \cite{ChuWanLiuHuaPeiPVLDB15}. 

All the above formulations, with their limitations, try to minimize the computational complexity of StQP using the standard game dynamics, whose complexity is $\mathcal{O}(n^2)$ for each  iteration. 

In this work we propose a fast approach (listed in Algorithm \ref{alg:FastInimdyn}), based on InfImDyn approach which solves StQP in $\mathcal{O}(n)$, for the recently proposed formulation, $\x\T ( {\mat{A}} - \alpha  I_{\mathcal{Q}}) \x, $ which of-course generalizes the StQP.

InfImDyn is a game dynamics inspired by Evolutionary game theory. The dynamics extracts a dominant set using a two-steps approach (infection and immunization), that iteratively increases the compactness measure of the objective function by driving the (probability) distribution with lower payoff to extinction, by determining an ineffective distribution $\vct{y}\in \Delta$, that satisfies the inequality $ (\vct y - \vct x)\T\mat{A}\vct x > 0$, the dynamics combines linearly
the two distributions ($\vct{x}$ and $\vct{y}$), thereby engendering a new population $\vct{z}$ which is immune to $\vct{y}$ and guarantees a maximum increase in the expected payoff. In our setting, given a set of instances (tracks, tracklets) and their affinity, we first assign all of them an equal probability (a distribution at the centre of the simplex, a.k.a. barycenter). The dynamics then drives the initial distribution with lower affinity to extinction; those which have higher affinity start getting higher, while the other get lower values. A selective function, $\mathcal{S}(\vct{x})$, is then run to check if there is any infective distribution; a distribution which contains instances with a better association  score. By iterating this process of infection and immunization the dynamics is said to  reach the equilibrium, when the population is driven to a state that cannot be infected by any other distribution, that is there is no distribution, whose support contains a set of instances with a better association score. The selective function, however, needs whole affinity matrix, which makes the InfImDyn inefficient for large graphs. We  propose an algorithm, that reduces the search space using the Karush-Kuhn-Tucker (KKT) condition of the constrained quadratic optimization, effectively enforcing the user constraints. In the constrained optimization framework \cite{ZemPelECCV16}, the algorithm computes the eigenvalue of the submatrix for every extraction of the compact sets, which contains the user constraint set. Computing eigenvalues for large graphs is computationally intensive, which makes the whole algorithm inefficient. 

In our approach, instead of running the dynamics over the whole graph, we localize it on the sub-matrix, selected using the dominant distribution, that is much smaller than the original one. To alleviate the issue with the eigenvalues, we utilize the properties of eigenvalues; a good approximation for the parameter $\alpha$ is to use the maximum degree of the graph, which of-course is larger than the eigenvalue of corresponding matrix. The computational complexity, apart from eigenvalue computation, is reduced to $\mathcal{O}(r)$ where $r$, which is much smaller than the original affinity,  is the size of the sub-matrix where the dynamics is run.

Let us summarize the KKT conditions for quadratic program reported in eq. (\ref{eqn:parQP}). By adding Lagrangian multipliers, $n$ non-negative constants $ \mu_1, ...., \mu_n$ and a real number $\lambda$, its Lagrangian function is defined as follows: 

\begin{displaymath}
\mathcal{L}(x,\mu,\lambda) = f_{\mathcal{Q}}^\alpha(\x) + 
\lambda \left(1 - \sum\limits_{i+1}^{n}x_i \right) +
\sum\limits_{i+1}^{n}\mu_ix_i.
\end{displaymath}

For a distribution $x \in \Delta$ to be a KKT-point, in order to satisfy the first-order necessary conditions for local optimality \cite{Lue84}, it should satisfy the following two conditions:

\begin{displaymath}
2*[(\mat{A} - \alpha I_{\mathcal{Q}}) \x]_i - \lambda + \mu_i = 0,
\end{displaymath}
for all $i=1 \ldots n$, and
\begin{displaymath}
\sum_{i=1}^n x_i \mu_i = 0~.
\end{displaymath}
Since both the $x_i$ and the $\mu_i$ values are nonnegative, the
latter condition is equivalent to saying that $i \in \sigma(\x)$ which implies that 
$\mu_i= 0$, from which we obtain:
\begin{equation}
\label{KKT}
[(\mat{A} - \alpha I_{\mathcal{Q}}) \x]_i ~
\begin{cases} 
~ = ~   \lambda/2, ~ \mbox{ if } i \in \sigma(\x) \\ 
~ \le ~ \lambda/2, ~ \mbox{ if } i \notin \sigma(\x)  
\end{cases}
\end{equation}

We then need to define a \textit{Dominant distribution}
\begin{Definition}
	\label{def:dominantDistribution}
	A distribution $\y \in \Delta$ is said to be a \textbf{dominant distribution} for $\x \in \Delta$ if 
	
	\begin{equation}
	\left\{\sum\limits_{i,j=1}^{n}   x_iy_ja_{ij} - \alpha x_iy_j \right\} > \left\{\sum\limits_{i,j=1}^{n}   x_ix_ja_{ij} - \alpha x_ix_j\right\}
	\end{equation}
\end{Definition}


Let the "support" be
$
\sigma(\x) = \{i \in V~:~x_i > 0\}
$ 
and  $e_i$ the $i^{th}$ unit vector (a zero vector whose $i^{th}$ element is one).	

\begin{proposition}
	Given an affinity $\mat{A} $ and a distribution $\x \in \Delta$, if $(\mat{A}\x)_i             > \x'\mat{A}\x - \alpha \x'_{\mathcal{Q}} \x_{\mathcal{Q}}, \mbox{for} \;i \notin \sigma(\x)$,	
	\begin{enumerate}
		\item $\x$ is not the maximizer of the parametrized quadratic program of (\ref{eqn:parQP})
		\item $e_i$ is a \textbf{dominant distribution} for $\x$
	\end{enumerate}	
\end{proposition}
%
\proof To show the first condition holds: Let's assume $\x$ is a KKT point

\[
\x\T ({\mat{A}} - \alpha I_{\mathcal{Q}}) \x = \sum\limits_{i=1}^{n} x_i [(\mat{A} - \alpha I_{\mathcal{Q}}) \x]_i\]
Since x is a KKT point
\[ \x\T ({\mat{A}} - \alpha I_{\mathcal{Q}}) \x = \sum\limits_{i=1}^{n} x_i * \lambda/2 = \lambda/2\]
From the second condition, we have: 
\[ [(\mat{A} - \alpha I_{\mathcal{Q}}) \x]_i \le \lambda/2 = \x\T ({\mat{A}} - \alpha I_{\mathcal{Q}}) \x \]
Since $i \notin \sigma(\x)$
\[ (\mat{A}\x)_i \le  \x\T ({\mat{A}} - \alpha I_{\mathcal{Q}}) \x \]
Which concludes the proof showing that the inequality does not hold.

For the second condition, if $e_i$ is a \textbf{dominant distribution} for $\x$, it should satisfy the inequality

\[ \left\{ e_i\T ({\mat{A}} - \alpha I_{\mathcal{Q}}) \x \right\} > \left\{\x\T ({\mat{A}} - \alpha I_{\mathcal{Q}}) \x \right\}\]
Since $i \notin \sigma(\x)$

\[ (\mat{A}x)_i > \left\{\x\T ({\mat{A}} - \alpha I_{\mathcal{Q}}) \x \right\}\]
Which concludes the proof

\endproof

The proposition provides us with an easy-to-compute dominant distribution.

Let a function, $\mathcal{S}(\mat{A},x)$, returns a dominant distribution for distribution, $x$, $\emptyset$ otherwise and $\mathcal{G}(\mat{A},\mathcal{Q},x)$ returns the local maximizer of program \eqref{eqn:parQP}. We summarize the details of our proposed algorithm in Algorithm (\ref{alg:FastInimdyn}).

\begin{algorithm}[H]
	\caption{Fast CDSC}\label{alg:FastInimdyn} 
	\textbf{Input}: Affinity $\mat{B}$, Constraint set $\mathcal{Q}$\\
	Initialize $\x$ to the barycenter of $\Delta_{\mathcal{Q}}$\\
	$\x_d$ $\leftarrow$ $\x$, initialize \textit{dominant distribution} \\
	\textbf{Output}: Fixed point $\vct x$
	\begin{algorithmic}[1]
		\While {true} \\
		$\x_d$ $\leftarrow$ $\mathcal{S}(\mat{B},\x)$, Find dominant distribution for $x$
		\If{$\x_d = \emptyset$} break
		\EndIf\\
		$\mathcal{H}$ $\leftarrow$ $\sigma(\x_d) \cup \mathcal{Q}$, subgraph nodes\\
		$\mat{A}\leftarrow \mat{B}_{\mathcal{H}}$\\
		$\x_l$ $\leftarrow$ $\mathcal{G}(\mat{A},\mathcal{Q},x)$\\
		$\x$ $\leftarrow$ $\x$*0\\
		$\x (\mathcal{H})$ $\leftarrow$ $\x_l$	
		\EndWhile \\
		\Return $\vct x$
	\end{algorithmic}
	\vspace{-.03in}
\end{algorithm}

The selected dominant distribution always increases the value of the objective function. Moreover, the objective function is bounded  which guaranties the convergence of the algorithm.

\section{Summary}

In the first part of this chapter, we briefly introduced dominant set clustering framework, a pairwise clustering approach, based on the notion of a dominant set, which can be seen as an edge weighted generalization of a clique. In the second part of the chapter, we have discussed an approach, constrained dominant-set, that finds a collection of dominant set clusters constrained to contain user-defined elements. The approach is based on some properties of a family of quadratic optimization problems related to dominant sets which show that, by properly selecting a regularization parameter that controls the structure of the underlying function, we are able to "force" all solutions to contain user specified elements. Finally, We have presented a novel fast approach to extract constrained dominant sets from the graph.


\chapter{Multi-Object Tracking Using Dominant Sets}
\label{DSC_tracker}
Multi-object tracking is an interesting but challenging task in the field of computer vision. Most previous works based on data association techniques merely take into account the relationship between detection responses in a locally limited temporal domain, which makes them inherently prone to identity switches and difficulties in handling long-term occlusions. In this chapter, a dominant set clustering based tracker is presented,  which formulates the tracking task as finding dominant sets  on the constructed undirected edge weighted graph. Meanwhile, temporal sliding window technique is utilized to find tracklets and perform further merging on them. Our robust tracklet merging step renders our tracker to long term occlusions with more robustness. We present results on three different challenging datasets (i.e. PETS2009-S2L1 \cite{pets}, TUD-standemitte \cite{anton2011} and ETH dataset (\dquote{sunny day} sequence) \cite{Sunnyday_dataset}), and show significant improvements compared with several state-of-art methods.

\section{Introduction}
In order to ensure security of people and places by means of intelligent video surveillance, the current trend is to increase the number of cameras installed, both as a deterrent and to better monitoring the surveyed area. As a consequence, we have witnessed an exponential increase of the data to be watched and stored, requiring inevitably the use of automatic processing of videos for scene analysis and understanding, both for real time and a-posterior mining. In this way, a large variety of video data provided by installed cameras is automatically analyzed for event detection, object and people tracking as well as behaviour analysis. These actions offer a valid support to investigations and crime detection.

Tracking targets is a challenging task: variations in the type of camera, lighting conditions, scene settings (e.g. crowd or occlusions), noise in images, variable appearance of moving targets and the point of view of the camera must be accounted. Following multiple targets while robustly maintaining data association remains a largely open problem.
In \cite{survey} a large experimental survey of various tracking approaches is presented, evaluating suitability of each approach in different situations and with different constraints (e.g assumptions on the background, motion model, occlusions, etc.).

In recent years, due to significant improvements in object detection, several researchers have proposed tracking methods that associate detection responses into tracks, also referred as Association Based Tracking (ABT) techniques \cite{anton2011,Perera,Pirsiavash, xing, nevatia,AmiShaECCV12} . An offline trained people detector is used to generate detection responses,  then tracklets are produced by linking these responses and further associated into longer tracks. Similarity between tracklets (i.e., the linking probabilities) is based on the motion smoothness and appearance similarity. A Hungarian algorithm is often used to find the global optimum \cite{Perera,xing}. As compared to generative models,  ABT is powerful in assessing the presence of objects on the scene since it uses discriminatively trained detectors, and needs no manual initialization. Association-based approaches are prone to handling long-term occlusions between targets and the complexity is polynomial with the number of targets present. In order to differentiate between different targets, speed and distance between tracklet pairs are often used as motion descriptors, whereas appearance descriptors are often based on global or part-based color histograms. Nevertheless, how to deal with motion in moving cameras  and how to better distinguish nearby targets remain key issues that limit the performance of ABT.

In most of the ABT works \cite{Perera,Nevatia2}, the affinity score between detection responses is computed once and kept fixed for all the later processes. Conversely, in this proposal we develop a more flexible approach where associations are made at two levels and the affinity measure is iteratively refined based on the knowledge retrieved from the previous level. Moreover, most of the previous methods \cite{Sidney,Pascal,Todorovic,shah2} use locally-limited temporal information by focusing only on the pairwise relationship of the tracklets, rather than applying data association among multiple tracklets over the whole video in a global manner. As a consequence, existing approaches are prone to identity switches (i.e., assigning different labels/IDs to the same target) in cases where targets with small spatial and appearance differences move together (which are rather common cases in real security videos).

In this chapter, a dominant set clustering based tracker is proposed, which formulates the tracking task as the problem of finding dominant set clusters in an auxiliary undirected edge-weighted graph. Unlike the previous approaches, the proposed method for data association combines both appearance and position information in a \textit{global} manner. Object appearance is modelled with a $9\times9$ covariance matrix feature descriptor \cite{covariance} and the relative position between targets which is less influenced by the camera motion (angle of the view) is computed. Since when the camera moves all the targets shift together, this motion feature is quite invariant to camera movements, making it a suitable representation applicable also to videos acquired by a moving camera. Then, a temporal sliding window technique is utilized to find tracklets and perform further merging on them.  More specifically, given detection responses found along the temporal window, we will represent them as a graph in which all the detections in each frame are connected to all the other detections in the other frames, regardless of their closeness in time, and the edge weight depicts both appearance and position similarity between nodes. 

A two-level association technique follows: first, low-level association is performed linking detection responses of the last two consecutive frames of the window, which helps differentiating difficult pairs of targets in a crowded scene; then, a global association is performed to form tracklets along the temporal window. Finally, different tracklets of the same target are merged to obtain the final trajectory.

The main contributions presented in this chapter are:
\begin{itemize}
	\item this chapter presents the first example of the use of dominant set clustering for data association in multi-target tracking;
	\item the two-level association technique proposed in this chapter allows us to consider efficiently (also thanks to the superior performance of dominant set clustering on identifying compact structures in graphs) the temporal window at once, performing data association in a global manner; this helps in handling long-lasting occlusions and target disapperance/reappearance more properly;
	\item the consensus clustering technique developed to merge tracklets of the same target and obtain the final trajectories is firstly introduced in this work, although it has been used in different domains;
	\item the proposed technique outperforms state-of-the-art techniques on various publicly-available challenging data sets.  
\end{itemize}

The rest of the chapter is organized as follows: related works are discussed in Section \ref{related work}; while Section \ref{Multi-object Tracking through DSC} details our tracking framework (DSC tracker, hereinafter). Experimental results are shown in Section \ref{experiment}, followed by summary in section \ref{summary_chapter_DSC}.
\section{Related Work}
\label{related work}
Target tracking has been and still is an active research area in computer vision and the amount of related literature is huge. Here, we concentrate on some of earlier related works on tracking-by-detection (or ABT) methods.

Recently, tracking-by-detection methods \cite{anton2011,Perera,Pirsiavash, xing, nevatia,AmiShaECCV12} become the most exploited strategy for multi-target tracking. Obtaining multiple detection responses from different frames, it performs data association to generate consistent trajectories of multiple targets. As the number of frames and targets increases, the complexity of data association problem also increases along with it. As a consequence, most approaches aim to either approximate the problem or to find locally-optimal solutions. Early methods for multi-target tracking includes joint probabilistic data association (JPDA) \cite{Jpda} and multi-hypothesis tracker (MHT) \cite{MHT}. These techniques aim to find optimal assignment over the heuristically-pruned hypothesis tree which is built over several frames of the video.

More recently,  researchers formalize data-association task as a matching problem, in which detections in consecutive frames with similar motion pattern and appearance are matched. Bipartite matching is the best known example of such methods \cite{bipartite}: the method is temporally local (considering only two frames) and utilizes Hungarian algorithm to find the solution. However, their approach suffers from the limited-temporal-locality in cases where target motion follows complex patterns, long-lasting occlusions are present or targets with similar spatial and appearances exist. On the other hand, other researchers in \cite{AmiShaECCV12,Pascal,navatia12,dp} follow a method generally termed as \textit{global}: recently this approach is becoming more popular since it allows to remove the limited-temporal-locality assumptions and this allows them to incorporate more global properties of a target during optimization, which helps overcoming to problems caused by noisy detection inputs. 

In \cite{navatia12} data association problem is mapped into a cost-flow network with a non-overlap
constraint on trajectories. The optimal solution is found by a min-cost flow algorithm in the network. In \cite{dp}  the graph was defined  similar to \cite{navatia12} and showed that dynamic programming can be used to find  high quality sub-optimal solutions. The paper in \cite{Pascal} uses a grid  of potential spatial locations of a target and solve association problem using the K-Shortest Path algorithm. Unfortunately, their method completely ignored appearance features into the data association process, which result unwarranted identity switches in complex scenes. To solve this limitation, the proposal in \cite{Horesh} incorporates the global appearance constraints.

The above-mentioned global methods use simplified version of the problem by only considering the relationships of detections in consecutive frames. Despite the good performance of these methods, they came short in identifying targets with similar appearance and spatial positions. Conversely, in recent approaches \cite{anton2011,AmiShaECCV12}, no simplifications in problem formulation are made. However, the proposed solutions are approximate due to the complexity of the models. More specifically, authors in \cite{anton2011} proposed a continuous energy minimization based approach in which the local minimization of a non-convex energy function is performed exploiting the conjugate gradient method and periodic trans-dimensional jumps. Compared to the aforementioned global methods, this approach is more suitable for real-world tracking scenarios. On the negative side, however, the solutions found to the non-convex function can be attracted by local minima.

Recently, in \cite{AmiShaECCV12} tracking problem is represented in a more complete manner, where pairwise relationships between different detections across the temporal span of the video are considered, such that a complete K-partite graph is built. A target track will be represented by a clique (sub graph where all the nodes are connected to each other), and the tracking problem is formulated as a constraint maximum weight clique problem. However, since a greedy local neighborhood search algorithm is followed to solve the problem, also this approach is prone to the local minima case. Moreover, due to heuristic line fitting approach used for outlier detection, the approach is prone to identity switches, particularly when targets with close position move together.

\section{Multi-object Tracking through Dominant Set Clusters}
\label{Multi-object Tracking through DSC}
\begin{figure*}[htb]
	\begin{center}
		\includegraphics[width=1\linewidth ,trim=0cm 9cm 0cm 4.3cm, clip]{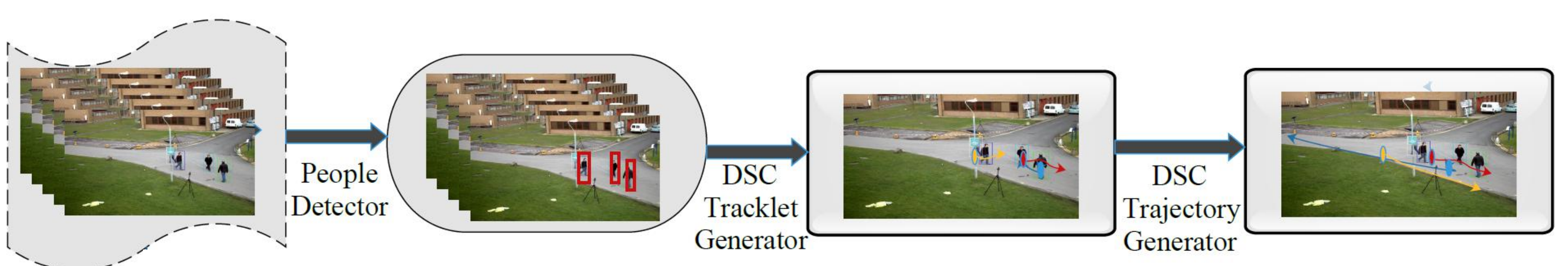}
	\end{center}
	
	\caption{Overview of the proposed approach}
	\label{fig:Overview}
\end{figure*}

The overall scheme of the proposed system is depicted in Figure \ref{fig:Overview}. Our tracking framework is composed of three phases. As a first phase, objects of interest need to be detected in each frame. In this chapter and corresponding experiments in Section \ref{experiment}, we will refer to people as targets/objects of interest, but the approach described in the following can be easily adapted to different objects, supposed that a proper detector is available. In fact, any detection technique can be used, although in this work we will use the well-known people detector based on HOG proposed in \cite{HOG}. Once objects are detected as candidate bounding boxes, labels must be consistently assigned to different instances of the same person. To this aim, in the second phase we employ a \dquote{sliding temporal window} approach that iteratively constructs tracklets using dominant set clustering. Finally, tracklets found along the whole video will be merged to form a trajectory over the whole course of the video. The next subsection will detail the last two phases, while readers interested in details about first phase can refer to \cite{HOG}.

%
%

In order to obtain more accurate results with more efficiency (i.e., lower latency) it is vital to use sliding temporal windows to track targets. Moreover, in the whole video there might be many targets most of which may not have temporal overlap. As a consequence, performing a global optimization over the whole video frames is not only impractical, but also inefficient due to the huge search space and long latency, and also needless as many targets only appear for some period but not on the whole course of the video. 

\begin{figure*}[h!]
	\begin{center}
		\includegraphics[width=1\linewidth ,trim=0cm 3cm 0cm 1cm, clip]{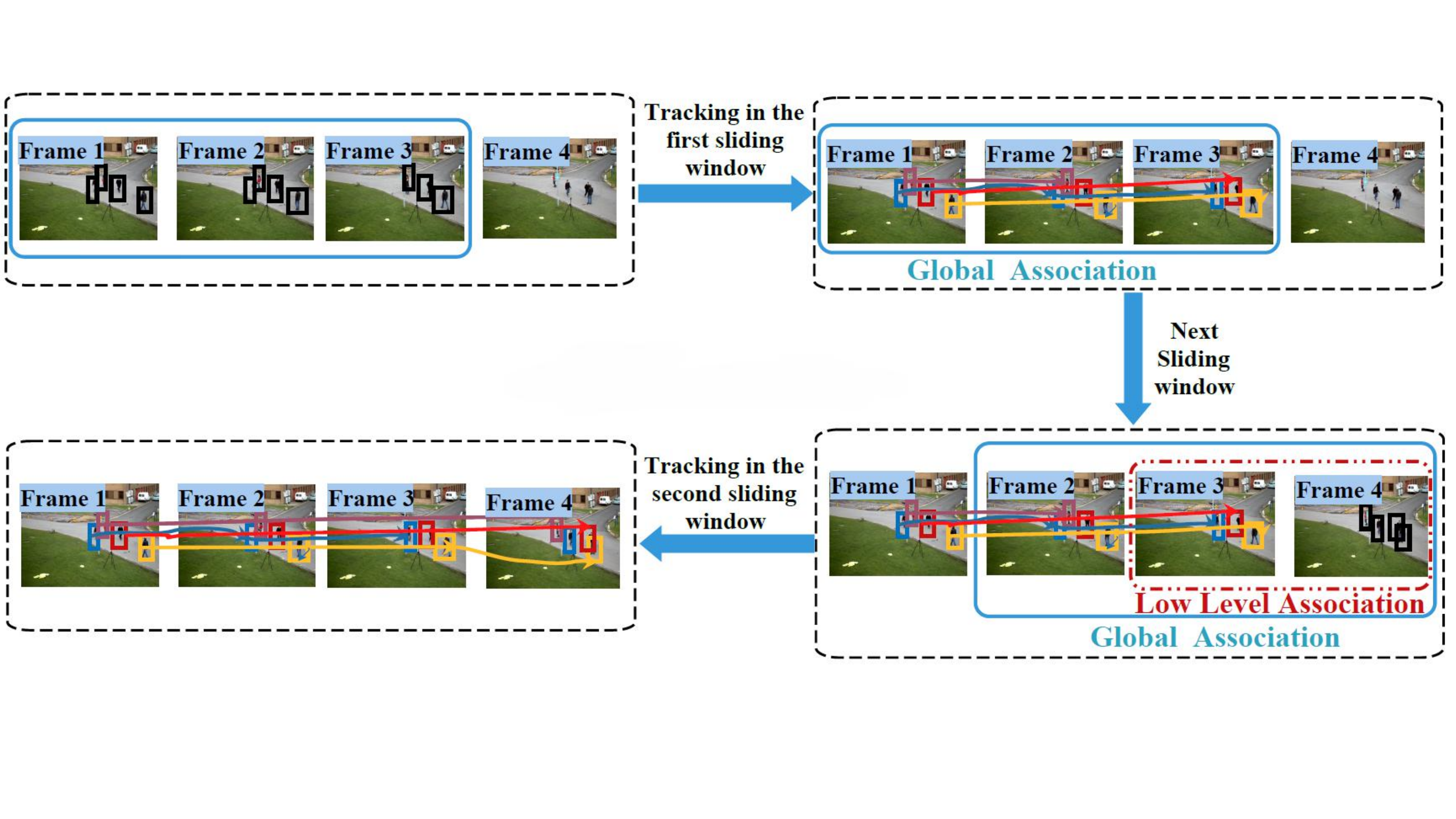}
		
	\end{center}
	
	\caption{Toy example of sliding window approach. Bounding boxes denote detection responses, and the different colors represent different targets. The sliding window moves one frame per step. \textbf{This figure is best viewed in color}.}
	\label{fig:slidingwindow}
\end{figure*}

To generate tracklets over the sliding temporal window, we use overlapping neighbouring windows of frames. A toy example is given in Figure \ref{fig:slidingwindow} to ease the description of the approach. Given an input video and the detection responses found within the first sliding window, tracking is performed within the window. Then, the window is moved with the step of one frame at a time, so that the new window will have more than 90\% overlap with the previous one, and the detection algorithm is applied on the new frame (number 4 in Fig. \ref{fig:slidingwindow}) so to generate detection responses. At this point, a \textit{low-level association} algorithm (detailed in section \ref{subsection:tracklate generation}) is employed to associate the detection responses found in the last two frames (3 and 4) which have high appearance similarities and small spatial distances. Then, a \textit{global association} algorithm (see section \ref{subsection:tracklate generation}) is applied over all the frames in the current window in order to create consistent tracklets. As a result, we are able to associate efficiently and accurately the new frame detection responses to the right tracklet.

It is worth noticing that selecting the size of the sliding window is crucial. In fact, there is a trade-off between latency and accuracy: the bigger the size (more frames) the more accurate the association task will be since more data are available, but at the cost of higher latency (preventing online real-time functionality); on the other hand, the smaller the size, the quicker the results will be available, but with lower accuracy. In our experiments (reported in section \ref{experiment}), we select a window size between 5 and 15 frames depending on the dynamics as well as frame rate of the video: the faster the frame rate, the bigger will be the size.

\subsection{Tracklet Generation in a Single Window}\label{subsection:tracklate generation}

As described above, the generation of tracklets in each single window is obtained by applying a two-level association procedure, low-level and global.

\textbf{Low-level association:} 
Low-level association is performed to connect detection responses found in the last two frames (which have high appearance similarities and small spatial distances) into reliable tracklets (pair of responses). This will enable us to separate responses which have very small temporal differences and also distinguish between targets in complex scenes with similar appearances, since the people will not drastically change in appearance as well as position in just two consecutive frames.

Once extracted, the detection responses from the last two frames are represented as nodes on a graph (as described in section \ref{subsect:DominantSetClustering}), where edges define connections between them and \emph{edge weights reflect both spatial and appearance similarity between nodes}. 
Let $G_s =(V_s,E_s,\omega_s)$ be our graph constructed using the detection responses of the last two frames where $V_s=\{v_{1}^{a},v_{2}^{b},.....\}$ and the vertice $v^{i}_{j}$ represents the $j^{th}$ node from the $i^{th}$ frame. We denote for the sake of simplicity with $a$ and $b$ the second last and last frames, respectively. The set of edges $E_s=\{(v_{q}^{i},v_{r}^{j})|(i\neq j)\}$ is composed of edges which connect \textit{only nodes coming from diffrent frames}, while $\omega_{s}$ is the positive edge weight function. 

As explained in section \ref{subsect:DominantSetClustering}, given $n$ nodes, we can represent the graph with a $n \times n$ affinity matrix $A_s = (a_{ij})$ where $a_{ij} = \omega(i,j), $ if $(i,j)\in E_s$, and $a_{i,j}=0$ otherwise. Using this representation, one association pair of responses of person $i$ is simply a \textit{dominant set} $G_{s}^i =(V_{s}^{i},E_{s}^{i},\omega_{s}^{i})$ in which $V_{s}^{i}$ contains detection responses of only target $i$. Here the task is to find a pair of detection responses (if any) which has maximum similarity. For this task we utilized the DSC technique described in detail in section \ref{subsect:DominantSetClustering}.

\textbf{Global association:} Low-level association results are useful to initialize tracklet generation based on reliable associations (based on two consecutive frames). However, it may happen that some people are not visible in the last two frames of the current window. As a toy example, low-level association in Figure \ref{fig:slidingwindow}(c) on frame 3 and 4 fails in creating an association for the pink person, since it is visible only in frame 4 and not frame 3. As a consequence, a global association algorithm ought to be used over the whole window. As a result, the tracklet generated in the previous window for the pink person (between frames 1 and 2 in Figure \ref{fig:slidingwindow}(b)) can be successfully associated to the new detection in frame 4 (see Figure \ref{fig:slidingwindow}(d)).

Similarly to the low-level association case, we denote the detection responses to be tracked in one temporal window of $f$ frames as an  undirected edge weighted graph $G_n =(V_n,E_n,\omega_n)$ with no self loop, represented as a similarity matrix $A_n$. One critical point in the global association algorithm is that the knowledge about both the previous window and the low-level association is exploited to update the new association matrix $A_n$. In particular, let $G_{o}^{k}$ be the subgraph representing tracklet of person $k$ with $V_o^k$ vertices from the previous window. Then, the similarity matrix $A_n = (a_{ij})$ is computed as

\begin{equation}
a_{i,j}  =
\begin{cases}
1 &  \text{if $(i,j)\in V_{o}^{k}$ or $(i,j)\in  V_{s}^{k}$}\\
0&\text{$(i,j)\notin E_n$}\\
\omega_n(i,j) &\text{$otherwise$}
\end{cases}
\label{eqn:updaterule}
\end{equation}
\noindent where $V_{s}^{k}$ represents tracklet of target $k$ obtained from low-level association,  $(i,j)\in V_{o}^{k}$ or $(i,j)\in  V_{s}^{k}$ means both node $i$ and $j$ belong to same tracklet of $k^{th}$ target obtained from the previous window or low-level association, respectively.

In order to capture the individual similarities in people (patches) of the same target and differentiate between different targets, it is compulsory that the graph is made using a meaningful and robust similarity measure. A node $v^{i}_{j} $ is represented with a location feature $l^{i}_{j}$ , which is the 2-dimensional spatial coordinates of the bottom center  and upper left corner of the corresponding detection.

Moreover, we decided to model people appearance using covariance matrix feature descriptors \cite{covariance}. This representation has been adopted in various approaches \cite{appearance,people,Liu,Smeulders} thanks to its robustness in capturing information like color, shape and location and also to its scale and rotation invariance. Considering \textit{$d$} different selected pixel features extracted independently from the image patch, benefiting from being a low dimensional representation, the resulting covariance matrix $C$ is a square symmetric matrix $d \times d$ where the diagonal entries represent the variance of each feature and the non-diagonal entries represent the correlations. 

Let us consider \textit{Im} as a three-channels color image and \textit{Y} be  the $W \times H \times d$ dimensional image feature extracted from $\textit{Im}$. Let $Y(x,y) = \rho(Im,x,y)$, where the function $\rho$ could be any mapping such as gradients, color, filter responses, intensity, etc.. Let {$\{t_i\}_{i=1...M}$ be the d-dimensional feature points inside \textit{Y}, with $M = W \times H$. The image \textit{Im} is represented with the $d \times d$ covariance matrix of the feature points:
	
	\begin{equation}
	C_R =\dfrac{1}{M-1} \sum_{i=1}^{M}(t_{i}-\mu)(t_{i}-\mu)^{T}
	\label{eq:cov}
	\end{equation}
	
	\noindent where vector $\mu$ represents the mean of the corresponding features of each point in a region $R$.
	
	In our case, we decided to model each pixel within a people patch with its HSV color values, its position $(x,y)$, $G_x$ and $G_y$ which are the first order derivatives of the
	intensities calculated through Sobel operator with respect to $x$ and $y$, the magnitude $mag(x,y)$ = $\sqrt{G_{x}^{2}+G_{y}^{2}}$ and the angle of the first derivative $\Theta(x,y) = arctan \left(  \dfrac{G_y}{G_x}\right) $. Therefore, each pixel of the people (patch) is mapped to a 9-dimensional feature vector: $t_i = [x\:; y\:; H\:; S \:; V \:; G_x \:; G_y \:; mag(x,y) \:; \Theta(x,y) ]^T $. Based on this 9-dimensional feature vector representation the covariance of a patch is $9 \times 9$ matrix.

	The distance between covariance matrices is computed using technique proposed in \cite{covariance,distance} which is the sum of the squared logarithms of their generalized eigenvalues. Formally, the distance between two matrices $C_{i}^{j}$ and $C_{m}^{n}$ is expressed as:	
	\begin{equation}
	\varpi(C_{i}^{j},C_{m}^{n})=\sqrt{\sum_{k=1}^{d}ln^{2}\lambda_{k}(C_{i}^{j},C_{m}^{n})}
	\end{equation}	
	\noindent where $\lambda_k(C_{i}^{j},C_{m}^{n})_{k=1....d}$ are the generalized eigenvalues of $C_{i}^{j}$ and $C_{m}^{n}$.
	
	The weight of an edge between two nodes, $\omega : E\rightarrow\mathbb{R}^ {+}$, represents both appearance and spatial  similarities (correlation) between detection responses and the distance between two nodes is computed as :
	\begin{equation}
	D(v_{j}^{i},v_{m}^{n})=\sqrt{\kappa \cdot\varpi(C_{j}^{i},C_{m}^{n})^{2}+ \iota \cdot d(l_{j}^{i},l_{m}^{n})^{2}}		
	\label{eq:distance}
	\end{equation}
	
	\noindent where $ d(l_{j}^{i},l_{m}^{n})= \Vert l_{j}^{i}-l_{m}^{n}\Vert^2$ ($l_{j}^{i}$ and $l_{m}^{n}$  representing the location feature vectors of the corresponding nodes) and $\kappa$ and $\iota$ are values which are used to control contributions of appearance and location features, respectively.
	
	The affinity between the two nodes is defined as: 
	\begin{equation}
	\label{eq:similarity}
	\omega(v_{j}^{i},v_{m}^{n}) = \exp(-\dfrac{D(v_{j}^{i},v_{m}^{n})}{2\gamma^2})
	\end{equation}	
	where $\varpi$ is the regularization term. The smaller the sigma, less likely are the detection responses to be associated, leading to fewer identity switches but more fragments (interruption of the correct trajectory) and vice versa.

	Our task of finding the tracklet of one target in one temporal window requires to identify a target in each frame and then to represent the feasible solution as a dominant set (subgraph) $G_f$  in which at most one node (detection response) is selected from each frame's detection responses which are highly coherent and similar to each other. Therefore, we represent the dominant set as a subgraph $G_{f} = (V_f,E_{f},\omega_{f})$. We ought to note that the feasible solution (tracklet) $G_f$ only contains the detection responses of one target and not all visible detections found in the temporal window. In Figure \ref{fig:tracklet_example}(c) we can see one feasible solution found from current temporal window of size 3.  
	

	By solving DSC for the graph $G_n$ of the global association algorithm, we will end up with the optimal solution that is a dominant set ($\equiv$ subgraph $G_f$) containing detections of one target, which corresponds to the feasible solution with the most consistent and highly coherent in both spatial and appearance features over the whole span of temporal window. Moreover, in order to find the tracklets of all the people found in a temporal window, the optimization problem \eqref{eq2} needs to be solved several times. Therefore, at each iteration, the algorithm finds the tracklet which has the highest internal coherency. Then, vertices selected in $G_f$ will be removed from $G_n$ and the above optimization process is repeated to find the tracklet for the next target, and so on until zero or few nodes remains in $G_n$. This approach is commonly referred as \textit{\dquote{peeling off strategy}}.
	
	\begin{figure*}[htb]
		\begin{center}
			
			\includegraphics[width=1\linewidth ,trim=0cm 4.2cm 0cm 1cm, clip]{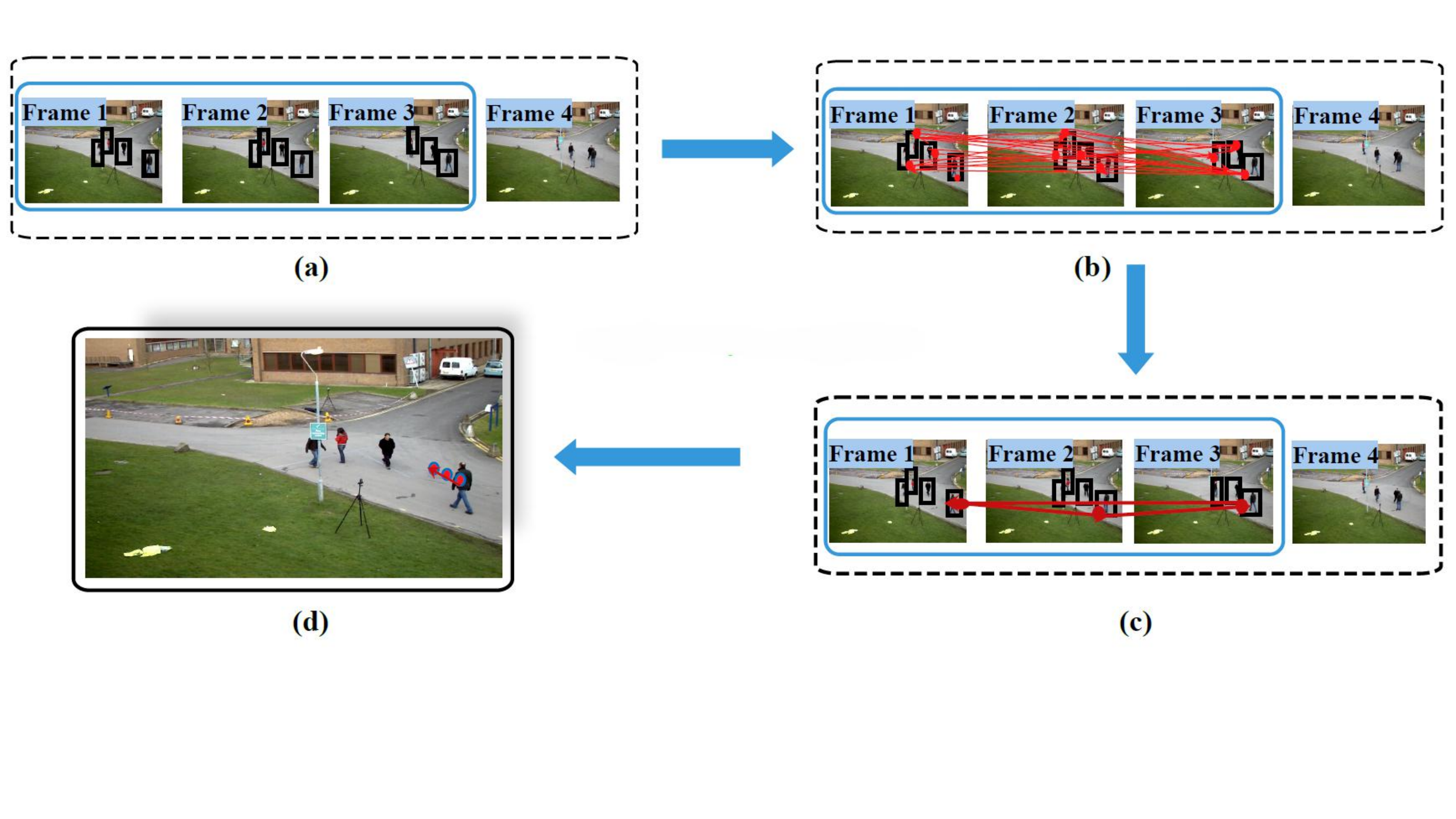}
		\end{center}
		\caption{Illustration of tracklet generation in one temporal window;(a) shows the possible detection responses in the window; (b) shows the graph built between the detection responses; (c) shows the tracking result from one iteration, containing dominant sets of one target as a subgraph; (d) shows the obtained trajectory. \textbf{This figure is best viewed in color}.}
		\label{fig:tracklet_example}
	\end{figure*}
	
	\subsection {Tracklet Merging using DSC}
	\label{section: merging}
	
	If a person is occluded or exits from the scene for a period which is longer than the temporal window size, the association algorithm described in the previous subsection will not work properly, since a new label/id will be assigned to the person when he/she reappears in the scene. As a consequence, we need to merge those tracklets representing the same target into one single trajectory along different windows.
	
	For this data association problem, we once again utilized the same DSC-based data association method. It is worth emphasizing that merging different tracklets of one person on different windows is by far the hardest task. In fact, it is very common that the person's appearance is changed when he/she reappears in the scene, for both the possible changes in illumination conditions and the different poses with which the person re-enters in the scene. To solve this problem, we need a robust approach accounting for different situations which might arise like: target might appear to the scene heavily occluded and stay occluded for the most part of the tracklet; a target might enter in the scene with a different pose but then might get back to his/her original pose for the rest part resulting similar tracklets in their (average) appearances similarity; etcetera. For this reason, it turns out that there exists no a single approach or similarity measure capable to handle all these situations carefully. Therefore, we borrow the idea from \textit{consensus clustering} \cite{clustering_ensemble,consensus_Wang_2014_CVPR,consensus2}. Consensus clustering (also known as \textit{clustering ensemble}) combines different clustering techniques and able to exploit advantages from all of them and to handle all (or most of) the situations above. 
	
	First of all, we consider only tracklets of length greater than 10 nodes since very short tracklets were considered as false positives. Then, three different approaches are used concurrently:
	
	\begin{enumerate}
		
		\item In the first approach, each tracklet is divided in two equal parts: each tracklet partition is represented as a node in a graph and the weight between the nodes (tracklets) is computed as a mean distance between their appearance features. Let $I$ and $ J $ be two tracklets ($\equiv$ nodes), then the distance between the two tracklets is formulated as: 
		\begin{equation}
		D_{ {I}, {J}} = \frac{1}{|I|} \left\lbrace \sum\limits_{ i \in I } \frac{1}{| J|}\left(\sum\limits_{ j \in  J }\varpi(C_i,C_j)\right) \right\rbrace
		\end{equation}
		
		The similarity matrix is built as $A^\mu = (a_{ I  J}^\mu)$ where $ a_{ I  J}^\mu= \exp(- D_{ {I}, {J}})$.
		This approach is adequate in merging two tracklets (of the same person) where the target  appears more or less with the same appearance in both tracklets for the most part, since it depicts the mean distance over their appearance similarity the approach is prone to few big changes made in appearance. But the approach fails in cases where the target appears in a totally different pose for  most of the considered frames.
		
		\item The second approach again divides each tracklet in two equal parts, represented as nodes in a graph. However, in this case the weight is computed by taking the minimum of their mean distances on their appearance features only:

		\begin{equation}
		D_{{I}, {J}} = \min\limits_{ i\in  I}\frac{1}{| J|}\left(\sum\limits_{ j \in  J}\varpi(C_i,C_j)\right)
		\end{equation}
		
		The similarity matrix is built as: $A^{M} = (a_{ I  J}^ M)$ where $a_{ I  J}^ M =  \exp(- D_{ {I}, {J}})$.
		
		Unlike the previous technique this approach works best in merging tracklets where the target makes a big change in appearance and stayed the same for the most part of the tracklet. In such cases, taking the minimum distance between their appearances is the best choice, supposed that there exist at least some frames/nodes where the appearance is similar. This approach will come short in cases where most of the targets have high appearance similarity between each other, causing identity switches.

		\item The third approach represents each tracklet by means of their detection with highest weighted characteristic vector value (strict local solution of the optimization equation \ref{eq2}). In other words, the best representative detection of each tracklet is selected for comparing with the other tracklets. The distance between the nodes of the new graph will be computed by taking their difference in appearance.
		More specifically, let us consider a node $i$ in a tracklet $ I$, $ i \in  I$, to be that with highest characteristic vector value:
		\begin{equation}
		C_{i} = \argmax_{ i} \frac{w_{ I}(i)}{ W( I)}
		\end{equation}
		
		The similarity matrix is built as: $A^{C} = (a_{ I  J}^ C)$ where $a_{ I  J}^ C=\exp(-\gamma(C_i,C_j))$.  
		
		In cases where the two separate tracklets are created due to occlusion between people which lasts longer than the size of the temporal window, it is less likely that the person will make a big change on his/her new appearance. Hence, it will be enough to represent a tracklet with  two representative detection responses with the highest characteristic vector values. But yet again this technique fails if the selected representative detection results to be a bad representative (e.g., it can be partially occluded), generating multiple identity switches.

	\end{enumerate}
	DSC is performed three times, each using one of the different approaches listed above ($A^\mu$,$A^ M$ and $A^ C$). The clustering results are then combined in to a single affinity matrix according to the Evidence Accumulation Paradigm \cite{evidence_accumulation}. We built a matrix $ B = (b_{ij})$ known as \emph{co-association matrix}: where  $ b_{ij}=0$ if $i=j$, otherwise $ b_{ij}=\varphi(i,j)$, being $\varphi(i,j)$ the fraction of times object $i$ and object $j$ were clustered together among all $C$ clustering results in the ensemble and computed as:
	\begin{equation}
	\varphi(i,j)=\frac{1}{|C|}\sum\limits_{r=1}^{|C|} \delta( C_r(i), C_r(j))
	\end{equation}
	
	\noindent where $C_r(i)$ is the label of $i^{th}$ object in the $r^{th}$ clustering result and $\delta(n,m)$ is Kronecker's delta function, that is, $\delta(n,m)= 1$ if $n=m$ and 0 otherwise.

	To get the final clustering result, we run the DSC on the co-association matrix.

	\section{Experiments}
	\label{experiment}
	\subsection{Experimental setup}
	We evaluate our approach on three publicly available data sets: sequence S2L1 from PETS2009 benchmark \cite{pets}, TUD-standemitte dataset \cite{anton2011}, and "sunny day" sequence from ETH mobile dataset \cite{eth}, which are commonly used by previous multi-target tracking works. To be fair and transparent on the comparison, we used publicly available detection responses \cite{Bo_sunny,online_crf}  and ground truth \cite{MOTChallenge2015} in all our experiments. We also used publicly available code \cite{Milan:2015:CVPR} for the evaluation.
	
	\textbf{Brief description of the datasets:} PETS2009-S2L1-View one \cite{pets} consists of 795 frames and comprised of challenges like non-linear target motion, targets in close proximity and several occlusions in the scene. TUD-Stadtmitte \cite{anton2011} consists of 179 frames recorded in a busy pedestrian street with low camera angle, which generates frequent occlusions. Both these datasets are recorded using static cameras. On the contrary, ETH dataset (\dquote{sunny day} sequence) \cite{Sunnyday_dataset} is recorded with a pair of moving cameras in a busy street scene. The cameras move forward and have a panning motion at times, which potentially makes the exploitation of known motion models less reliable. Most targets move with a speed of 2-4 pixels per frame, whereas the camera exhibits a motion of around 20 pixels per frame. This causes imprecise results while using linear motion estimations. However, the relative positions of the targets do not change between two consecutive frames since all targets move according to the same camera motion. The sequence contains 354 frames and people size detected on image plane varies significantly. Similar to \cite{nevatia_sunny}, we used the sequence from the left camera. No ground plane information is used.

	\textbf{Parameter analysis:} There is no fixed window size which will work for all the datasets, rather depending on the dynamics and frame rate of the video. Therefore, we set window size of 15,10 and 5 for PETS2009-S2L1, TUD-standemitte and ETH-sunnyday sequences, respectively. 
	Our algorithm performs well in a wide range of $\gamma$ (the scaling parameter in eq. \ref{eq:similarity}) from 0.2 to 3.0. The good invariance (or low sensitivity) to $\gamma$ parameter is due to the exploitation of results from both the previous window and the low-level association along the current window to update our affinity matrix, which results in the replacement of most of the values of our affinity matrix (with the value 1, if the two nodes belong to same cluster according to results from previous window or low-level association and with the value 0 otherwise).

	For PETS2009-S2L1 and TUD-standemitte, $\kappa$ and $\iota$ in eq. \ref{eq:distance} (which are factors for controlling the contributions of appearance and position information, respectively) are typically set to one which will give equal weight to appearance and position features. However, based on our experiments, the appearance features are more informative than position in our formulation. In fact, in many cases using appearance features only is sufficient for tracklet generation. In the case of  ETH-sunny day sequence, we set  $\kappa$ and $\iota $ values to 1 and 1.25 respectively to get the best result. In fact, since this sequence is recorded with a moving camera, large changes in poses and sizes of the people are present and this results in significant changes in their appearance. Consequently, appearance information is less informative than positions in this case.
	
	
	\textbf{Evaluation methodology:} The correct and fair evaluation of multi-target tracking systems relies mainly on two issues: the definition of proper and agreed evaluation metrics; and the use of publicly-available ground truth on which the evaluation can be based.

	Regarding the evaluation metrics, we adopted those defined in \cite{Milan:2015:CVPR} which uses the most widely accepted protocol CLEAR MOT \cite{clear}. 
	CLEAR MOT defined several values: 
	\begin{itemize}
		\item \emph{Multi-Object Tracking Accuracy (MOTA)} combines all error types (false positives (FP), false negatives/missing detections (FN) and identity switches (IDS)) - the higher the better; 
		\item \emph{Multi-object Tracking Precision (MOTP)} measures the normalized distance between tracker output and ground truth location, i.e. the precision in the bounding box (or center of gravity) localization - the higher the better; 
		\item \emph{Mostly Tracked (MT)} measures how many ground truth (GT) trajectories are successfully tracked for at least 80\% of frames - the higher the better; 
		\item \emph{Mostly Lost (ML)} measures how many of the ground truth (GT) trajectories are tracked for less than 20\% of the whole trajectory - the lower the better; 
		\item \emph{identity switches} (IDS): number of times that a tracked trajectory switches its matched ground truth identity - the lower the better. 
	\end{itemize}
	
	In addition to those metrics, we also compute the following values:
	\begin{itemize}
		\item \emph{False alarm per frame (FAF)} measures the average false alarms per frame - the lower the better; 
		\item \emph{Precision} (Prcsn) is the average of correctly matched detections per total detections in the tracking result - the higher the better; 
		\item \emph{Recall (Rcll)} computes the average of matched detections per total detections in the ground truth - the higher the better.
	\end{itemize}
	
	All the reported results in Tables \ref{table:Tablepets} and \ref{table:TableTUD} are generated using tracking outputs provided by the authors \cite{Milan:2015:CVPR} with the same overlap threshold for CLEAR MOT metrics. Instead, quantitative results of the compared approach reported in Table \ref{table:Tablesunny} are taken from \cite{nevatia_sunny}.

	Regarding the ground truth, in order to provide a common ground for further comparison, we used the publicly-available detection responses from \cite{online_crf} for PETS2009 and TUD datasets and those from \cite{nevatia_sunny} for ETH. As ground truth, for all the datasets we used that provided in \cite{MOTChallenge2015}. It is worth noting that we achieved exactly the same results even when we used modified (stricter) ground truths, assuring that all the people have one unique ID troughout the whole video. In fact, it often happens that when a person disappears from the scene (due to occlusions or because he/she exits temporary from the scene), when he/she reappears a different ID is assigned (also in the public ground truth). The stricter ground truth, instead, assigns the same ID, which is not always the case in \cite{MOTChallenge2015}.

	\subsection{Results on Static Camera Videos}
	Let us first introduce the results in the two most used datasets for multi-target tracking (PETS2009-S2L1 \cite{pets} and TUD-Stadtmitte \cite{anton2011}) which are recorded with static cameras.

	Our quantitative results on PETS2009-S2L1 are shown in Table \ref{table:Tablepets}. Compared with up-to-date methods, our DSC tracker attains best performance on MOTA, precision, IDs and FAF over the majority of the approaches in Table \ref{table:Tablepets}, while keeping recall and MT comparable.

	Some visual results are presented in Fig. \ref{fig:smplresult}, first two rows (first row shows our results, while second row reports the ground truth). Even if targets appear quite close and similar in appearance, our approach is still able to maintain correct IDs, as in the case of targets 3 and 7 on frame 149 or targets 3 and 8 on frame 149 and 161. Furthermore, thanks to our robust tracklet merging step, our approach is capable to correctly identify a target on his/her reappearance after a long absence from the scene, as in the case of targets 1, 3, 4, 5 and 6 on frame 716. Please note that the ground truth (second row of Fig. \ref{fig:smplresult}) contains different IDs when the targets reapper in the scene (see also the comment reported above about the grund truth).

	\begin{table}[bth]
		\caption{Tracking results on PETS2009-S2L1 sequence. For all approaches the number of ground truth (GT) trajectories is the same (19).}
		\label{table:Tablepets}
			\begin{tabular}{clclclclclclclclclclclc} 
				\hline\noalign{\smallskip}
				Method &  MOTA(\%)  & MOTP(\%) &  Rcll(\%) &  Prcsn(\%) &  FAF  &  MT & ML  & IDs   \\
				\hline
					\cline{1-9}
				\cite{dp} & 81.8 & 71.5     &	 89.2 & 96.6   & 0.19  & 		17 &	\textbf{0} &	97 \\ \cline{1-9}
				\cite{cl2} & 53.4 & 70.9   &	72.6 & 80.3 & 1.04   & 	7 &	1 &	65 \\ \cline{1-9}
				\cite{dco} &  \textbf{90.8} &  \textbf{74.2}      &	 97.1 & 94.4  & 0.34  & 		18 &	\textbf{0} &	26 \\ \cline{1-9}
				\cite{Milan:2015:CVPR} &  87.9 &	64.5  &   	\textbf{98.6}   &   	90.8      &  0.59  &     		\textbf{19} &	\textbf{0} &	29	 \\ \cline{1-9}
				\textbf{DSC} &    90.0 & 56.8        &	 91.7 &  \textbf{98.5}    & \textbf{0.08}  & 		17 &	\textbf{0} &	\textbf{15} \\ \cline{1-9}

				\hline
			\end{tabular}
	\end{table}

	\begin{figure} [h!]
		\centering
		\includegraphics[width=1\linewidth ,trim=7cm 0cm 9cm 0cm, clip]{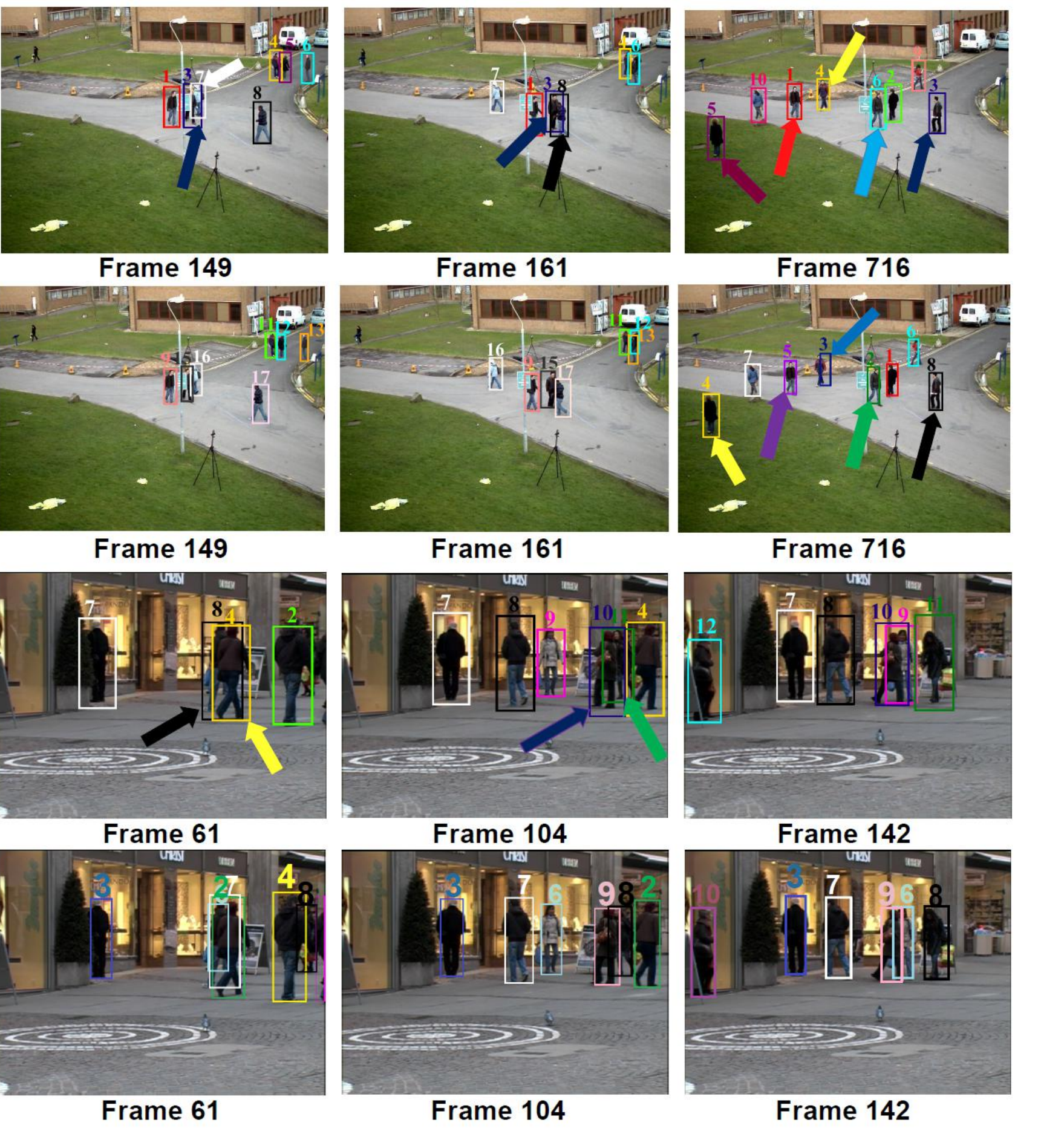}
		\caption{Tracking result on PETS2009-S2L1 and TUD-Stadtmitte datasets. First two rows refer to PETS2009-S2L1 (first row represents our results, while second row represents the ground truth), wherease the last two rows refers to TUD-Stadtmitte (with third row showing our results, while fourth showing the ground truth). \textbf{This figure is best viewed in color}.}
		\label{fig:smplresult}
	\end{figure}
	
	\begin{table}[bth]
		\caption{Tracking results on TUD-Stadtmitte sequence. For all approaches the number of ground truth (GT) trajectories is the same (10).}
		\label{table:TableTUD}
			\begin{tabular}{lllllllllllll } 
				\hline\noalign{\smallskip}
				Method &   MOTA(\%)  & MOTP(\%) & Rcll(\%) & Prcsn(\%) &  FAF &  MT & ML  & IDs   \\
			\cline{1-9}
				\cite{dp} & 67.6 & \textbf{72.6}     &   	75.0 & 92.9      &   0.37  &		5 &	0 &	96  \\ \cline{1-9}
				\cite{cl2} &  60.0 & 56.5       &   	\textbf{83.7} & 78.7        &  1.46  &		\textbf{7} &	0 &	12  \\ \cline{1-9}
				\cite{dco} &  53.5 & 72.4       &   	 79.1 &  82.0        &   1.12  &		\textbf{7} &	0 &	20  \\ \cline{1-9}
				\cite{Milan:2015:CVPR} & 69.7 & 53.4   &   	74.7 &  94.3    &   0.29  &		6 &	0 &	\textbf{4}  \\ \cline{1-9}
				\textbf{DCS}& \textbf{72.4} & 52.6         &   	75.1 & \textbf{99.8}      &  \textbf{0.01}  &		6 &	0 &	10  \\ \cline{1-9}
				\\
				\hline
			\end{tabular}
	\end{table}
	
	Quantitative results on TUD-Stadtmitte dataset are provided in Table \ref{table:TableTUD}. Our method attains superior results in MOTA, precision and FAF while remaining comparable in ML and IDs. However, our approach obtains relatively lower performance in recall: this is mainly because DCS tracker focuses on assigning the right person ID to detection responses generated by the people detector, i.e. no motion model (linear or any other) for prediction of next locations has been used. As a result, our method generates a higher number of false negatives.

	Fig. \ref{fig:smplresult} shows some visual tracking results on TUD-Stadtmitte sequence (third and fourth row). Our approach is able to maintain correct IDs, as in the case of targets 4 and 8 on frame 61 or targets 10 and 11 on frame 104, regardless of their similarity in position and appearance and the severe occlusions.

	\subsection{Results on Moving Camera Videos}

	Table \ref{table:Tablesunny} shows the results on ETH-Sunny day sequence recorded from moving cameras. Compared with \cite{nevatia_sunny}, our DSC tracker achieves best performance on precision and false alarm per frame, while having relatively higher number of identity switches. This is mainly due to the appearance and size of targets which highly vary along with camera movement. However, the higher precision of our approach shows that it is able to recover the correct IDs. Fig. \ref{sunny_smpl_result} shows few visual results of our method (first row shows our results, while second row represents the ground truth). Even if people are fully occluded for a long time and reappear, our approach is able to correctly re-identify the person, as in the case of target 5 on frame 79, 98 and 316. We ought to note that the ground truth (second row of Fig. \ref{sunny_smpl_result}) gives different IDs when the target reappear after a long-lasting occlusion, as in the case of targets 8 and 29 on frame 79 and 316, respectively.

	
	\begin{table}[bth]
		\caption{Tracking results on ETH-Sunny day sequence. For all approaches the number of ground truth (GT) trajectories is the same (30).}
		\label{table:Tablesunny}
			\begin{tabular}{lllllllllllllll} 
				\hline\noalign{\smallskip}
				Method  &  MOTA(\%)  & MOTP(\%) & Rcll(\%) & Prcsn(\%)  & 	 FAF  & 	 	MT  & 	ML  & 	IDS \\
			\cline{1-9}
				\cite{nevatia_sunny}  & - &   - &\textbf{77.9}  & 		86.7 	 & 		0.65 & 	 	 		\textbf{22}   & 		\textbf{3} &  	 	\textbf{1}\\ \cline{1-9}
				\textbf{DSC} &  61.5  & 66.8 &  71.4 &  \textbf{89.4} & \textbf{0.45}  &   19 &    \textbf{3}  & 8
				\\ \cline{1-9}
				\hline
			\end{tabular}
	\end{table}

	\begin{figure} [h!]
		\centering
		
		\includegraphics[width=0.9\textwidth]{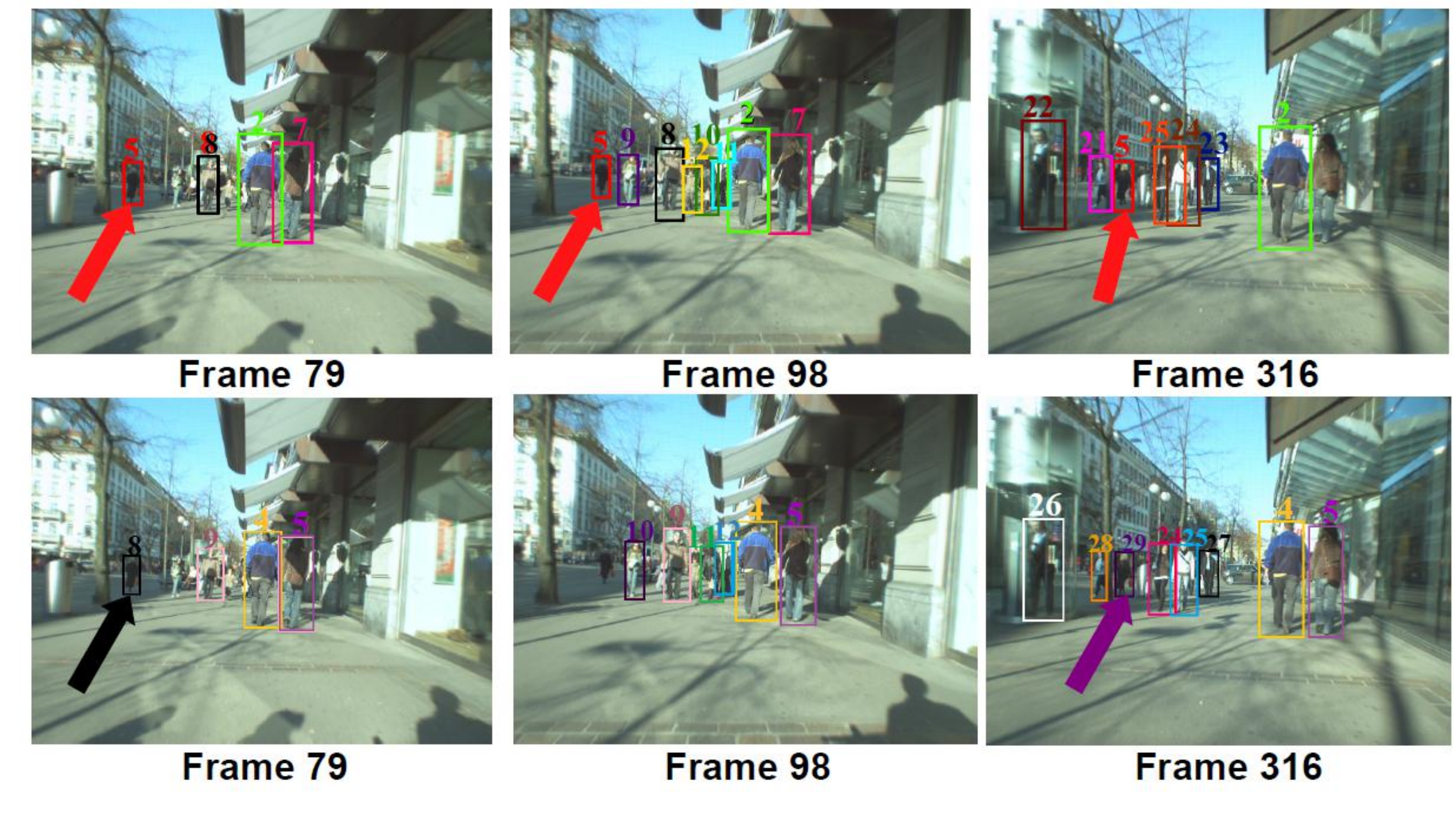}
		\caption{Sample Tracking result on ETH-Sunny day sequence (first row represents our results, while second row represents the ground truth). \textbf{This figure is best viewed in color}.}
		
		\label{sunny_smpl_result}
	\end{figure}


	\section{Summary}
	\label{summary_chapter_DSC}
	In this chapter, a dominant set clustering based tracker is proposed, which formulates the tracking task as finding dominant set (cluster) on the constructed undirected edge weighted graph. The development of a complete multi-target tracking using dominant set clustering is the main contribution of this work. We utilized both information (appearance and position) for data association in a \textit{global} manner, avoiding the locally-limited approach typically present in previous works. Experimental results compared with the state-of-the-art tracking approaches show the superiority of our tracker, especially in terms of MOTA and precision, as well as lower identity switches in some cases. Generally speaking, the tradeoff between precision and recall in tracking is hard to balance, thus resulting in lower performance on one of the two values. Our claim is that precision is more relevant for tracking purposes and that assigning the right ID to targets (people) when they re-appear after an occlusion or the exit from the scene should be the key objective of tracking systems.

	Regarding the efficiency of the proposed approach, for a window size of 10 frames with approximately 10 pedestrians, processing a frame using the full proposed framework, excluding human detection, takes an average time of 1.6 and 0.003 s for affinity matrix generation and tracklet generation, respectively. Moreover, it takes only 0.001 s for tracklet merging step. These values are computed by running a non-optimised Matlab code on a core i5 2.5 GHz machine. Using an optimised parallel implementation in C, the algorithm is likely to work in real time.
	
	As future directions of research, we first would extend the methodology to multi-camera setups. As a feeling, we believe that our approach will be straightforward to extend multiple cameras, since no motion or temporal information are used in it. Another possible future work consists in evaluating different similarity measures as well as to consider different types of targets instead of people.


	\chapter{Multi-target tracking in multiple nonoverlapping cameras using constrained dominant sets} \label{CDSC_tracker}
		
		In this chapter, we present a unified three-layer hierarchical approach for solving tracking problems in multiple non-overlapping cameras. Given a video and a set of detections (obtained by any person detector), we first solve {\em within-camera tracking} employing the first two layers of our framework  and, then, in the  third layer, we solve {\em across-camera tracking} by  merging tracks  of the same person in all cameras in a simultaneous fashion. To best serve our purpose, a constrained dominant sets clustering (CDSC) technique, a parametrized version of standard quadratic optimization, is employed to solve both tracking tasks. The tracking problem is caste as finding constrained dominant sets from a graph. That is, given a constraint set and a graph, CDSC  generates cluster (or clique), which forms a compact and coherent set that contains a subset of the constraint set. In addition to having a unified framework that simultaneously solves within- and across-camera tracking, the third layer helps link broken tracks of the same person occurring during within-camera tracking. We have tested this approach on a very large and challenging dataset (namely, MOTchallenge DukeMTMC \cite{RisSolZouCucTomECCV16,solera2016groups,ErgCarACCV2014}) and show that the proposed framework outperforms the current state of the art. Even though the main focus of this work is on multi-target tracking in non-overlapping cameras, proposed approach can also be applied to solve {\em re-identification} problem. Towards that end, we also have performed  experiments on MARS \cite{ZheBieSunWanSuWanTiaECCV16}, one of the largest and challenging video-based person re-identification dataset, and have obtained excellent results. 
		These experiments demonstrate the general applicability of the proposed framework for non-overlapping across-camera tracking and  person re-identification tasks.

	\section{Introduction}
	As the need for visual surveillance grow, a large number of cameras have been deployed to cover large and wide areas like airports, shopping malls, city blocks etc.. Since the fields of view of single cameras are limited,  in most wide area surveillance scenarios, multiple cameras are required to cover larger areas. Using multiple cameras with overlapping fields of view is costly from both economical and computational aspects. Therefore, camera networks with non-overlapping fields of view are preferred and widely adopted in real world applications.
	
	In the work presented in this chapter, the goal is to track  multiple targets  and maintain their identities as they move from one camera to the another camera with non-overlapping fields of views. In this context, two problems need to be solved, that is, within-camera data association (or tracking)  and across-cameras data association by employing the tracks obtained from within-camera tracking. Although there have been significant progresses in both problems separately, tracking multiple target jointly in both within and across non-overlapping cameras remains a less explored topic.  Most approaches, which solve multi-target tracking in multiple non-overlapping cameras \cite{YinKaiTieICPR08,AmiPanSurWeiBMVC04, OmaZeeKhuMubICCV03, OmaKhuMubCVPR05, YouSenMusMTA14,RomHorICPR08, JinIsaGerWACV05, CheBhaTCS16}, assume tracking within each camera has already been performed and try to solve tracking problem only in  non-overlapping cameras; the results obtained from  such approaches are far from been optimal \cite{YouSenMusMTA14}. 
	
	\begin{figure*}
		\centering 
		\includegraphics[width=1.07\linewidth,trim=1.4cm 2.58cm 0cm 2.88cm,clip]{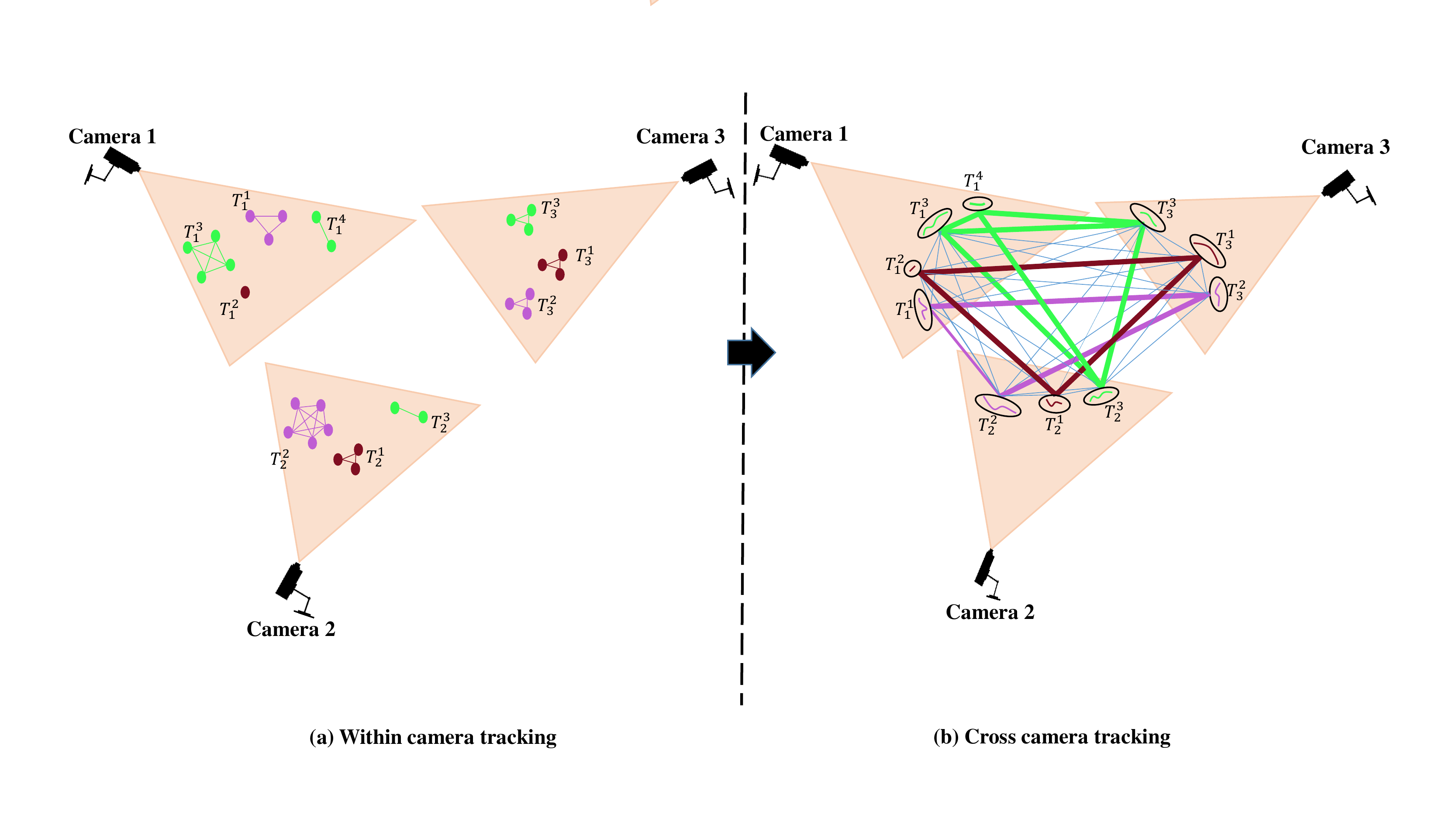}
		\caption{A general idea of the proposed framework. (a) First, tracks are determined within each camera, then (b) tracks of the same person from different non-overlapping cameras are associated, solving the across-camera  tracking. Nodes in (a) represent tracklets and nodes in (b) represent tracks. The $i^{th}$ track of camera $j$, $T^i_j$, is a set of tracklets that form a clique. In (b) each clique in different colors represent tracks of the same person in non-overlapping cameras. Similar color represents the same person. (Best viewed in color)}
		\label{fig:overview}
	\end{figure*}
	
	In this work, we propose a hierarchical approach in which we first determine tracks within each camera, (Figure \ref{fig:overview}(a)) by solving data association, and later we associate  tracks of the same person in different cameras in a unified approach (Figure \ref{fig:overview}(b)), hence solving the across-camera tracking. Since appearance and motion cues of a target tend to be consistent in a short temporal window in a single camera tracking,  solving tracking problem in a hierarchical manner is common: tracklets are generated within short temporal window first and later they are merged  to form  full tracks (or trajectories) \cite{AmiShaECCV12, gmmcp, YonEyaPelPraIET2016}. Often, across-camera tracking  is more challenging than solving within-camera tracking due to the fact that appearance of people may exhibit significant differences due to illumination variations and pose changes between cameras.

	
	Therefore, this chapter proposes a unified three-layer framework to solve both within- and across-camera tracking. In the first two layers, we generate tracks within each camera and in the third layer we  associate all tracks of the same person across all cameras in a simultaneous fashion.

	To best serve our purpose, a constrained dominant sets clustering (CDSC) technique, a parametrized version of standard quadratic optimization, is employed to solve both tracking tasks. The tracking problem is caste as finding constrained dominant sets from a graph. That is, given a constraint set and a graph, CDSC  generates cluster (or clique), which forms a compact and coherent set that contains all or part of the constraint set. \textit{Clusters} represent tracklets and tracks in the first and second layers, respectively. The proposed within-camera tracker can robustly handle long-term occlusions,  does not change the scale of original problem as it does not remove nodes from the graph during the extraction  of compact clusters and is several orders of magnitude faster (close to real time) than existing methods. Also, the proposed across-camera tracking method using CDSC and later followed by refinement step offers several advantages. More specifically, CDSC not only considers the affinity (relationship) between tracks, observed in different cameras, but also takes into account the affinity  among tracks from the same camera. As a consequence, the proposed approach not only accurately associates  tracks from different cameras but also makes it possible to link multiple short broken tracks obtained during within-camera tracking, which may belong to a single target track.  
	For instance, in Figure \ref{fig:overview}(a)  track $T_1^3$ (third track from camera 1) and $T_1^4$ (fourth track from camera 1) are tracks of same person which were mistakenly broken from a single track. However, during the third layer, as they are highly similar to tracks in camera 2 ($T_2^3$) and camera 3 ($T_3^3$), they form a clique, as shown in Figure \ref{fig:overview}(b).   Such across-camera formulation is able to associate these broken tracks with the rest of tracks from different cameras, represented with the green cluster in Figure \ref{fig:overview}(b).

	The main contributions of this chapter are summarized as follows:
	\begin{itemize}
		\item We formulate multi-target tracking in  multiple non-overlapping cameras as finding constrained dominant sets from a graph. We propose a three-layer hierarchical approach, in which we first solve within-camera tracking using the first two layers, and using the third layer we solve the across-camera tracking problem.  
		\item Experiments are performed on  MOTchallenge DukeMTMCT dataset and  MARS dataset,  and show improved effectiveness of our method with respect to the state of the art.
		
	\end{itemize}
	
	The rest of the chapter is organized as follows. In Section \ref{Related Work}, we review relevant previous works. Overall proposed approach for within- and across-cameras tracking modules is summarized in section \ref{approach}, while sections \ref{within-camera tracking} and  \ref{across-camera tracking} provide more in details of the two modules. In section \ref{speedin up data association }, we present the proposed approach to further speed up our method. Experimental results are presented in Section \ref{experiments}. Finally, section \ref{sufmmary_CDSC_tracker}  summarizes the chapter.
	
	\section{Related Work}\label{Related Work}
	
	Object tracking is a challenging computer vision problem and has been one of the most active research areas for many years. In general, it can be divided in two broad categories: tracking in single and  multiple cameras. Single camera object tracking associates object detections across frames in a video sequence, so as to generate the object motion trajectory over time. Multi-camera tracking aims to solve handover problem from one camera view to another and hence establishes target correspondences among  different cameras, so as to achieve consistent object labelling across all the camera views. Early multi-camera target tracking research works fall in different categories as follows. Target tracking with partially overlapping camera views has been researched extensively during the last decade \cite{NadAndAVSS09, calderara2008bayesian, CarRauNarLuiFerACM08, SohMubPAMI03, BirThoGerICPR08, SenJasCheWayJasIVP08}. Multi target tracking across multiple cameras with disjoint views has also been researched in \cite{YinKaiTieICPR08,AmiPanSurWeiBMVC04, OmaZeeKhuMubICCV03, OmaKhuMubCVPR05, YouSenMusMTA14,RomHorICPR08, JinIsaGerWACV05, CheBhaTCS16}. Approaches for overlapping field of views  compute spatial proximity of tracks in the overlapping area, while approaches for tracking targets across cameras with disjoint fields of view, leverage appearance cues together with spatio-temporal information.
	
	
	Almost all  early multi-camera research works try to address only across-camera tracking problems, assuming that within-camera tracking results for all cameras are given. Given tracks from each camera, similarity among tracks is computed and target correspondence across cameras is solved, using the assumption that a track of a target in one camera view can match with at most one target track in another camera view.  Hungarian algorithm \cite{KuhHarNRLQ56} and  bipartite graph matching \cite{OmaZeeKhuMubICCV03} formulations are  usually used to solve this problem. Very recently, however, researchers have argued that assumptions of cameras having overlapping fields of view and the availability of intra-camera tracks  are unrealistic \cite{YouSenMusMTA14}. Therefore, the work proposed in this chapter addresses the more realistic problem by solving both within- and across-camera tracking in one joint framework. 
	
	In the rest of this section, we first review the most recent works for single camera tracking, and then describe the previous related works on multi-camera multi-view tracking.

	Single camera target tracking associates target detections across frames in a video sequence in order to generate the target motion trajectory over time.
	Zamir \etal \cite{AmiShaECCV12} formulate tracking problem as generalized maximum clique problem (GMCP), where the relationships between all detections in a temporal window are considered. In \cite{AmiShaECCV12}, a cost to each clique is assigned and  the selected clique maximizes a score function. Nonetheless, the approach  is prone to local optima as it uses greedy local neighbourhood search. Deghan \etal \cite{gmmcp} cast tracking as a generalized maximum multi clique problem  (GMMCP) and follow a joint optimization for all the tracks simultaneously. To handle outliers and weak-detections associations they introduce  dummy nodes. However, this solution is computationally expensive. In addition, the hard constraint in their optimization makes the approach impractical for large graphs. Tesfaye \etal \cite{YonEyaPelPraIET2016} consider all the pairwise relationships between detection responses in a temporal sliding window, which is used as an input to their optimization based on fully-connected edge-weighted graph. They formulate tracking as finding dominant set clusters. They follow a pill-off strategy to enumerate all possible clusters, that is, at each iteration they remove the cluster from the graph which results in a change in scale (number of nodes in a graph) of the original problem. 
	In this chapter, we propose a multiple target tracking approach, which in contrast to  previous works, does not need additional nodes to handle occlusion nor encounters change in the scale of the problem.

	Across-camera tracking aims to establish target correspondences among trajectories from different cameras so as to achieve consistent target labelling across all camera views. It is a challenging problem due to the illumination and pose changes across cameras, or track discontinuities due to the blind areas or miss detections. Existing  across-camera tracking methods try to deal with the above problems  using appearance cues. The variation in illumination of the appearance cues has been leveraged using different techniques such as Brightness Transfer Functions (BTFs). To handle the appearance change of a target as it moves from one camera to another, the authors in  \cite{OmaKhuZeeMubCVPR08} show that all brightness transfer functions from a given camera to another camera lie in a low dimensional subspace, which is learned by employing probabilistic principal component analysis and used for appearance matching. Authors of \cite{AndRicECCV06} used an incremental
	learning method to model the colour variations and \cite{BryShaTaoBMVC08} proposed a Cumulative Brightness Transfer Function, which is a better use of the available colour information from a very sparse training set. Performance comparison of different variations of Brightness Transfer Functions can be found in \cite{TizPiePaoICDSC09}. Authors in \cite{SatKaEdwTCASSP11} tried to achieve color consistency using colorimetric principles, where the image analysis system is modelled as an observer and  camera-specific transformations are determined, so that images of the same target appear similar to this observer. Obviously,  learning Brightness Transfer Functions or color correction models requires large amount of training data and they may not be robust against drastic illumination changes across different cameras. Therefore,  recent approaches have combined them with spatio-temporal cue which improve multi-target tracking performance \cite{DeYihJinQiqNanNeuro17, CheChaRamECCV10,YueRonLonAle,ShuYinAmiCVIU15,YinGerWACV14,XiaKaiTiePR14}. Chen \etal \cite{DeYihJinQiqNanNeuro17} utilized human part configurations for every target track from different cameras to describe the across-camera spatio-temporal constraints for across-camera track association, which is formulated as a multi-class classification problem via Markov Random Fields (MRF). Kuo \etal \cite{CheChaRamECCV10} used Multiple Instance Learning (MIL) to learn an appearance model, which effectively combines multiple image descriptors and their corresponding similarity measurements. The proposed appearance model combined with spatio-temporal information improved across-camera track association solving the "target handover" problem across cameras. Gao \etal \cite{YueRonLonAle} employ tracking results of different trackers and use their spatio-temporal correlation, which help them enforce tracking consistency and establish pairwise correlation among multiple tracking results. Zha \etal \cite{ShuYinAmiCVIU15} formulated tracking of multiple interacting targets as a network flow problem, for which the solution can be obtained by the K-shortest paths algorithm. Spatio-temporal relationships among targets are utilized to identify group merge and split events. In \cite{YinGerWACV14} spatio-temporal context is used  for collecting samples for discriminative appearance learning, where target-specific appearance models are learned to distinguish different people from each other. And the relative appearance context models inter-object appearance similarities for people walking in proximity and helps disambiguate individual appearance matching across cameras. 
	
	The problem of target tracking across multiple non-overlapping cameras is also tackled in \cite{RisSolZouCucTomECCV16} by extending their previous single camera tracking method \cite{ErgCarACCV2014}, where they formulate the tracking task as a graph partitioning problem. Authors in \cite{XiaKaiTiePR14}, learn across-camera transfer models including both spatio-temporal and appearance cues. While a color transfer method is used to model the changes of color across cameras for learning across-camera appearance transfer models, the spatio-temporal model is learned using an unsupervised topology recovering approach.  Recently Chen \etal \cite{CheBhaTCS16} argued that low-level information (appearance model and spatio-temporal information) is unreliable for tracking across non-overlapping cameras, and integrated contextual information such as social grouping behaviour. They formulate tracking using an online-learned Conditional Random Field (CRF), which favours track associations that maintain group consistency. In this chapter, for tracks to be associated, besides their high pairwise similarity (computed using appearance and spatio-temporal cues), their corresponding constrained dominant sets should also be similar. 
	
	Another recent popular research topic, video-based person re-identification(ReID)  \cite{JinAncXianWeiCVPR16, NiaJesPaulCVPR16, TaiShaXiaECCV14, DunCatLouLouICIAP09,LiaHuZhuLiCVPR15,FarBazPerMurCriCVPR10,XioGouCamSznECCV14,ZheSheTiaWanWanTiaICCV15,MaSuJurIVC14}, is closely related to across-camera multi-target tracking. Both problems aim to match tracks of the same persons across non-overlapping cameras. However, across-camera tracking aims at 1-1 correspondence association  between tracks of different cameras. Compared to most video-based ReID approaches, in which only  pairwise similarity between the probes and gallery is exploited, our across-camera tracking framework not only considers the relationship between probes and gallery but it also takes in to account the relationship among tracks in the gallery.
	
	\section{Overall Approach} \label{approach}

		In our formulation, in the first layer, each node in our graph represents a  short-tracklet along a temporal window (typically 15 frames). Applying constrained dominant set clustering here aim at determining cliques in this graph, which correspond to  tracklets. Likewise, each node in a graph in the second layer represents a tracklet, obtained from the first layer, and CDSC is applied here to determine cliques, which correspond to tracks. Finally, in the third layer, nodes in a graph correspond to tracks from different non-overlapping cameras, obtained from the second layer, and CDSC is applied to determine cliques, which relate  tracks of the same person across non-overlapping cameras.

	\subsection{Within-Camera Tracking} \label{within-camera tracking}
	Figure \ref{fig:within_camera_2} shows proposed within-camera tracking framework. First, we divide a video into multiple short segments, each segment contains 15 frames, and generate short-tracklets, where human detection bounding boxes in two consecutive frames with 70\%  overlap, are connected \cite{gmmcp}. Then, short-tracklets from 10 different non-overlapping segments are used as input to our first layer of tracking. Here the nodes are short-tracklets (Figure \ref{fig:within_camera_2}, bottom left). Resulting tracklets from the first layer are used as an input to the second layer, that is, a tracklet from the first layer is now represented by a node in the second layer (Figure \ref{fig:within_camera_2}, bottom right). In the second layer, tracklets of the same person from different segment are associated forming tracks of a person within a camera. 
	
	\begin{figure*}[!h]
		\centering 
		\includegraphics[width=12.5cm,height=12cm]{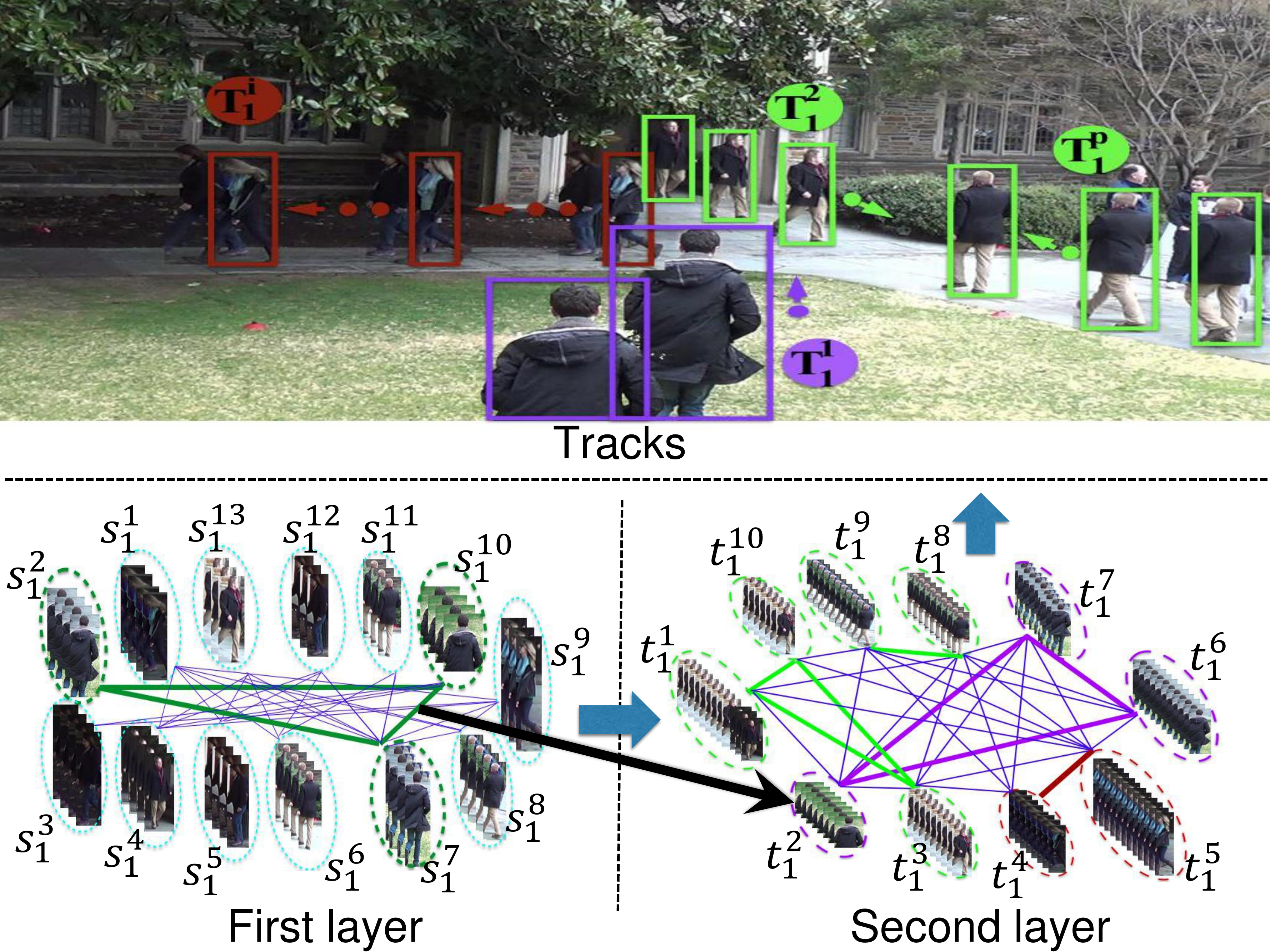}
		\caption{The figure shows within-camera tracking where short-tracklets from different  segments are used as input to our first layer of tracking. The resulting tracklets from the first layer are inputs to the second layer, which determine a tracks for each person. The three dark green short-tracklets ($s_1^2, s_1^{10}, s_1^7$), shown by dotted ellipse in the first layer, form a cluster resulting in tracklet ($t_1^2$) in the second layer, as shown with the black arrow. In the second layer, each cluster, shown in purple, green and dark red colors, form tracks of different targets, as can be seen on the top row. tracklets and tracks with the same color indicate same target. The two green cliques (with two tracklets and three tracklets) represent tracks of the person going in and out of the building (tracks $T^p_1$ and $T^2_1$ respectively) }
		
		\label{fig:within_camera_2}
	\end{figure*}
	\subsubsection{Formulation Using Constrained Dominant Sets}
	
	We build an input graph, $G(V,E,w)$,  where nodes represent short-tracklet ($s_i^j$, that is, $j^{th}$ short-tracklet of camera $i$) in the case of first layer (Figure \ref{fig:within_camera_2}, bottom left) and tracklet ($t_k^l$, that is, $l^{th}$ tracklet of camera $k$), in the second layer (Figure \ref{fig:within_camera_2}, bottom right). The corresponding affinity matrix  $\mat{A}=\left\{a_{i,j}\right\}$,  where $a_{i,j}=w(i,j)$ is built. The weight $w(i,j)$ is assigned to each edge, by considering both motion and appearance similarity between the two nodes. Fine-tuned CNN features are used to model the appearance of a node.  These features are extracted from the last fully-connected layer of Imagenet pre-trained 50-layers Residual Network (ResNet 50) \cite{KaiXiaShaJiaCVPR2016} fine-tuned using the trainval sequence of DukeMTMC dataset. Similar to \cite{AmiShaECCV12}, we employ a global constant velocity model to compute motion similarity between two nodes.
	
	
	\textbf{Determining cliques}: In our formulation, a clique of graph $G$  represents  tracklet(track) in the first  (second) layer. 
	Using short-tracklets/tracklets as a constraint set (in eq. \ref{eqn:parQP}), we enumerate all clusters, using game dynamics, by utilizing intrinsic properties of constrained dominant sets. Note that we do not use peel-off strategy to remove the nodes of found cliques from the graph, this keeps the scale of our problem (number of nodes in a graph) which guarantees that all the found local solutions are the local solutions of the (original) graph. 
	After the extraction of each cluster,  the constraint set is changed in such a way to make the extracted cluster unstable under the dynamics. The within-camera tracking starts with all nodes as constraint set. Let us say $\Gamma^i$ is the $i^{th}$ extracted cluster, $\Gamma^1$ is then the first extracted cluster which contains a subset of elements from the whole set.  After our first extraction, we change the constraint set to a set $V\backslash \Gamma^1$, hence rendering its associated nodes unstable (making the dynamics not able to select sets of nodes in the interior of associated nodes). The procedure iterates, updating the constraint set at the $i^{th}$ extraction as $V\backslash \bigcup\limits_{l=1}^i \Gamma^l$, until the constraint set becomes empty. Since we are not removing the nodes of the graph (after each extraction of a compact set), we may end up with a solution that assigns a node to more than one cluster.
	
	To find the final solution, we use the notion of centrality of constrained dominant sets. The true class of a node $j$, which is assigned to $\mat{K} > 1$ cluster, $\psi = \left\{\Gamma^1  \dots \Gamma^\mat{K} \right\}$, is  computed as:
	
	\[\argmax_{\Gamma^i \in \psi} ~ \left(|\Gamma^i|*\delta^i_j\right),\]
	where the cardinality $|\Gamma^i|$ is the number of nodes that forms the $i^{th}$ cluster and $\delta^i_j$ is the membership score of node $j$ obtained when assigned to cluster $\Gamma^i$. The normalization using the cardinality is important to avoid any unnatural bias to a smaller set.
	
	Algorithm (\ref{alg:Algorithm1_track_association}), putting the number of cameras under consideration ($\mathcal{I}$) to 1 and $\mathcal{Q}$  as short-tracklets(tracklets) in the first(second) layer, is used to determine constrained dominant sets which correspond to tracklet(track)  in the first (second) layer.
	
	%

	%
	
	\subsection{Across-Camera Tracking} \label{across-camera tracking}
	\subsubsection{Graph Representation of Tracks and the Payoff Function}
	Given tracks ($T_i^j $, that is, the $j^{th} $ track of camera $i$) of different cameras from previous step, we build graph $G'(V',E',w')$, where nodes represent tracks and their corresponding affinity matrix $\mat{A}$ depicts the similarity between tracks.  
	
	

	Assuming we have $\mathcal{I}$ number of cameras and $ \mat{A}^{i\times j}$ represents the similarity among tracks of camera $i$ and $j$, the final track based affinity $\mat{A}$, is built as
	
	$$
	\mat{A} = 
	\begin{pmatrix} 
	\mat{A}^{1 \times 1} & .  .  & \mat{A}^{1 \times j} & . . & \mat{A}^{1 \times \mathcal{I}} \\ 
	. & ~~ . ~~ &  .  & . & .\\
	~~ ~~ & ~~ ~~ & ~~ ~~ &   \\
	\mat{A}^{i \times 1} & .  .  & \mat{A}^{i \times j} & . & \mat{A}^{i \times \mathcal{I}}\\ 
	. & ~~ ~~ &  . & . & .\\
	~~ ~~ & ~~ ~~ & ~~ ~~\\
	\mat{A}^{\mathcal{I} \times 1} & .  .  & \mat{A}^{\mathcal{I} \times j} & . . & \mat{A}^{\mathcal{I} \times \mathcal{I}}\\ 
	\end{pmatrix}.
	$$ 
	
	Figure \ref{fig:ExamplarGraphWithinAndAcross} shows exemplar graph for  across-camera tracking among three cameras.  $T^i_j$ represents the $i^{th}$ track of camera $j$. Black and orange edges, respectively, represent within- and across-camera relations of the tracks. From the affinity $\mat{A}$, $\mat{A}^{i\times j}$ represents the black edges of camera $i$ if $i = j$, which otherwise represents the across-camera relations using the orange edges. 
	
	The colors of the nodes depict the track ID; nodes with similar color represent tracks of the same person. Due to several reasons such as long occlusions, severe pose change of a person, reappearance and others, a person may have more than one track (a \textit{broken track}) within a camera. The green nodes of camera 1 (the second and the $p^{th}$ tracks) typify two \textit{broken tracks} of the same person, due to reappearance as shown in Figure \ref{fig:within_camera_2}. The proposed unified approach, as discussed in the next section, is able to deal with such cases. 
	
	\begin{figure}[!h]
		\centering
		\includegraphics[width=1\linewidth]{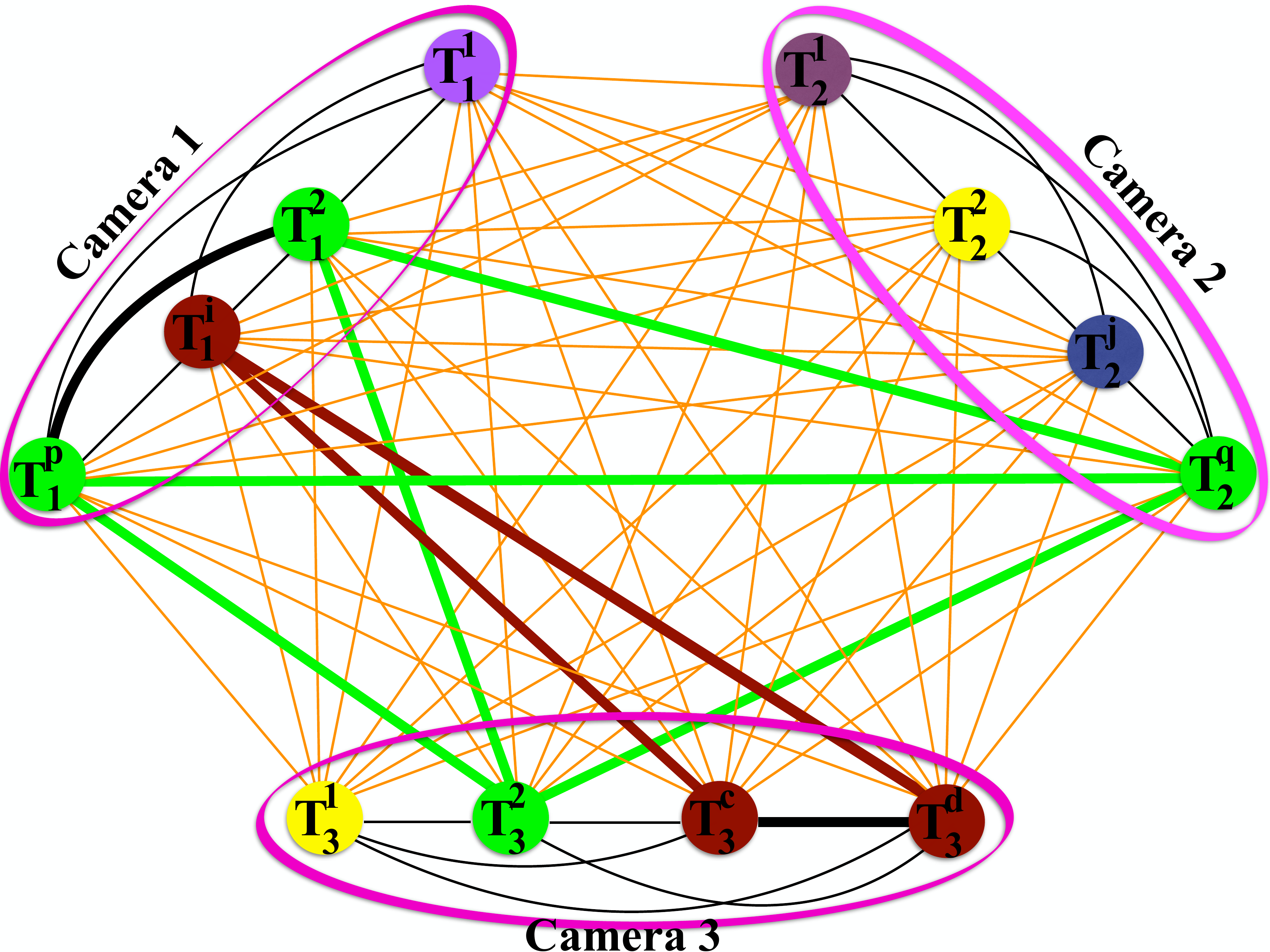}
		\caption{\small Exemplar graph of tracks from three cameras. $T^i_j$ represents the $i^{th}$ track of camera $j$.  Black and colored edges, respectively, represent within- and across-camera relations of tracks. Colours of the nodes depict track IDs, nodes with similar colour represent tracks of the same person, and the thick lines show both within- and across-camera association. 
		}
		\label{fig:ExamplarGraphWithinAndAcross}
	\end{figure}

	\subsubsection{Across-camera Track Association}

	In this section, we discuss how we simultaneously solve within- and across-camera tracking.
	Our framework is  naturally able to deal with the  errors listed above. 
	A person, represented by the green node from our exemplar graph (Figure \ref{fig:ExamplarGraphWithinAndAcross}), has two tracks which are difficult to merge during within-camera tracking; however, they belong to clique (or cluster) with tracks in camera 2 and camera 3, since they  are highly similar. The algorithm applied to a such across-camera graph is able to cluster all the correct tracks. 
	This helps us linking \textit{broken tracks} of the same person occurring during within-camera track generation stage. 
	
	Using the graph with nodes of tracks from a camera as a constraint set, data association for both within- and across-camera are performed simultaneously. Let us assume, in our exemplar graph (Figure \ref{fig:ExamplarGraphWithinAndAcross}), our constraint set $\mathcal{Q}$ contains nodes of tracks of camera 1, $\mathcal{Q}$ = \{ $T^1_1, T^2_1, T^i_1, T^p_1$ \}.  $I_\mathcal{Q}$ is then $n \times n$ diagonal matrix, whose diagonal elements are  set to 1 in correspondence to the vertices contained in all cameras, except camera 1 which takes the value zero. That is,  the sub-matrix $I_\mathcal{Q}$, that corresponds to $A^{1\times 1}$, will be a zero matrix of size equal to number of tracks of the corresponding camera. Setting $\mathcal{Q}$ as above, we have guarantee that the maximizer of program in eq. (\ref{eqn:parQP}) contains some elements from set $\mathcal{Q}$: i.e., $\mathcal{C}_1^1$=$\left\{T^2_1, T^p_1, T^q_2, T^2_3\right\}$ forms a clique which contains set $\left\{T^2_1, T^p_1\right\} \in \mathcal {Q}$. 
	This is  shown in Figure \ref{fig:ExamplarGraphWithinAndAcross}, using the thick green edges (which illustrate across-camera track association) and the thick black edge (which typifies the within camera track association). The second set, $\mathcal{C}_1^2$, contains tracks shown with the dark red color, which illustrates the case where within- and across-camera tracks are in one clique. Lastly, the $\mathcal{C}_1^3$ = $T^1_1$ represents a track of a person that appears only in camera 1. 
	As a general case, $C^i_j$, represents the $i^{th}$ track set using tracks in camera $j$ as a constraint set and $C_j$ is the set that contains track sets generated using camera $j$ as a constraint set, e.g. $C_1 = \left\{ C^1_1, C^2_1, C^3_1\right\}$. We iteratively  process all the cameras and then apply track refinement step.
	
	Though Algorithm (\ref{alg:Algorithm1_track_association}) is applicable to within-camera tracking also, here we show the specific case for across-camera track association.
	Let $\mathcal{T}$ represents the set of tracks from all the cameras we have and $C$ is the set which contains sets of tracks, as $C^i_p$, generated using our  algorithm. $T^{\vartheta}_p$ typifies the $\vartheta^{th}$ track from camera $p$ and $T_p$ contains all the tracks in camera $p$. The function $\mathcal{F}(\mathcal{Q},\mat{A} $) takes as an input a constraint set $\mathcal{Q}$ and the affinity $\mat{A}$, and provides as output all the $m$ local solutions $\mathcal{X}^{n\times m}$ of program \eqref{eqn:parQP} that contain element(s) from the constraint set. This can be accomplished by iteratively finding a local maximizer of equation (program) (\ref{eqn:parQP}) in $\Delta$, e.g. using game dynamics, and then changing the constraint set $\mathcal{Q}$, until all members of the constraint set have been clustered.
	
\begin{algorithm}[t]
	\caption{Track Association}	
	\label{alg:Algorithm1_track_association}	
	{\bf INPUT:} Affinity $\mat{A}$, Sets of tracks $\mathcal{T}$ from $\mathcal{I}$ cameras\\
	$C$ $\leftarrow \emptyset$ Initialize the set with empty-set \\
	Initialize $\x$ to the barycenter and $i$ and $p$ to 1
	
	\rule{1\linewidth}{0.02cm}
	\begin{algorithmic}[1]
		\While {$p\le \mathcal{I}$} \\
		$\mathcal{Q}$ $\leftarrow$ $T_p$, define constraint set\\
		$\mathcal{X}$ $\leftarrow$ $\mathcal{F}(\mathcal{Q},\mat{A} $)\\
		$C^i_p$ = $\leftarrow \sigma(\mathcal{X}^i)$, compute for all $i = 1 \dots m$\\
		$p$ $\leftarrow$ $p+1$
		\EndWhile \\	    
		$C$ = $\bigcup\limits_{p=1}^\mathcal{I} C_p$\\
		{\bf OUTPUT:} \{$C$\}
		
	\end{algorithmic}
\end{algorithm}

	
%
	
	\subsection{Track Refinement}
	The proposed framework, together with the notion of centrality of constrained dominant sets and the notion of reciprocal neighbours, helps us in refining tracking results  using tracks from different cameras as different constraint sets. Let us assume we have $\mathcal{I}$ cameras and $\mathcal{K}^i$ represents the set corresponding to track $i$,  while $\mathcal{K}^i_p$ is the subset of $\mathcal{K}^i$ that corresponds to the $p^{th}$ camera.  $\mathcal{M}^{l^i}_p$ is  the membership score assigned to the $l^{th}$ track in the set $\mathcal{C}^i_p$. 
	
	We use two constraints during 	track refinement stage, which  helps us refining false positive association.
	
	\noindent {\bf Constraint-1:}  {\em A track can not be found in two different sets generated using same constraint set}, i.e. it must hold that:
	
	\[|\mathcal{K}_p^i| \le 1 \] 
	
	Sets that do not satisfy the above inequality should be refined as there is one or more tracks that exist in different sets of tracks collected using the same constraint, i.e. $T_p$. The corresponding track is removed  from all the sets which contain it and is assigned to the right set based on its membership score in each of the sets. Let us say the $l^{th}$ track exists in $q$ different sets, when tracks from camera $p$ are taken as a constraint set, $|\mathcal{K}^l_p|=q$. The right set which contains the track, $C^r_p$, is chosen as:  
	
	\[C^r_p = \argmax_{C^i_p \in \mathcal{K}^l_p} ~ \left(|C^i_p|*\mathcal{M}^{l^i}_p \right). \]
	where $i = 1, \dots, |\mathcal{K}^l_p| $.
	This must be normalized with the cardinality of the set to avoid a bias towards smaller sets.
	
	\noindent {\bf Constraint-2:} {\em The maximum number of sets that contain track $i$ should be the number of cameras under consideration}. If we consider $\mathcal{I}$ cameras, the cardinality of the set which contains sets with track $i$, is not larger than $\mathcal{I}$, i.e.: 
	
	\[|\mathcal{K}^i| \le \mathcal{I}. \] 
	
	If there are sets that do not satisfy the above condition, the tracks are refined based on the cardinality of the intersection of sets that contain the track, i.e. by enforcing the reciprocal properties of the sets.
	
	If there are sets that do not satisfy the above condition, the tracks are refined based on the cardinality of the intersection of sets that contain the track by enforcing the reciprocal properties of the sets which contain a track. 
	Assume we collect sets of tracks considering tracks from camera $q$ as constraint set and assume a track $\vartheta$ in the set $C^j_p$, $p\ne q$, exists in more than one sets of $C_q$.  The right set, $C^r_q$, for $\vartheta$ considering tracks from camera $q$ as constraint set is chosen as:
	
	\[C^r_q = \argmax_{C^i_q \in \mathcal{K}^\vartheta_q} ~ \left( C^i_q \cap C^j_p \right).\] where $i = 1, \dots, |\mathcal{K}^\vartheta_q| $.

	\section{Experimental Results} \label{experiments}
	The proposed framework has been evaluated on recently-released large dataset, MOTchallenge DukeMTMC \cite{RisSolZouCucTomECCV16,solera2016groups,ErgCarACCV2014}. Even though the main focus of this chapter is on multi-target tracking in multiple non-overlapping cameras, we also perform additional experiments on MARS  \cite{ZheBieSunWanSuWanTiaECCV16}, one of the largest and challenging video-based person re-identification dataset, to show that the proposed across-camera tracking approach can efficiently solve this task also.  
	
	\textbf{DukeMTMC}  is recently-released dataset to evaluate the performance of multi-target multi-camera tracking systems. It is the largest (to date),  fully-annotated and calibrated high resolution 1080p, 60fps dataset, that covers a single outdoor scene from 8 fixed synchronized cameras, the topology of cameras is shown in Fig. 4. The dataset consists of 8 videos of 85 minutes each from the 8 cameras, with 2,700 unique identities (IDs) in more than 2 millions frames in each video  containing 0 to 54 people. The video is split in three parts: (1) Trainval (first 50 minutes of the video), which is for training and validation; (2) Test-Hard (next 10 minutes after Trainval sequence); and (3) Test-Easy, which covers the last 25 minutes of the video. Some of the properties which make the dataset more challenging include: huge amount of data to process, it contains 4,159 hand-overs, there are more than 1,800 self-occlusions (with 50\% or more overlap),  891 people walking in front of only one camera.
	\begin{figure}[h]
		\centering 
		\includegraphics[width=1\linewidth, scale = 0.4 ,trim=0cm 0cm 0cm 0cm,clip]{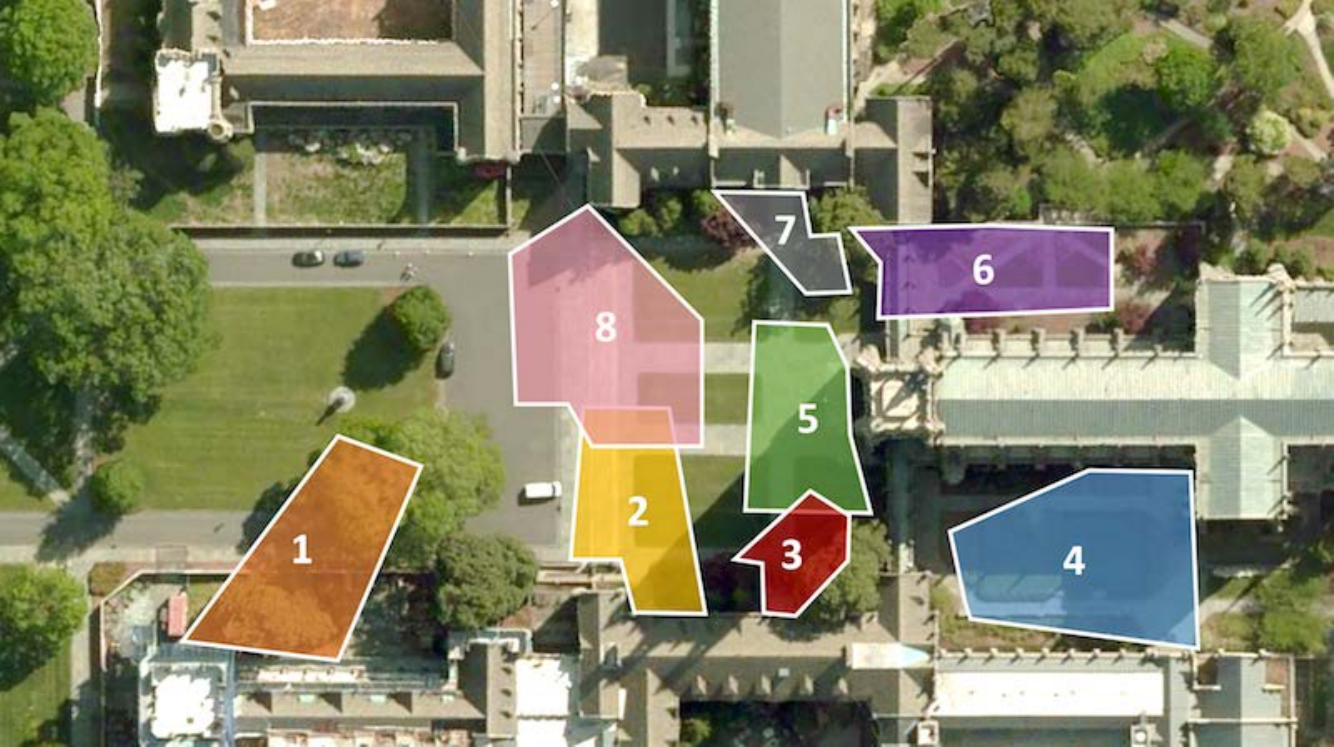}
		\caption{ Camera topology for DukeMTMC dataset. Detections from the overlapping fields of view are not considered. More specifically, intersection occurred between camera (8 \& 2) and camera (5 \& 3). }
		\label{fig:Camera_topology}
	\end{figure}
	
	\textbf{MARS (Motion Analysis and Re-identification Set)}  is an extension of the Market-1501 dataset \cite{ZheBieSunWanSuWanTiaECCV16}. It has been collected from six near-synchronized cameras. It consists of 1,261 different pedestrians, who are captured by at least 2 cameras. The variations in poses, colors and illuminations of pedestrians, as well as the poor image quality, make it very  difficult to yield high matching accuracy. Moreover, the dataset contains 3,248 distractors in order to make it more realistic. Deformable Part Model (DPM) \cite{Felzenszwalb_DPBM} and GMMCP tracker \cite{gmmcp} were used to automatically generate the tracklets (mostly 25-50 frames long). Since the video and the detections are not available we use the generated tracklets as an input to our framework.

	\textbf{Performance Measures:} In addition to the standard Multi-Target Multi-Camera tracking performance measures, we evaluate our framework using additional measures recently proposed in \cite{RisSolZouCucTomECCV16}: Identification F-measure (IDF1), Identification Precision (IDP) and Identification Recall (IDR) \cite{RisSolZouCucTomECCV16}. The standard performance measures such as CLEAR MOT report the amount of  incorrect decisions made by a tracker. Ristani \etal \cite{RisSolZouCucTomECCV16} argue and demonstrate that some system users may instead be more interested in how well they can determine who is where at all times. After pointing out that different measures serve different purposes, they proposed the three measures (IDF1, IDP and IDR) which can be applied both within- and across-cameras. These measure tracker's performance not by how often ID switches
	occur, but by how long the tracker correctly tracks targets.
	
	\textbf{\textit{Identification precision IDP (recall IDR)}:} is the fraction of computed (ground truth) detections that are correctly identified. 
	
	\textbf{\textit{Identification F-Score IDF1}:} is the ratio of correctly identified detections over the average number of ground-truth and computed detections. Since MOTA and its related performance measures under-report across-camera errors \cite{RisSolZouCucTomECCV16}, we use them for the evaluation of our single camera tracking results.
	
	The performance of the algorithm for re-identification is evaluated employing rank-1 based accuracy  and confusion matrix using average precision (AP).

	\textbf{Implementation:} In the implementation of our framework, we do not have parameters to tune. The affinity matrix $\mat{A}$ adapting kernel trick distance function from \cite{GroBayBenACM05}, is constructed as follows:
	
	\[ \mat{A}_{i,j} = 1 - \sqrt{\frac{\mat{K}(x_i,x_i)+\mat{K}(x_j,x_j) - 2*\mat{K}(x_i,x_j)} {2}},\]
	where $\mat{K}(x_i,x_j)$ is chosen as the Laplacian kernel 
	
	\[ exp(-\gamma\parallel x_i - x_j \parallel _1).\]
	The kernel parameter $\gamma$ is set as the inverse of the median of pairwise distances.
	
	In our similarity matrix for the final layer of the framework, which is sparse, we use spatio-temporal information based on the time duration and the zone of a person moving from one zone of a camera to other zone of another camera which is learned from the Trainval sequnece of DukeMTMC dataset. The affinity between track $i$ and track $j$ is different from zero , if and only if they have a possibility, based on the direction a person is moving and the spatio-temporal information, to be linked and form a trajectory (across camera tracks of a person). However, this may have a drawback due to \textit{broken tracks} or track of a person who is standing and talking or doing other things in one camera which results in a track that does not meet the spatio-temporal constraints. To deal with this problem, we add, for the across camera track's similarity, a path-based information as used in \cite{EyaMarICIAP2015}, i.e if a track in camera $i$ and a track in camera $j$ have a probability to form a trajectory,  and track $j$ in turn have linkage possibility with a track in camera $z$, the tracks in camera $i$ and camera $z$ are considered to have a possibility to be linked.
	
	The similarity between two tracks is computed using the Euclidean distance of the max-pooled features. The max-pooled features  are computed as the row maximum of the feature vector of individual patch, of the given track, extracted from the last fully-connected layer of Imagenet pre-trained 50-layers Residual Network (ResNet\_50) \cite{KaiXiaShaJiaCVPR2016}, fine-tuned using the Trainval sequence of DukeMTMC dataset. The network is fine-tuned  with classification loss on the Trainval sequence, and activations of its last fully-connected layer are extracted, L2-normalized and taken as visual features. Cross-view Quadratic Discriminant Analysis (XQDA) \cite{LiaHuZhuLiCVPR15} is then used for pairwise distance computation between instances. For the experiments on MARS, patch representation is obtained using CNN features used in \cite{ZheBieSunWanSuWanTiaECCV16}. The pairwise distances between instances are then computed in XQDA, KISSME \cite{KosHirWohRotBisCVPR12} and euclidean spaces.
	

	\begin{small}
		\begin{table*}[t]
			\centering
			\scalebox{0.915}{		
				\begin{tabular}{cc|c|c|c|c|c|c|c|c|r}
					\centering
					& {Mthd} &{MOTA$\uparrow$} &{MT$\uparrow$} & {ML$\downarrow$}  &{FP$\downarrow$} & {FN$\downarrow$} &{IDS$\downarrow$}  & {IDF1}$\uparrow$ & IDP$\uparrow$ &  IDR$\uparrow$ \\ \cline{1-10}
					\cline{1-11}
					\multicolumn{1}{ c  }{\multirow{2}{*}{C1} } &
					\multicolumn{1}{ |c| }{\cite{RisSolZouCucTomECCV16}} & 43.0	& 24&46&2,713&107,178&39 &57.3&91.2&41.8   \\ \cline{2-11}
					\multicolumn{1}{ c  }{}                        &
					\multicolumn{1}{ |c| }{Ours} &  69.9  & 137& 22 & 5,809 & 52,152 & 156 & 76.9 & 89.1 & 67.7 \\ \cline{1-11}
					\multicolumn{1}{ c  }{\multirow{2}{*}{C2} } &
					\multicolumn{1}{ |c| }{\cite{RisSolZouCucTomECCV16}} & 44.8& 133 & 8 & 47,919 & 53,74 & 60 & 68.2 &	69.3 &67.1   \\ \cline{2-11}
					\multicolumn{1}{ c  }{}                        &
					\multicolumn{1}{ |c| }{Ours} & 71.5 & 134 & 21 & 8,487 & 43,912 & 75 & 81.2 & 90.9 & 73.4  \\ \cline{1-11}
					\multicolumn{1}{ c  }{\multirow{2}{*}{C3} } &
					\multicolumn{1}{ |c| }{\cite{RisSolZouCucTomECCV16}} & 	57.8 &	52 &	22 &	1,438 & 28,692 &	16 & 60.3&	78.9&	48.8   \\ \cline{2-11}
					\multicolumn{1}{ c  }{}                        &
					\multicolumn{1}{ |c| }{Ours} & 67.4 & 44 & 9 & 2,148 & 21,125 & 38 & 64.6 & 76.3 & 56.0 \\ \cline{1-11}
					\multicolumn{1}{ c  }{\multirow{2}{*}{C4} } &
					\multicolumn{1}{ |c| }{\cite{RisSolZouCucTomECCV16}} & 63.2&	36&	18&	2,209&	19,323&	7 & 	73.5&	88.7&	62.8   \\ \cline{2-11}
					\multicolumn{1}{ c  }{}                        &
					\multicolumn{1}{ |c| }{Ours} & 76.8  & 45 & 4 & 2,860 & 10,689 & 18 & 84.7 & 91.2 & 79.0   \\ \cline{1-11}
					
					\multicolumn{1}{ c  }{\multirow{2}{*}{C5} } &
					\multicolumn{1}{ |c| }{\cite{RisSolZouCucTomECCV16}} & 72.8 &	107&17&	4,464&35,861&	54&73.2&	83.0&	65.4    \\ \cline{2-11}
					\multicolumn{1}{ c  }{}                        &
					\multicolumn{1}{ |c| }{Ours} & 68.9 & 88 & 11 & 9,117 & 36,933 & 139 & 68.3& 76.1 & 61.9   \\ \cline{1-11}
					\multicolumn{1}{ c  }{\multirow{2}{*}{C6} } &
					\multicolumn{1}{ |c| }{\cite{RisSolZouCucTomECCV16}} & 73.4&	142&27&	5,279&45,170& 55&	77.2&	87.5&	69.1   \\ \cline{2-11}
					\multicolumn{1}{ c  }{}                        &
					\multicolumn{1}{ |c| }{Ours} & 77.0 & 136 & 11 & 4,868 & 38,611 & 142 & 82.7 & 91.6 &75.3  \\ \cline{1-11}
					\multicolumn{1}{ c  }{\multirow{2}{*}{C7} } &
					\multicolumn{1}{ |c| }{\cite{RisSolZouCucTomECCV16}} &	71.4&	69& 13&1,395&18,904&	23 & 80.5&	93.6&	70.6    \\ \cline{2-11}
					\multicolumn{1}{ c  }{}                        &
					\multicolumn{1}{ |c| }{Ours} & 73.8 & 64 & 4 & 1,182 & 17,411 & 36 & 81.8 & 94.0& 72.5    \\ \cline{1-11}
					\multicolumn{1}{ c  }{\multirow{2}{*}{C8} } &
					\multicolumn{1}{ |c| }{\cite{RisSolZouCucTomECCV16}} & 60.7&	102&53&2,730&52,806&	46&	 72.4&	92.2&	59.6     \\ \cline{2-11}
					\multicolumn{1}{ c  }{}                        &
					\multicolumn{1}{ |c| }{Ours} & 63.4  & 92 & 28 & 4,184 & 47,565 & 91 & 73.0& 89.1 & 61.0  \\ \cline{1-11}
					\multicolumn{1}{ c  }{\multirow{2}{*}{Av} } &
					\multicolumn{1}{ |c| }{\cite{RisSolZouCucTomECCV16}} & 59.4& 665&234&68,147&361,672&\textbf{300}& 70.1 &83.6&	60.4    \\ \cline{2-11}
					\multicolumn{1}{ c  }{}                        &
					\multicolumn{1}{ |c| }{Ours} &  \textbf{70.9} & \textbf{740} &\textbf{110} & \textbf{38,655} &  \textbf{268,398} & 693 & \textbf{77.0} & \textbf{87.6} &   \textbf{68.6}\\ \cline{1-11}
					
				\end{tabular}}
				
				\caption{The results show detailed (for each camera C1 to C8) and average performance (Av) of our and state-of-the-art approach \cite{RisSolZouCucTomECCV16} on the Test-Easy sequence of DukeMTMC dataset.}
				\label{table:TestEasyDuke}
			\end{table*}	
		\end{small}
		
		\begin{small}
			\begin{table*}[t]
				\centering
				\scalebox{0.915}{	
					\begin{tabular}{cc|c|c|c|c|c|c|c|c|r}
						\centering
						& {Mthd} &{MOTA$\uparrow$} &{MT$\uparrow$} & {ML$\downarrow$}  &{FP$\downarrow$} & {FN$\downarrow$} &{IDS$\downarrow$}  & {IDF1}$\uparrow$ & IDP$\uparrow$ &  IDR$\uparrow$ \\ \cline{1-10}
						\cline{1-11}
						\multicolumn{1}{ c  }{\multirow{2}{*}{C1} } &
						\multicolumn{1}{ |c| }{\cite{RisSolZouCucTomECCV16}} &  37.8& 6	  & 34  & 1,257	      & 78,977	& 55	           & 52.7     & 92.5	  & 36.8     \\ \cline{2-11}
						\multicolumn{1}{ c  }{}                        &
						\multicolumn{1}{ |c| }{Ours} &   63.2 & 65 & 17 & 2,886 & 44,253 & 408 & 67.1 & 83.0 & 56.4 \\ \cline{1-11}
						\multicolumn{1}{ c  }{\multirow{2}{*}{C2} } &
						\multicolumn{1}{ |c| }{\cite{RisSolZouCucTomECCV16}} &  47.3	& 68  & 12   & 26526    & 46898	& 194        & 60.6	    & 65.7	  & 56.1      \\ \cline{2-11}
						\multicolumn{1}{ c  }{}                        &
						\multicolumn{1}{ |c| }{Ours} & 54.8 & 62 & 16 & 8,653 & 54,252 & 323 & 63.4 & 78.8 & 53.1 \\ \cline{1-11}
						\multicolumn{1}{ c  }{\multirow{2}{*}{C3} } &
						\multicolumn{1}{ |c| }{\cite{RisSolZouCucTomECCV16}} &  46.7	& 24	  & 4	    & 288	      & 18182	& 6	             & 62.7	    & 96.1	  & 46.5	  \\ \cline{2-11}
						\multicolumn{1}{ c  }{}                        &
						\multicolumn{1}{ |c| }{Ours} & 68.8  & 18 & 2 & 2,093 & 8,701 & 11 & 81.5 & 91.1 & 73.7 \\ \cline{1-11}
						\multicolumn{1}{ c  }{\multirow{2}{*}{C4} } &
						\multicolumn{1}{ |c| }{\cite{RisSolZouCucTomECCV16}} &  85.3	& 21	  & 0	    & 1,215	      & 2,073	& 1	              & 84.3   & 86.0	  & 82.7    \\ \cline{2-11}
						\multicolumn{1}{ c  }{}                        &
						\multicolumn{1}{ |c| }{Ours} & 75.6 & 17 & 0 & 1,571 & 3,888 & 61 & 82.3 & 87.1 & 78.1 \\ \cline{1-11}
						
						\multicolumn{1}{ c  }{\multirow{2}{*}{C5} } &
						\multicolumn{1}{ |c| }{\cite{RisSolZouCucTomECCV16}} &  78.3	& 57	  & 2	    & 1,480	      & 11,568	& 13	             & 81.9	    & 90.1	  & 75.1    \\ \cline{2-11}
						\multicolumn{1}{ c  }{}                        &
						\multicolumn{1}{ |c| }{Ours} & 78.6  & 47 & 2 & 1,219 & 11,644 & 50& 82.8 & 91.5 & 75.7   \\ \cline{1-11}
						\multicolumn{1}{ c  }{\multirow{2}{*}{C6} } &
						\multicolumn{1}{ |c| }{\cite{RisSolZouCucTomECCV16}} &  59.4	& 85	  & 23 & 5,156	      & 77,031	& 225      & 64.1	    & 81.7	  & 52.7      \\ \cline{2-11}
						\multicolumn{1}{ c  }{}                        &
						\multicolumn{1}{ |c| }{Ours} & 53.3 & 68 & 36 & 5,989 & 88,164 & 547 & 53.1 & 71.2 & 42.3  \\ \cline{1-11}
						\multicolumn{1}{ c  }{\multirow{2}{*}{C7} } &
						\multicolumn{1}{ |c| }{\cite{RisSolZouCucTomECCV16}} &  50.8	& 43	  & 23  & 2,971	      & 38,912	& 148       & 59.6	    & 81.2	  & 47.1     \\ \cline{2-11}
						\multicolumn{1}{ c  }{}                        &
						\multicolumn{1}{ |c| }{Ours} & 50.8  & 34 & 20 & 1,935 & 39,865 & 266 & 60.6 & 84.7 & 47.1   \\ \cline{1-11}
						\multicolumn{1}{ c  }{\multirow{2}{*}{C8} } &
						\multicolumn{1}{ |c| }{\cite{RisSolZouCucTomECCV16}} &  73.0	& 34	  & 5	     & 706	      & 9735	& 10	            & 82.4   &94.9	  & 72.8     \\ \cline{2-11}
						\multicolumn{1}{ c  }{}                        &
						\multicolumn{1}{ |c| }{Ours} & 70.0  & 37 & 6 & 2,297 & 9,306 & 26 & 81.3 & 90.3 & 73.9   \\ \cline{1-11}
						\multicolumn{1}{ c  }{\multirow{2}{*}{Av} } &
						\multicolumn{1}{ |c| }{\cite{RisSolZouCucTomECCV16}} & 54.6	& 338	&103	& 39,599	& 283,376	& \textbf{652}	   &64.5	& 81.2	& 53.5     \\ \cline{2-11}
						\multicolumn{1}{ c  }{}                        &
						\multicolumn{1}{ |c| }{Ours} & \textbf{59.6} & \textbf{348} & \textbf{99}& \textbf{26,643} & \textbf{260,073} & 1637 & \textbf{65.4} & \textbf{81.4}& \textbf{54.7}  \\ \cline{1-11}
						
					\end{tabular}}
					
					\caption{The results show detailed (for each camera) and average performance of our and state-of-the-art approach \cite{RisSolZouCucTomECCV16} on the Test-Hard sequence of DukeMTMC dataset.}
					\label{table:TestHardDuke}
				\end{table*}
			\end{small}

			\begin{table}[h]
				\centering
				\begin{tabular}{cc|c|c|r}
					
					& Methods  &{IDF1}$\uparrow$ & IDP$\uparrow$ &  IDR$\uparrow$ \\ \cline{3-4}
					\cline{1-5}
					\multicolumn{1}{ c  }{\multirow{2}{*}{Multi-Camera} } &
					\multicolumn{1}{ |c| }{\cite{RisSolZouCucTomECCV16}} & 
					56.2 &	67.0 &	48.4 \\ \cline{2-5}
					\multicolumn{1}{ c  }{}                        &
					\multicolumn{1}{ |c| }{Ours} & \textbf{60.0} & \textbf{68.3} &  \textbf{53.5}   \\ \cline{1-5}
					
				\end{tabular}
				
				\caption{Multi-camera performance of our and state-of-the-art approach \cite{RisSolZouCucTomECCV16} on the Test-Easy sequence of DukeMTMC dataset.}
				\label{table:MCTestEasyDuke}
			\end{table}

			\begin{table}[h]	
				\centering
				\begin{tabular}{cc|c|c|r}
					
					& Methods  &{IDF1}$\uparrow$ & IDP$\uparrow$ &  IDR$\uparrow$ \\ \cline{3-4}
					\cline{1-5}
					\multicolumn{1}{ c  }{\multirow{2}{*}{Multi-Camera} } &
					\multicolumn{1}{ |c| }{\cite{RisSolZouCucTomECCV16}} & 47.3	& 59.6 &39.2
					\\ \cline{2-5}
					\multicolumn{1}{ c  }{}                        &
					\multicolumn{1}{ |c| }{Ours} & \textbf{50.9} & \textbf{63.2} &   \textbf{42.6}  \\ \cline{1-5}
					
				\end{tabular}
				
				\caption{Multi-Camera performance of our and state-of-the-art approach \cite{RisSolZouCucTomECCV16} on the Test-Hard sequence of DukeMTMC dataset.}
				\label{table:MCTestHardDuke}
			\end{table}
	\subsection{Evaluation on DukeMTMC dataset:} In Table \ref{table:TestEasyDuke} and Table \ref{table:TestHardDuke}, we compare quantitative performance of our method with state-of-the-art multi-camera multi-target tracking method on the DukeMTMC dataset. The symbol $\uparrow$ means higher scores indicate better performance, while $\downarrow$ means lower scores indicate better performance. The quantitative results of the trackers shown in table \ref{table:TestEasyDuke} represent the performance on the  Test-Easy sequence, while those in table \ref{table:TestHardDuke} show the performance on the  Test-Hard sequence. For a fair comparison, we use the same detection responses obtained from MOTchallenge DukeMTMC as the input to our method. In both cases, the reported results of row 'Camera 1' to 'Camera 8' represent the within-camera tracking performances. The last row of the tables represent the average performance over 8 cameras. Both tabular results demonstrate that the proposed approach improves tracking performance for both sequences. In the Test-Easy sequence, the performance is improved by 11.5\% in MOTA and 7\% in IDF1 metrics, while in that of the Test-Hard sequence, our method produces 5\% larger average MOTA score than \cite{RisSolZouCucTomECCV16}, and 1\% improvement is achieved in IDF1. Table \ref{table:MCTestEasyDuke} and Table \ref{table:MCTestHardDuke} respectively present Multi-Camera performance of our and state-of-the-art approach \cite{RisSolZouCucTomECCV16} on the Test-Easy and  Test-Hard sequence (respectively) of DukeMTMC dataset. We have improved IDF1 for both Test-Easy and Test-Hard sequences by 4\% and 3\%, respectively.
	
	Figure \ref{fig:QualitativeResults} depicts sample qualitative results. Each person is represented by (similar color of) two bounding boxes, which represent the person's position at some specific time, and a track which shows the path s(he) follows. In the first row, all the four targets, even under significant illumination and pose changes, are successfully tracked in four cameras, where they appear. In the second row, target 714 is successfully tracked through three cameras. Observe its significant illumination and pose changes from camera 5 to camera 7. In the third row, targets that move through camera 1, target six, seven and eight are tracked. The last row shows tracks of targets that appear in cameras 1 to 4.
		
		\begin{figure*}[h!]
			\centering
			\includegraphics[width=12.5cm,height=12cm]{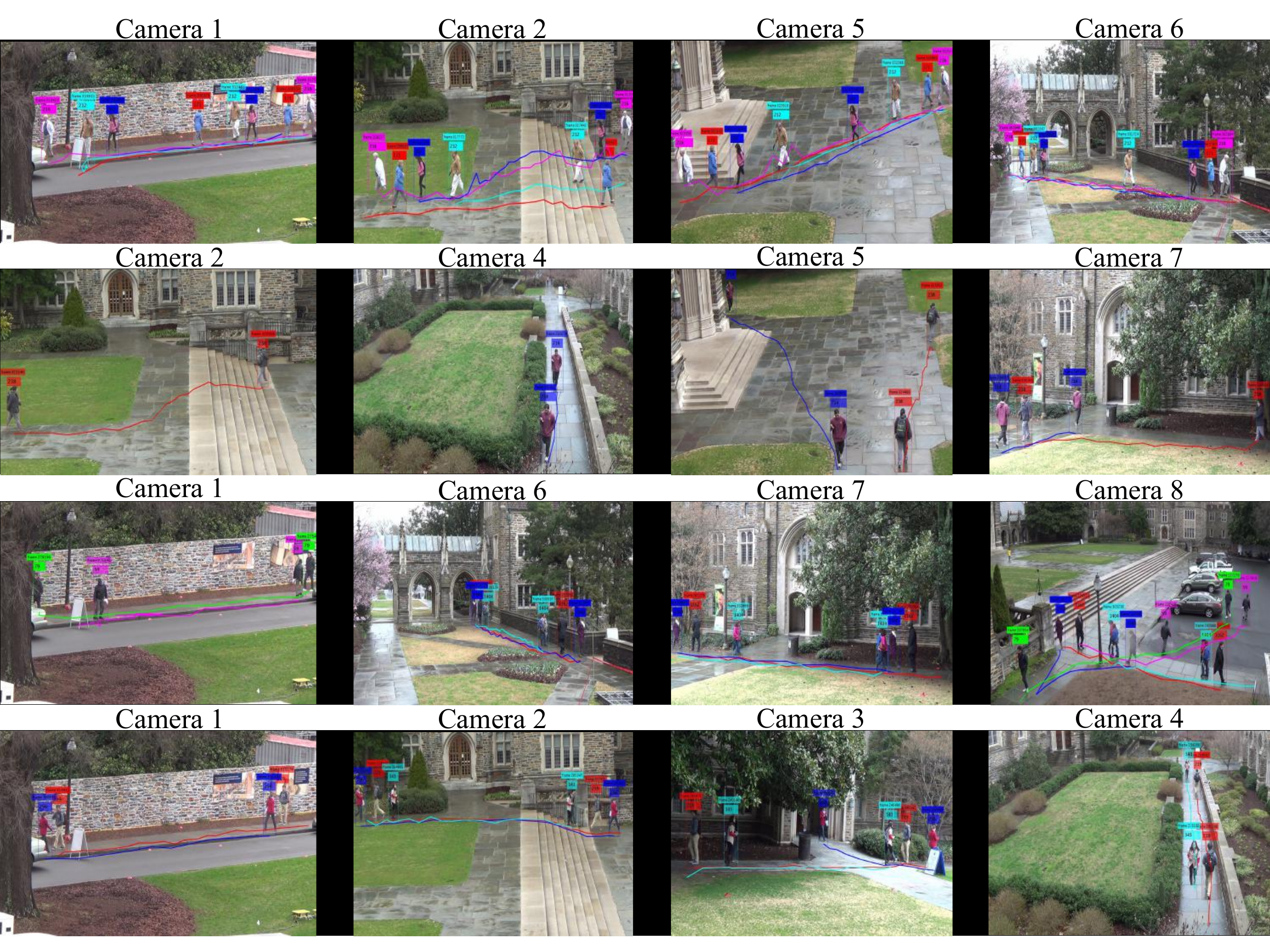}
			\caption{\small Sample qualitative results of the proposed approach on DukeMTMC dataset. Bounding boxes and lines with the same color indicate the same target (Best viewed in color).
			}
			\label{fig:QualitativeResults}
		\end{figure*}

	
	\begin{figure*}[h]
		\centering 
		\includegraphics[width=12.5cm,height=8cm ,trim=0cm 0cm 0cm 0cm,clip]{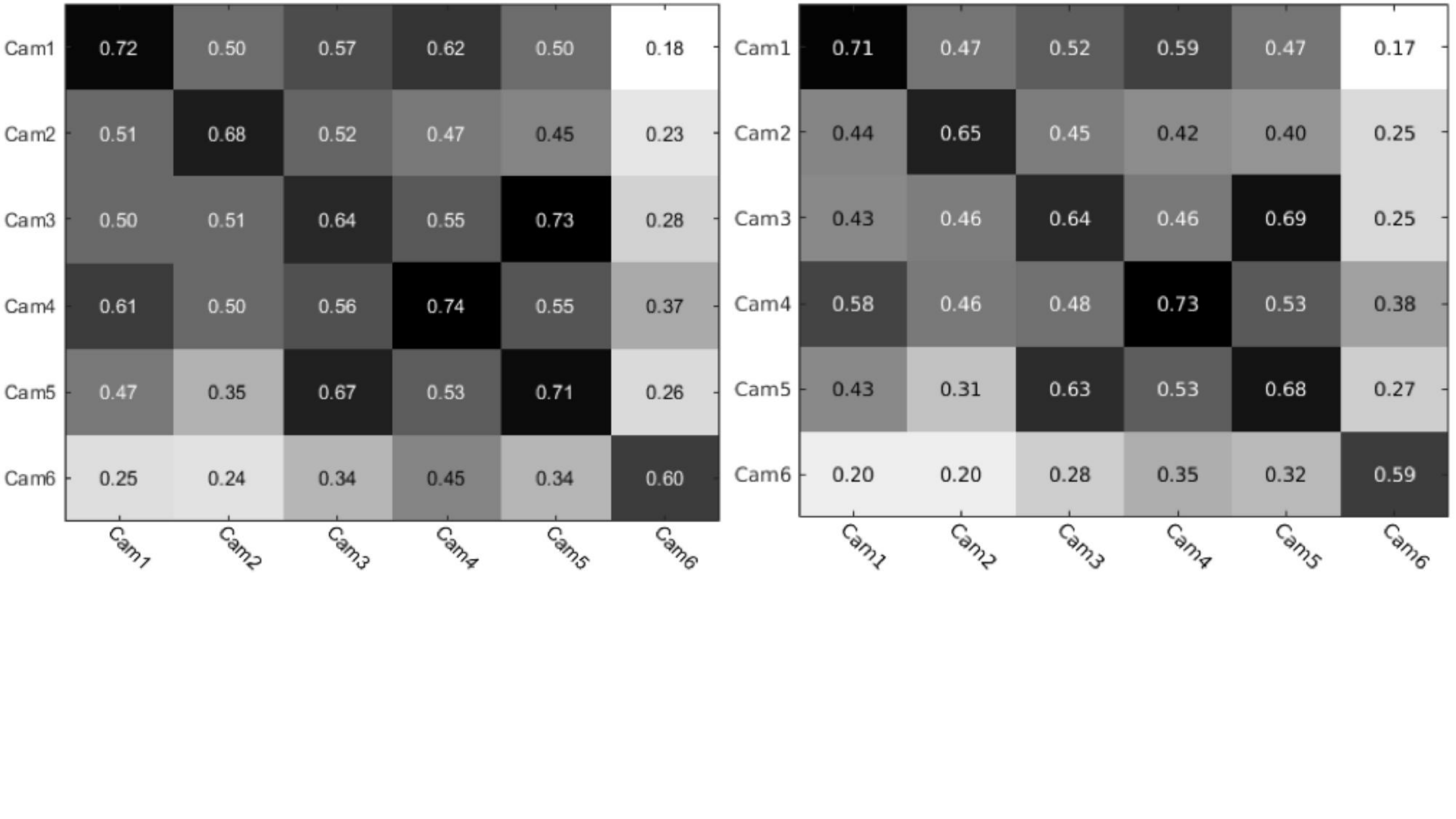}
		\vspace*{-2cm}
		\caption{The results show the performance of our algorithm on MARS (both using CNN + XQDA) when the final ranking is done using membership score (\textbf{left}) and using pairwise euclidean distance (\textbf{right}). 
		}
		\label{fig:NotionOfMembershipScore1}
	\end{figure*}
	
	
	\subsection{Evaluation on MARS dataset:} In Table \ref{table:rank1} we compare our results (using the same settings as in \cite{ZheBieSunWanSuWanTiaECCV16}) on MARS dataset with the state-of-the-art methods.  The proposed approach achieves  3\% improvement. In table \ref{table:NotionOfMembershipScore} the results show performance of our  and state-of-the-art approach \cite{ZheBieSunWanSuWanTiaECCV16} in solving the within- (average of the diagonal of the confusion matrix, Fig. \ref{fig:NotionOfMembershipScore1}) and across-camera (off-diagonal average) ReID using average precision. Our approach shows up to 10\% improvement in the across-camera ReID and up to 6\% improvement in the within camera ReID.  
	
	\begin{table}[h!]
		{
			\centering
			\begin{tabular}{l|r}
				Methods                          & rank 1 \\ \hline
				HLBP + XQDA             & 18.60 \\ \hline
				BCov + XQDA              & 9.20 \\ \hline
				LOMO + XQDA                   & 30.70 \\ \hline
				BoW + KISSME            & 30.60 \\ \hline
				SDALF + DVR     & 4.10 \\ \hline
				HOG3D + KISSME    & 2.60 \\ \hline
				CNN + XQDA \cite{ZheBieSunWanSuWanTiaECCV16}   & 65.30 \\ \hline
				CNN + KISSME \cite{ZheBieSunWanSuWanTiaECCV16}   & 65.00 \\ \hline
				Ours   & \textbf{68.22} \\ \hline
			\end{tabular}
			\caption{The table shows the comparison (based on rank-1 accuracy) of our approach  with the state-of-the-art approaches: SDALF \cite{FarBazPerMurCriCVPR10}, HLBP \cite{XioGouCamSznECCV14}, BoW \cite{ZheSheTiaWanWanTiaICCV15}, BCov \cite{MaSuJurIVC14}, LOMO \cite{LiaHuZhuLiCVPR15}, HOG3D \cite{KlaMarSchBMCV08} on MARS dataset.}
			\label{table:rank1}
		}
	\end{table}


	\begin{table}[h!]
		\centering
		\begin{tabular}{c|c|c|r}		
			{Feature+Distance}&{Methods}  &{Within } & Across \\ \cline{1-4}
			\cline{1-4}
			\multicolumn{1}{ c  }{\multirow{3}{*}{CNN + Eucl} } &
			\multicolumn{1}{ |c| }{\cite{ZheBieSunWanSuWanTiaECCV16}} & 0.59 & 0.28     \\ \cline{2-4}
			\multicolumn{1}{ c  }{}        &
			\multicolumn{1}{ |c| }{Ours (PairwiseDist)} & 0.59 & 0.29     \\ \cline{2-4}
			\multicolumn{1}{ c  }{}                        &
			\multicolumn{1}{ |c| }{Ours (MembershipS)} & \textbf{0.60} & \textbf{0.29}     \\ \cline{1-4}
			\multicolumn{1}{ c  }{\multirow{3}{*}{CNN + KISSME} } &
			\multicolumn{1}{ |c| }{\cite{ZheBieSunWanSuWanTiaECCV16}} & 0.61 & 0.34  \\ \cline{2-4}
			\multicolumn{1}{ c  }{}                       &
			\multicolumn{1}{ |c| }{Ours (PairwiseDist)} & 0.64 & 0.41   \\ \cline{2-4}
			\multicolumn{1}{ c  }{}                        &
			\multicolumn{1}{ |c| }{Ours (MembershipS)} & \textbf{0.67} & \textbf{0.44} \\
			\cline{1-4}
			\multicolumn{1}{ c  }{\multirow{3}{*}{CNN + XQDA} } &
			\multicolumn{1}{ |c| }{\cite{ZheBieSunWanSuWanTiaECCV16}} & 0.62 & 0.35  \\ \cline{2-4}
			\multicolumn{1}{ c  }{}                      &
			\multicolumn{1}{ |c| }{Ours (PairwiseDist)} & 0.65 & 0.42   \\ \cline{2-4}
			\multicolumn{1}{ c  }{}                        &
			\multicolumn{1}{ |c| }{Ours (MembershipS)} & \textbf{0.68} & \textbf{0.45} \\ \cline{2-4}

			\cline{1-4}
			
		\end{tabular}
		
		
		\caption{The results show performance of our(using pairwise distance and membership score) and state-of-the-art approach \cite{ZheBieSunWanSuWanTiaECCV16} in solving within- and across-camera ReID using average precision on MARS dataset using CNN feature and different distance metrics.
		}
		\label{table:NotionOfMembershipScore}
	\end{table}
	To show how much meaningful the notion of centrality of constrained dominant set is, we  conduct an experiment on the MARS dataset computing the final ranking using the membership score and pairwise distances. The confusion matrix in Fig. \ref{fig:NotionOfMembershipScore1} shows the detail result of both the within cameras (diagonals) and across cameras (off-diagonals), as we consider tracks from each camera as query. Given a query, a set which contains the query is extracted using the constrained dominant set framework.  Note that constraint dominant set comes with the membership scores for all members of the extracted set. We show  in Figure \ref{fig:NotionOfMembershipScore1} the results  based on the final ranking obtained using membership scores (\textbf{left}) and using pairwise Euclidean distance between the query and the extracted nodes(\textbf{right}). As can be seen from the results in Table \ref{table:NotionOfMembershipScore} (average performance) the use of membership score outperforms the pairwise distance approach, since it captures the interrelation among  targets.

		\begin{figure*}[h!]
			\centering
			\includegraphics[width=.7\linewidth,trim=4.3cm 8cm 4.5cm 9.5cm,clip]{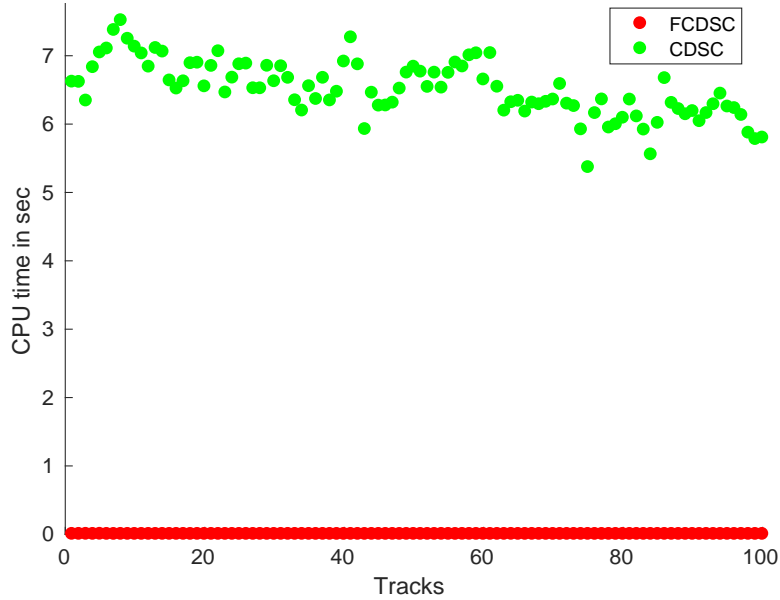}
			\caption{CPU time taken for each track association using our proposed fast approach (FCDSC - fast CDSC) and CDSC.}
			\label{plot:time_plot}
		\end{figure*}

		\begin{figure*}[h!]
			\centering
			\includegraphics[width=.7\linewidth,trim=3.8cm 8cm 4.5cm 9.5cm,clip]{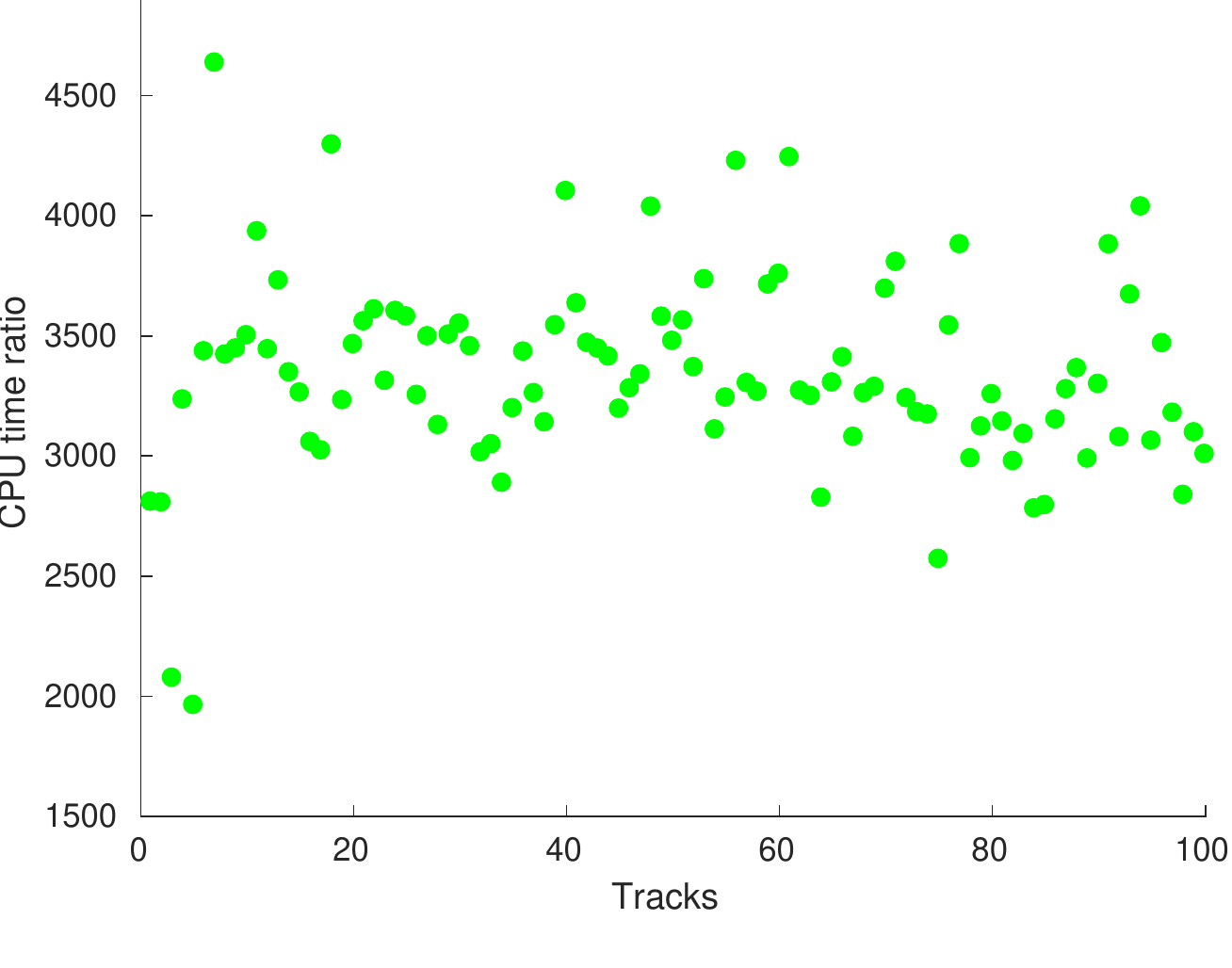}
			\caption{The ratio of CPU time taken between CDSC and proposed fast approach (FCDSC), computed as CPU time for CDSC/CPU time for FCDSC.}
			\label{plot:cpuTimeRatio}
		\end{figure*}

	\subsection{Computational Time.} Figure \ref{plot:time_plot} shows the time taken for each track - from 100 randomly selected (query) tracks - to be associated, with the rest of the (gallery) tracks, running CDSC over the whole graph (CDSC without speedup) and running it on a small portion of the graph using the proposed approach (called FCDSC, CDSC with speedup). The vertical axis is the CPU time  in seconds and horizontal axis depicts the track IDs.  As it is evident from the plot,our approach takes a fraction of second  (red points in Fig. \ref{plot:time_plot}). Conversely, the CDSC takes up to 8 seconds for some cases (green points in Fig. \ref{plot:time_plot}). Fig. \ref{plot:cpuTimeRatio} further elaborates how fast our proposed approach is over CDSC, where the vertical axis represents the ratio between CDSC (numerator) and FCDSC (denominator) in terms of CPU time. This ratio ranges from 2000 (the proposed FCDSC 2000x faster than CDSC) to a maximum of above 4500.

	\section{Summary} 
	\label{sufmmary_CDSC_tracker}
	In this chapter, we presented a constrained dominant set clustering (CDSC) based framework for  solving multi-target tracking problem in multiple non-overlapping cameras. The proposed method utilizes a three layers hierarchical approach, where within-camera tracking is solved using first two layers of our framework resulting in  tracks for each person, and later in the third layer the proposed across-camera tracker merges tracks of the same person across different cameras. Experiments on a challenging real-world dataset (MOTchallenge DukeMTMCT) validate the effectivness of our model. 
	
	We further perform additional experiments to show effectiveness of the proposed across-camera tracking on one of the largest video-based people re-identification datasets (MARS). Here each query is treated as a constraint set and its corresponding members in the resulting constrained dominant set cluster are considered as possible candidate matches to their corresponding query.

	There are few directions we would like to pursue in our future research. In this work, we consider a static cameras with known topology but it is important for the approach to be able to handle challenging scenario, were some views are from cameras with ego motion (e.g., PTZ cameras or taken from mobile devices) with unknown camera topology. Moreover, here we consider features from static images, however, we believe video features which can be extracted using LSTM could boost the performance and help us extend the method to handle challenging scenarios. 
 

\chapter{Large-scale Image Geo-Localization \\ Using Dominant Sets}\label{Geo-localization}

This chapter presents a new approach for the challenging problem of geo-localization  using image matching in a structured database of city-wide reference images with known GPS coordinates.  We cast geo-localization as a clustering problem of local image features. Akin to existing approaches to the problem, our framework builds on low-level features which allow local matching between images. For each local feature in the query image, we find its approximate nearest neighbors in the reference set. Next, we cluster the features from reference images using Dominant Set clustering. We obtain  multiple clusters (different local maximizers) and obtain a robust final solution to the problem using multiple weak solutions through constrained Dominant Set clustering on global image features, where we enforce the constraint that the query image must be included in the cluster. This second level of clustering also bypasses heuristic approaches to voting and selecting the reference image that matches to the query. We evaluate the proposed framework on an existing dataset of 102k street view images as well as a new larger dataset of 300k images, and show that it outperforms the state-of-the-art by 20\% and 7\%, respectively, on the two datasets.

	\section{Introduction}
Image geo-localization, the problem of determining the location of an image using just the visual information, is remarkably difficult. Nonetheless, images often contain useful visual and  contextual  informative cues which allow us to determine the location of an image with variable confidence. The foremost of these cues are landmarks, architectural details, building textures and colors, in addition to road markings and surrounding vegetation.
	
	Recently, the geo-localization through image-matching approach was proposed in \cite{zamir2010accurate, amirshahpami2014}. In \cite{zamir2010accurate}, the authors find the first nearest neighbor (NN) for each local feature in the query image, prune outliers and use a heuristic voting scheme for selecting the matched reference image. The follow-up work \cite{amirshahpami2014} relaxes the restriction of using only the first NN and proposed Generalized Minimum Clique Problem (GMCP) formulation for solving this problem. However, GMCP formulation can only handle a fixed number of nearest neighbors for each query feature. The authors used 5 NN, and found that increasing the number of NN drops the performance. Additionally, the GMCP formulation selects exactly one NN per query feature. This makes the optimization sensitive to outliers, since it is possible that none of the 5 NN is correct. Once the best NN is selected for each query feature, a very simple voting scheme is used to select the best match. Effectively, each query feature votes for a single reference image, from which the NN was selected for that particular query feature. This often results in identical number of votes for several images from the reference set. Then, both \cite{zamir2010accurate, amirshahpami2014} proceed with randomly selecting one reference image as the correct match to infer GPS location of the query image. Furthermore, the GMCP is a binary-variable NP-hard problem, and due to the high computational cost, only a single local minima solution is computed in \cite{amirshahpami2014}.
	
	In this chapter, we present a novel approach to image geo-localization by robustly finding a matching reference image to a given query image. This is done by finding correspondences between local features of the query and reference images. We first introduce automatic NN selection into our framework, by exploiting the discriminative power of each NN feature and employing different number of NN for each query feature. That is, if the distance between query and reference NNs is similar, then we use several NNs since they are ambiguous, and the optimization is afforded with more choices to select the correct match. On the other hand, if a query feature has very few low-distance reference NNs, then we use fewer NNs to save the computation cost. Thus, for some cases we use fewer NNs, while for others we use more requiring on the average approximately the same amount of computation power, but improving the performance, nonetheless. This also bypasses the manual tuning of the number of NNs to be considered, which can vary between datasets and is not straightforward. 
	
	Our approach to image geo-localization is based on \textit{Dominant Set clustering} (DSC) - a well-known generalization of maximal clique problem to edge-weighted graphs- where the goal is to extract the most compact and coherent set. It's intriguing connections to evolutionary game theory allow us to use efficient game dynamics, such as replicator dynamics and infection-immunization dynamics (InImDyn). InImDyn has been shown to have a linear time/space complexity for solving standard quadratic programs (StQPs), programs which deal with finding the extrema of a quadratic polynomial over the standard simplex \cite{RotBomGEB2011,BulPelBomCVIU11}. The proposed approach is on average 200 times faster and yields an improvement of 20\% in the accuracy of geo-localization compared to \cite{zamir2010accurate, amirshahpami2014}. This is made possible, in addition to the dynamics, through the flexibility inherent in DSC, which unlike the GMCP formulation avoids any hard constraints on memberships. This naturally handles outliers, since their membership score is lower compared to inliers present in the cluster. Furthermore, our solution uses a linear relaxation to the binary variables, which in the absence of hard constraints is solved through an iterative algorithm resulting in massive speed up.
	
	Since the dynamics and linear relaxation of binary variables allow our method to be extremely fast, we run it multiple times to obtain several local maxima as solutions. Next, we use a query-based variation of DSC to combine those solutions to obtain a final robust solution. The query-based DSC uses the soft-constraint that the query, or a group of queries, must always become part of the cluster, thus ensuring their membership in the solution. We use a fusion of several global features to compute the cost between query and reference images selected from the previous step. The members of the cluster from the reference set are used to find the geo-location of the query image. Note that, the GPS location of matching reference image is also used as a cost in addition to visual features to ensure both visual similarity and geographical proximity.
	
	GPS tagged reference image databases collected from user uploaded images on Flickr have been typically used for the geo-localization task. The query images in our experiments have been collected from Flickr, however, the reference images were collected from Google Street View. The data collected through Flickr and Google Street View differ in several important aspects: the images downloaded from Flickr are often redundant and repetitive, where images of a particular building, landmark or street are captured multiple times by different users. Typically, popular or tourist spots have relatively more images in testing and reference sets compared to less interesting parts of the urban environment. An important constraint during evaluation is that the distribution of testing images should be similar to that of reference images. On the contrary, Google Street View reference data used in this work contains only a single sample of each location of the city. However, Street View does provide spherical $360^{\circ}$ panoramic views, , approximately 12 meters apart, of most streets and roads. Thus, the images are uniformly distributed over different locations, independent of their popularity. The comprehensiveness of the data ensures that a correct match exists; nonetheless, at the same time, the sparsity or uniform distribution of the data makes geo-localization difficult, since every location is captured in only few of the reference images. The difficulty is compounded by the distorted, low-quality nature of the images as well.

	The main contributions of this chapter are summarized as follows: 
	\begin{itemize}
	 
		\item We present a robust and computationally efficient approach for the problem of large-scale image geo-localization by locating images in a structured database of city-wide reference images with known GPS coordinates.   
		
		\item  We formulate geo-localization problem in terms of a more generalized form of dominant sets framework which incorporates weights from the nodes in addition to edges. 
		
		\item We take a two-step approach to solve the problem. The first step uses local features to find putative set of reference images (and is therefore faster), whereas the second step uses global features and a constrained variation of dominant sets to refine results from the first step, thereby, significantly boosting the geo-localization performance.
		
		\item We have collected new and more challenging high resolution reference dataset (\textit{\textbf{WorldCities}} dataset) of 300K Google street view images.
		
	\end{itemize}
	
	The rest of the chapter is structured as follows. We present literature relevant to our problem in Sec. \ref{secRelatedWork}, followed by technical details of the proposed approach in Sec. \ref{secFramework}, while constrained dominant set based post processing step is discussed in Sec. \ref{post-processing}. This is followed by dataset description  in section \ref{Dataset_discription}. Finally, we provide results of our extensive evaluation in Sec. \ref{secExperiments} and summary in Sec. \ref{geo_summary}.

\section{Related Work} \label{secRelatedWork}
The computer vision literature on the problem of geo-localization can be divided into three categories depending on the scale of the datasets used: landmarks or buildings \cite{avrithis2010retrieving,chen2011city,quack2008world,zheng2009tour}, city-scale including streetview data \cite{jin2015predicting}, and worldwide \cite{hays2008im2gps,hays2015large,weyand2016planet}. Landmark recognition is typically formulated as an image retrieval problem \cite{avrithis2010retrieving,quack2008world, zheng2009tour,gammeter2009know,johns2011images}. For geo-localization of landmarks and buildings, Crandall \etal \cite{crandall2009mapping} perform structural analysis in the form of spatial distribution of millions of geo-tagged photos. This is used in conjunction with visual and meta data from images to geo-locate them. The datasets for this category contain many images near prominent landmarks or images. Therefore, in many works \cite{avrithis2010retrieving,quack2008world}, similar looking images belonging to same landmarks are often grouped before geo-localization is undertaken.

For citywide geo-localization of query images, Zamir and Shah \cite{zamir2010accurate} performed matching using SIFT features, where each feature votes for a reference image. The vote map is then smoothed geo-spatially and the peak in the vote map is selected as the location of the query image. They also compute 'confidence of localization' using the Kurtosis measure as it quantifies the peakiness of vote map distribution. The extension of this work in \cite{amirshahpami2014} formulates the geo-localization as a clique-finding problem where the authors relax the constraint of using only one nearest neighbor per query feature. The best match for each query feature is then solved using Generalized Minimum Clique Graphs, so that a simultaneous solution is obtained for all query features in contrast to their previous work \cite{zamir2010accurate}. In similar vein, Schindler \etal \cite{SchindlerBS07} used a dataset of 30,000 images corresponding to 20 kilometers of street-side data captured through a vehicle using vocabulary tree. Sattler \etal \cite{SatHavSchPolCVPR2016} investigated ways to explicitly handle geometric bursts by analyzing the geometric relations between the different database images retrieved by a query. Arandjelovic´ \etal \cite{AraGroTorPajSicCVPR2016} developed a convolutional neural network architecture for place recognition that aggregates mid-level (conv5) convolutional features extracted from the entire image into a compact single vector representation amenable to efficient indexing. 
Torii \etal \cite{TorSivOkuPajPAMI2015} exploited repetitive structure for visual place recognition, by robustly detecting repeated image structures and a simple modification of weights in the bag-of-visual-word model. Zeisl \etal \cite{BerTorMarICCV2015} proposed a voting-based pose estimation strategy that exhibits linear complexity in the number of matches and thus facilitates to consider much more matches.

For geo-localization at the global scale, Hays and Efros \cite{hays2008im2gps} were the first to extract coarse geographical location of query images using Flickr collected across the world. Recently, Weyand \etal \cite{weyand2016planet} pose the problem of geo-locating images in terms of classification by subdividing the surface of the earth into thousands of multi-scale geographic cells, and train a deep network using millions of geo-tagged images. In the regions where the coverage of photos is dense, structure-from-motion reconstruction is used for matching query images \cite{agarwal2009building,li2012worldwide,sattler2011fast,sattler2012improving}. Since the difficulty of the problem increases as we move from landmarks to city-scale and finally to worldwide, the performance also drops.
There are many interesting variations to the geo-localization problem as well. Sequential information such as chronological order of photos was used by \cite{kalogerakis2009image} to geo-locate photos. Similarly, there are methods to find trajectory of a moving camera by geo-locating video frames using Bayesian Smoothing \cite{vaca2012city} or geometric constraints \cite{hakeem2006estimating}. Chen and Grauman \cite{chen2011clues} present Hidden Markov Model approach to match sets of images with sets in the database for location estimation. Lin \etal \cite{lin2013cross} use aerial imagery in conjunction with ground images for geo-localization. Others \cite{lin2015learning,workman2015wide} approach the problem by matching ground images against a database of aerial images. Jacob \etal \cite{jacobs2007geolocating} geo-localize a webcam by correlating its video-stream with satellite weather maps over the same time period. Skyline2GPS \cite{ramalingam2010skyline2gps} uses street view data and segments the skyline in an image captured by an upward-facing camera by matching it against a 3D model of the city.

Feature discriminativity has been explored by \cite{arandjelovic2014dislocation}, who use local density of descriptor space as a measure of descriptor distinctiveness, i.e. descriptors which are in a densely populated region of the descriptor space are deemed to be less distinctive. Similarly, Bergamo \etal \cite{bergamo2013leveraging} leverage Structure from Motion to learn discriminative codebooks for recognition of landmarks. In contrast, Cao and Snavely \cite{cao2013graph} build a graph over the image database, and learn local discriminative models over the graph which are used for ranking database images according to the query. Similarly, Gronat \etal \cite{gronat2013learning} train discriminative classifier for each landmark and calibrate them afterwards using statistical significance measures. Instead of exploiting discriminativity, some works use similarity of features to detect repetitive structures to find locations of images. For instance, Torii \etal \cite{torii2013visual} consider a similar idea and find repetitive patterns among features to place recognition. Similarly, Hao \etal \cite{hao20123d} incorporate geometry between low-level features, termed 'visual phrases', to improve the performance on landmark recognition.

Our work is situated in the middle category, where given a database of images from a city or a group of cities, we aim to find the location where a test image was taken from. Unlike landmark recognition methods, the query image may or may not contain landmarks or prominent buildings. Similarly, in contrast to methods employing reference images from around the globe, the street view data exclusively contains man-made structures and rarely natural scenes like mountains, waterfalls or beaches.

\begin{figure*}[h!]
	\centering
	\includegraphics[width=1\linewidth ,trim=0cm 1.8cm 0cm 2.3cm, clip]{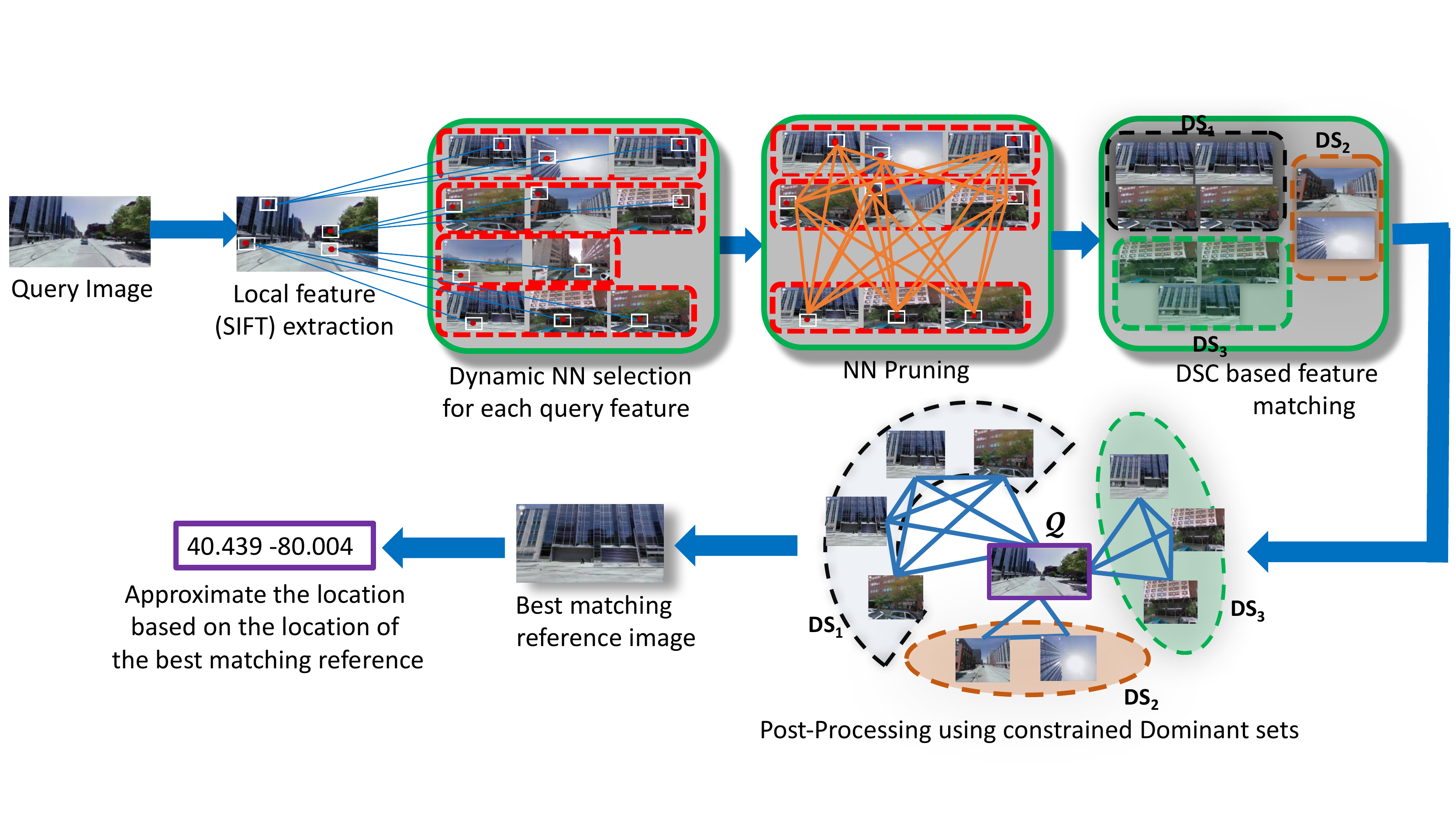}
	\caption{Overview of the proposed method.}
	\label{Overview}
\end{figure*}

\section{Image Matching Based Geo-Localization}\label{secFramework}
Fig. \ref{Overview} depicts the overview of the proposed approach. Given a set of reference images, e.g., taken from Google Street View, we extract local features (hereinafter referred as \textit{reference features}) using SIFT from each reference image. We then organize them in a k-means tree \cite{DBLP:conf/visapp/MujaL09}.

First, for each local feature extracted from the query image (hereinafter referred as \textit{query feature}), we dynamically collect nearest neighbors based on how distinctive the two nearest neighbors are relative to their corresponding query feature. Then, we remove query features, along with their corresponding reference features, if the ratio of the distance between the first and the last nearest neighbor is too large (Sec.~\ref{secAuto}). If so, it means that the query feature is not very informative for geo-localization, and is not worth keeping for further processing. In the next step, we formalize the problem of finding matching reference features to query features as a DSC (Dominant Set Clustering) problem, that is, selecting reference features which form a coherent and most compact set (Sec.~\ref{dsc-feature matching}). Finally, we employ constrained dominant-set-based post-processing step to choose the best matching reference image and use the location of the strongest match as an estimation of the location of the query image (Sec.~\ref{post-processing}).

\subsection{Dynamic Nearest Neighbor Selection and Query Feature Pruning}\label{secAuto}
For each of $N$ query features detected in the query image, we collect their corresponding nearest neighbors (\emph{NN}). Let $v_m^i$ be the $m^{th}$ nearest neighbor of $i^{th}$ query feature $q^i$, and $m \in \mathbb{N} : 1\leq m \leq \mid NN^i \mid$ and $i \in \mathbb{N} : 1 \leq i \leq N$, where $\mid\cdot\mid$ represents the set cardinality and $NN^i$ is the set of NNs of the $i^{th}$ query feature.
In this work, we propose a dynamic NNs selection technique based on how distinctive two consecutive elements are in a ranked list of neighbors for a given query feature, and employ different number of nearest neighbors for each query feature. 
%
%

As shown in Algorithm (\ref{alg:NNselection}), we add the $(m+1)^{th}$ NN of the $i^{th}$ query feature, $v_{m+1}^{i}$ , if the ratio of the two consecutive NN is greater than $\theta$, otherwise we stop. In other words, in the case of a query feature which is not very discriminative, i.e., most of its NNs are very similar to each other, the algorithm continues adding NNs until a distinctive one is found. In this way, less discriminative query features will use more NNs to account for their ambiguity, whereas more discriminative query features will be compared with fewer NNs.
\begin{algorithm}
	\caption{: Dynamic Nearest Neighbor Selection for $i^{th}$ query feature ($q^i$)}
	\label{alg:NNselection}
	
	\textbf{Input}: the $i^{th}$ query feature ($q^i$) and all its nearest neighbors extracted from K-means tree $\{ v_1^i,v_2^i.......v_{|NN^i|}^i\} $\\
	\textbf{Output}: Selected Nearest Neighbors for the $i^{th}$ query feature $(\mathbb{V}^i)$
	
	\rule{1\linewidth}{0.02cm}
	
	\begin{algorithmic}[1]
		\Procedure{Dynamic NN Selection}{$ $}
		\State Initialize $\mathbb{V}^i = \{v_1^i\}$ and m=1  
		\While {$m<|NN^i|-1$}
		\If {$\frac{\parallel \xi(q^i)-\xi(v_m^i)\parallel}{\parallel \xi(q^i)-\xi (v_{m+1}^i) \parallel} >\theta$ } 
		\State $\mathbb{V}^i=\mathbb{V}^i \cup v_{m+1}^i $ \Comment{If so, add $v_{m+1}^i$ to our solution }
		\State $m=m+1$ \Comment{Go to the next neighbor}
		\Else
		\State  Break \Comment{If not, stop adding and exit}
		\EndIf
		
		\EndWhile
		\EndProcedure
	\end{algorithmic}
\end{algorithm}
\noindent \textbf{Query Feature Pruning.} For the geo-localization task, most of the query features that are detected from moving objects (such as cars) or the ground plane, do not convey any useful information. If such features are coarsely identified and removed prior to performing the feature matching, that will reduce the clutter and computation cost for the remaining features. Towards this end, we use the following pruning constraint which takes into consideration distinctiveness of the \textit{first} and the \textit{last} NN. In particular, if  $\left\|\xi\left(q^i\right)-\xi\left(v_1^i\right)\right\| / \left\| \xi\left(q^i\right)-\xi\left(v_{\left|NN^i\right|}^i\right) \right\| > \beta$, where $\xi(\cdot)$ represents an operator which returns the local descriptor of the argument node, then $q^i$ is removed, otherwise it is retained. That is, if the \textit{first} NN is similar to the \textit{last} NN (less than $\beta$), then the corresponding query feature along with its NNs are pruned since it is expected to be uninformative. 

We empirically set both thresholds, $\theta $ and $\beta$, in Algorithm (\ref{alg:NNselection})  and pruning step, respectively, to 0.7 and keep them fixed for all tests.

\subsection{Multiple Feature Matching Using Dominant Sets}\label{dsc-feature matching}

\subsubsection{Similarity Function and Dynamics for Multiple Feature Matching} \label{similarity_dynm}

In our framework, the set of nodes, $V$, represents all NNs for each query feature which survives the pruning step.
The edge set is defined as $E=\{(v^i_m ,v^j_n) \mid  i\neq j \} $, which signifies that all the nodes in $G$ are connected as long as their corresponding query features are not the same. The edge weight, $\omega : E \longrightarrow \mathbb{R}^+$ is defined as $\omega (v^i_m, v^j_n) = exp(-\Vert \psi (v^i_m)  - \psi (v^j_n) \Vert^2/{2\gamma^2})$, where $\psi(\cdot)$ represents an operator which returns the global descriptor of the parent image of the argument node and $\gamma$ is empirically set to $2^7$ . The edge weight, $\omega (v^i_m, v^j_n)$, represents a similarity between nodes $v^i_m $ and $v^j_n$ in terms of the global features of their parent images. The node score, $\zeta : V \longrightarrow \mathbb{R}^+$, is defined as $\zeta(v_m^i) = exp(-\Vert \xi(q^i)-\xi(v_m^i) \Vert^2/{2\gamma^2})$. The node score shows how similar the node $v^i_m$ is with its corresponding query feature in terms of its local features.

Matching the query features to the reference features requires identifying the correct NNs from the graph $G$ which maximize the weight, that is, selecting a node (NN) which forms a coherent (highly compact) set in terms of both global and local feature similarities.

Affinity matrix $\mat{A}$ represents the global similarity among reference images, which is built using GPS locations as a global feature and a node score $\vct{b}$ which shows how similar the reference image is with its corresponding query feature in terms of their local features. We formulate the following optimization problem, a more general form of the dominant set formulation:

\begin{equation}
	\label{general_quadratic_function}
	\begin{array}{ll}
		\text{maximize }  &  f(\vct x) = \vct x\T\mat{A}\vct x + \vct b\T\vct x, \\
		\text{subject to} &  \mathbf{x} \in \Delta.
	\end{array}
\end{equation}

The affinity $\mat{A}$ and the score $\vct{b}$ are computed as follows:
\begin{align}
	\mat{A}(v^i_m, v^j_n)&=
	\begin{cases}
		\omega (v^i_m, v^j_n)
		,&\text{for $i \neq j$,}\\
		0, &\text{otherwise,}
	\end{cases}\\
	\vct{b}(v_m^i)&= \zeta(v_m^i).
\end{align}

General quadratic optimization problems, like (\ref{general_quadratic_function}), are known to be NP-hard \cite{HorParTho2000}. However, in relaxed form, standard quadratic optimization problems can be solved using many algorithms which make full systematic use of data constellations. Off-the-shelf procedures find a local solution of (\ref{general_quadratic_function}), by following the paths of feasible points provided by game dynamics based on evolutionary game theory.

Interestingly, the general quadratic optimization problem can be rewritten in the form of standard quadratic problem. A principled way to do that is to follow the result presented in \cite{Bom98}, which shows that maximizing the general quadratic problem over the simplex can be homogenized as follows. Maximizing $ \vct x\T\mat{A}\vct x + \vct b\T\vct{x}, \text{subject to } \vct{x} \in \Delta $ is equivalent to maximizing $\vct x\T\mat B\vct{x}, \text{subject to } \vct{x} \in \Delta$, where $\mat{B}=\mat{A}+\vct{e}\vct{b}\T+\vct{b}\vct{e}\T$ and $\vct{e}$ = $\sum\limits_{i=1}^{n}\vct{e}_i=[1,1,....1]$, where $\vct{e}_i$ denotes the $i^{th}$ standard basis vector in $\mathbb{R}^n$. This can be easily proved by noting that the problem is solved in the simplex.

\begin{itemize}
	\item[] $\vct x\T\mat{A}\vct x + 2*\vct b\T\vct x = \vct x\T\mat{A}\vct x + \vct b\T\vct x + \vct{x}\T\vct{b}$,
	\item[]\hspace{2.55cm}$= \vct{x}\T\mat{A}\vct{x} + \vct{x}\T\vct{e}\vct{b}\T\vct{x} + \vct{x}\T\vct{b}\vct{e}\T\vct{x}$,
	\item[]\hspace{2.55cm}$= \vct{x}\T(\mat{A} + \vct{e}\vct{b}\T + \vct{b}\vct{e}\T)\vct{x}$,
	\item[]\hspace{2.55cm}$= \vct{x}\T\mat{B}\vct{x}$.
\end{itemize}

As can be inferred from the formulation of our multiple NN feature matching problem, DSC can be essentially used for solving the optimization problem. Therefore, by solving DSC for the graph $G$, the optimal solution that has the most agreement in terms of local and global features will be found. 



 It is important to note that, since our solution is a local solution and we do not know which local solution includes the best matching reference image, we determine several (typically three) locally optimal solutions. Unlike most of the previous approaches, which perform a simple voting scheme to select the best matching reference image, we introduce a post processing step utilizing a extension of dominant set called \textit{constrained dominant set}. 


\section{Post processing using constrained dominant sets}\label{post-processing}

Up to now, we devised a method to collect matching reference features corresponding to our query features. The next task is to select one reference image, based on feature matching between query and reference features, which best matches the query image. To do so, most of the previous methods follow a simple voting scheme, that is, the matched reference image with the highest vote is considered as the best match. 

This approach has  two important shortcomings.  First, if there are  equal votes for two or more reference images (which happens quite often), it will randomly select one, which makes it prone to outliers. Second, a simple voting scheme does not consider the similarity between the query image and the candidate reference images at the global level, but simply accounts for local matching of features. Therefore, we deal with this issue by proposing a post processing step, which considers the comparison of the global features of the images and employs \emph{constrained dominant set}, a framework that generalizes the dominant sets formulation and its variant \cite{PavPelPAMI07,PavPel03}.


In our post processing step, the user-selected query and the matched reference images are related using their {\em global} features and a unique local solution is then searched which contains the union of all the dominant sets containing the query. As customary, the resulting solution will be globally consistent both with the reference images and the query image and  due to the notion of \textit{centrality}, each element in the resulting solution will have a membership score, which depicts how similar a given reference image is with the rest of the images in the cluster. So, by virtue of this property of constrained dominant sets, we will select the image with the highest membership score as the final best matching reference image and approximate the location of the query with the GPS location of the best matched reference image.

 Given a user specified query, $\mathcal{Q} \subseteq \hat V$, we define the graph $\hat G=(\hat V,\hat E,\hat w)$, where the edges are defined as $\hat{E}=\{ (i,j) | i\neq j,  \{i,j\} \in \mathcal{DS}_n \vee  (i \in \mathcal{Q} \vee  j \in \mathcal{Q}) \}$, i.e., all the nodes are connected as long as they do not belong to different local maximizers, $\mathcal{DS}_n$, which represents the $n^{th}$ extracted dominant set. The set of nodes $\hat{V}$ represents all matched reference images (local maximizers) and query image, $\mathcal{Q}$. The edge weight $\hat{w}: \hat{E} \rightarrow\mathbb{R}^+$ is defined as:

$$
\hat{w}(i,j)=
\begin{cases}
\rho(i,j),&\text{for $i \neq j$, $i \in \mathcal{Q} \vee  j\in \mathcal{Q}$},\\
\mat{B}_n(i,j),&\text{for $i \neq j$, $\{i,j\} \in \mathcal{DS}_n$},\\
0, &\text{otherwise}
\end{cases}$$
\noindent where $\rho(i,j)$ is an operator which returns the global similarity of two given images $i$ and $j$, that is, $\rho (i, j) = exp(-\Vert \psi (i)  - \psi (j) \Vert^2/{2\gamma^2})$, $\mat{B}_n$ represents a sub-matrix of $\mat{B}$, which contains only members of $\mathcal{DS}_n$, normalized by its maximum value and finally $\mat{B}_n(i,j)$ returns the normalized affinity between the $i^{th}$ and $j^{th}$ members of $\mathcal{DS}_n$. The graph $\hat{G}$ can be represented by an $n\times n$ affinity matrix $\hat{\mat{A}} = (\hat{w}(i,j))$, where $n$ is the number of nodes in the graph.
Given a parameter $ \alpha > 0$, let us define the following parameterized variant of program (\ref{eq2}):
\begin{equation}
\label{eqn:parQP1}
\begin{array}{ll}
\text{maximize }  &  f_{\mathcal{Q}}^\alpha(\x) = \x\T (\hat {\mat{A}} - \alpha \hat I_{\mathcal{Q}}) \x, \\
\text{subject to} &  \mathbf{x} \in \Delta,
\end{array}
\end{equation}
\noindent where $\hat I_{\mathcal{Q}}$ is the $n \times n$ diagonal matrix whose diagonal elements are set to 1 in correspondence to the vertices contained in $\hat V \setminus \mathcal{Q}$ (a set $\hat V$ without the element $\mathcal{Q}$) and to zero otherwise.

Let $\mathcal{Q} \subseteq \hat V$, with $\mathcal{Q} \neq \emptyset$ and let $\alpha > \lambda_{\max}(\hat {\mat{A}}_{\hat V \setminus \mathcal{Q}}) $, where $\lambda_{\max}(\hat {\mat{A}}_{\hat V \setminus \mathcal{Q}})$ is the largest eigenvalue of the principal submatrix of $\hat{\mat{A}}$ indexed by the elements of $\hat V \setminus \mathcal{Q}$.
If $\x$ is a local maximizer of $f_{\mathcal{Q}}^\alpha$ in $\Delta$, then
$\sigma(\x) \cap \mathcal{Q} \neq \emptyset$. A complete proof can be found in \cite{ZemPelECCV16}.

The above result provides us with a simple technique to determine dominant-set clusters containing user-specified query vertices. Indeed, if $\mathcal{Q}$ is a vertex selected by the user, by setting
\begin{equation}
\label{alphabound_Geoo}
\alpha > \lambda_{\max}(\hat{\mat{A}}_{\hat V \setminus \mathcal{Q}}),
\end{equation}
we are guaranteed that all local solutions of (\ref{eqn:parQP1}) will have a support
that necessarily contains elements of $\mathcal{Q}$.
The performance of our post processing may vary dramatically among queries and we do not know in advance which global feature plays a major role. Figs. \ref{fig: Postprocessing1} and \ref{fig: Postprocessing2} show illustrative cases of different global features. In the case of Fig. \ref{fig: Postprocessing1}, HSV color histograms and GIST provide a wrong match (dark red node for HSV and dark green node for GIST), while both CNN-based global features (CNN6 and CNN7) matched it to the right reference image (yellow node). The second example, Fig. \ref{fig: Postprocessing2}, shows us that only CNN6 feature localized it to the right location while the others fail. Recently, fusing different retrieval methods has been shown to enhance the overall retrieval performance \cite{LiaSheLuFeiZiQi15,ShaMinTimKaiDim15}.

Motivated by \cite{LiaSheLuFeiZiQi15} we dynamically assign a weight, based on the effectiveness of a feature, to all global features based on the area under normalized score between the query and the matched reference images. The area under the curve is inversely proportional to the effectiveness of a feature. More specifically, let us suppose to have $\mathcal{G}$ global features and the distance between the query and the $j^{th}$ matched reference image ( $\mathcal{N}_j$), based on the $i^{th}$ global feature ($\mathcal{G}_i$), is computed as: $f_i^j=\psi_i(\mathcal{Q})-\psi_i(\mathcal{N}_j)$, where $\psi_i(\cdot)$ represents an operator which returns the $i^{th}$ global descriptor of the argument node. 
Let the area under the normalized score of $f_i$ be $\mathcal{A}_i$. The weight assigned for feature $\mathcal{G}_i$ is then computed as $w_i=\frac{1}{\mathcal{A}_i} / \sum\limits_{j=1}^{\mathcal{\left|G\right|}}\frac{1}{\mathcal{A}_j}$.

\begin{figure*}[h!]
	\includegraphics[width=1\linewidth,trim=5.00cm 0.0cm 5.55cm 0.0cm, clip]{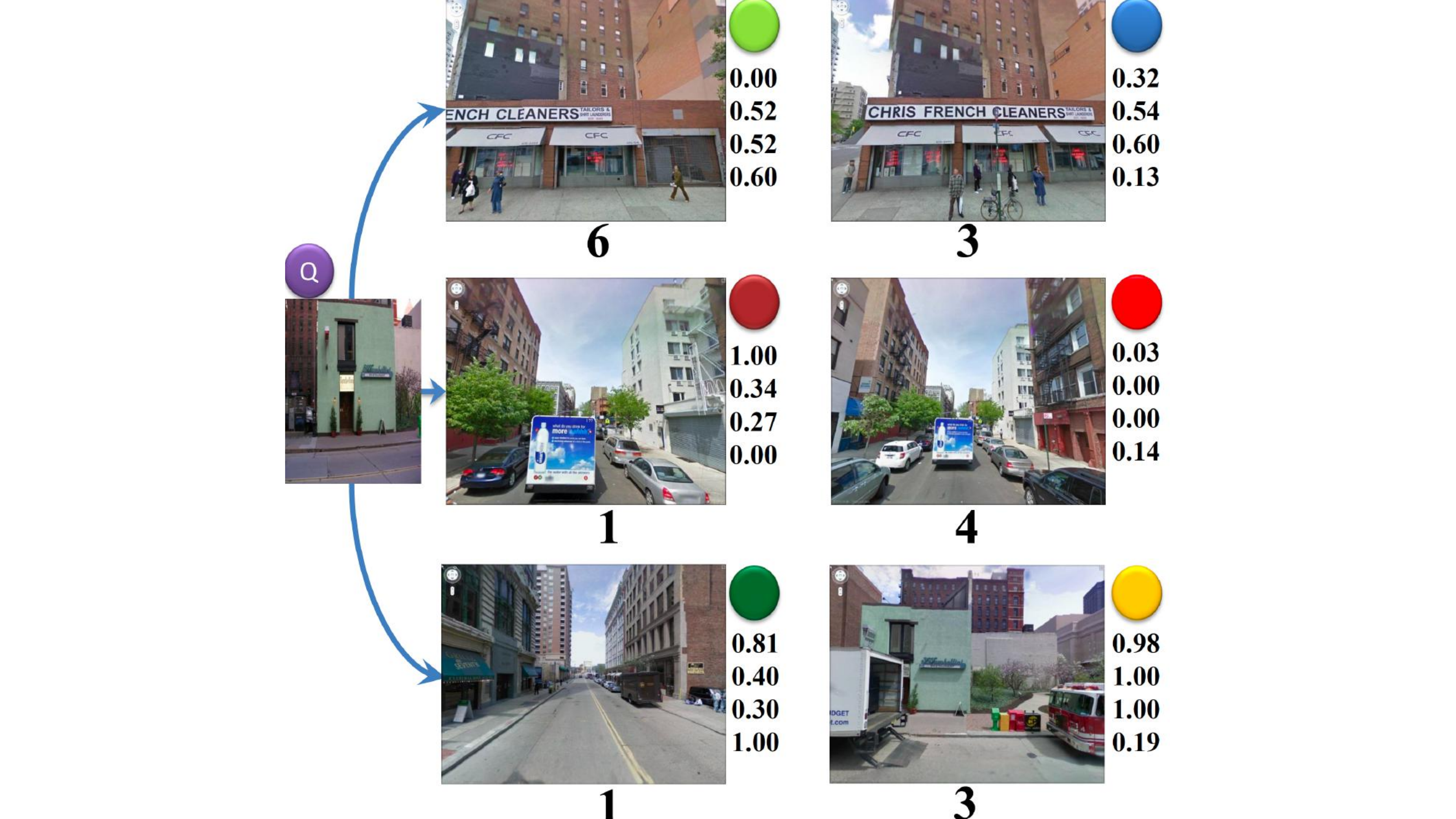}
	\caption{Exemplar output of the dominant set framework: {\bf Left:} query, {\bf Right:} each row shows corresponding reference images from the first, second and third local solutions (dominant sets), respectively, from top to bottom. The number under each image shows the frequency of the matched reference image, while those on the right side of each image show the min-max normalized scores of HSV, CNN6, CNN7 and GIST global features, respectively. The filled colors circles on the upper right corner of the images are used as reference IDs of the images. }
	\label{fig: Postprocessing1}
\end{figure*}

\begin{figure*}[h!]
	\includegraphics[width=1\linewidth,trim=0cm 0.0cm 0cm 0cm, clip]{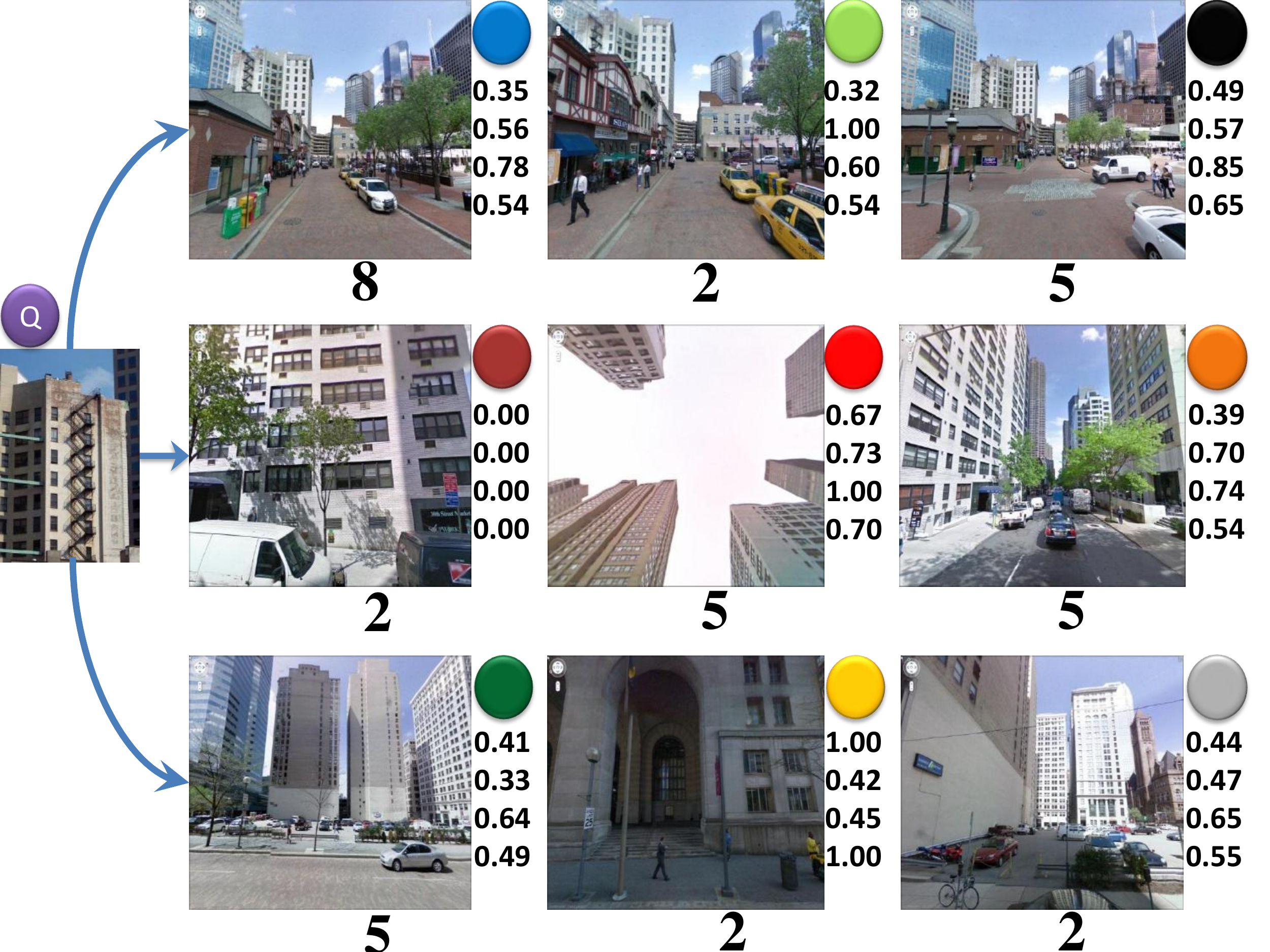}
	\caption{Exemplar output of the dominant set framework: {\bf Left:} query, {\bf Right:} each row shows corresponding reference images of the first, second and third local solutions (dominant sets), respectively, from top to bottom. The number under each image shows the mode of the matched reference image, while those on the right of each image show the min-max normalized scores of HSV, CNN6, CNN7 and GIST global features, respectively.}
	\label{fig: Postprocessing2}
\end{figure*}

Figs. \ref{fig: Postprocessing1} and \ref{fig: Postprocessing2} show illustrative cases of some of the advantages of having the post processing step. Both cases show the disadvantage of localization following heuristic approaches, as in \cite{zamir2010accurate,amirshahpami2014}, to voting and selecting the reference image that matches to the query. In each case, the matched reference image with the highest number of votes (shown under each image) is  the first node of the first extracted dominant set, but represents a wrong match. Both cases (Figs. \ref{fig: Postprocessing1} and \ref{fig: Postprocessing2}) also demonstrate that the KNN-based matching may lead to a wrong localization. For example, by choosing HSV histogram as a global feature, the KNN approach chooses as best match the dark red node in Fig. \ref{fig: Postprocessing1} and the yellow node in Fig. \ref{fig: Postprocessing2} (both with min-max value to 1.00). Moreover, it is also evident that choosing the best match using the first extracted local solution (i.e., the light green node in Fig. \ref{fig: Postprocessing1} and blue node in Fig. \ref{fig: Postprocessing2}), as done in \cite{amirshahpami2014}, may lead to a wrong localization,  since one cannot know in advance which local solution plays a major role. In fact, in the case of Fig. \ref{fig: Postprocessing1}, the third extracted dominant set contains the right matched reference image (yellow node), whereas in the case of Fig. \ref{fig: Postprocessing2} the best match is contained in the first local solution (the light green node).

The similarity between query $\mathcal{Q}$ and the corresponding matched reference images is computed using their global features such as HSV histogram, GIST \cite{OlivTorIJCV2001} and CNN (CNN6 and CNN7 are Convolutional Neural Network features extracted from ReLU6 and FC7 layers of pre-trained network, respectively \cite{SimZisCoRR2014}). For the different advantages and disadvantages of the global features, we refer interested readers to \cite{amirshahpami2014}.


Fig. \ref{fig: Postprocessing1} shows the top three extracted dominant sets with their corresponding frequency of the matched reference images (at the bottom of each image). Let $\mathcal{F}_i$ be the number  (cardinality) of local features, which belongs to $i^{th}$ reference image from the extracted sets and the total number of matched reference images be $\mathcal{N}$. We build an affinity matrix $\hat{\mat{B}}$ of size $\mathcal{S} = \sum\limits_{i=1}^{\mathcal{N}}\mathcal{F}_i + 1$ (e.g., for the example in Fig. \ref{fig: Postprocessing1}, the size $\mathcal{S}$ is 19 ). Fig. \ref{fig: PostprocessingGraph} shows the reduced graph for the matched reference images shown in Fig. \ref{fig: Postprocessing1}. Fig. \ref{fig: PostprocessingGraph} upper left, shows the part of the graph for the post processing. It shows the relation that the query has with matched reference images. The bottom left part of the figure shows how one can get the full graph from the reduced graph.
For the example in Fig. \ref{fig: PostprocessingGraph},  $\hat V=\{\mathcal{Q},1,2,2,2,3,4,4 .. ..... 6\}$.


\begin{figure}[t]
	\includegraphics[width=1\linewidth,trim=3cm 0.0cm 4cm 0cm, clip]{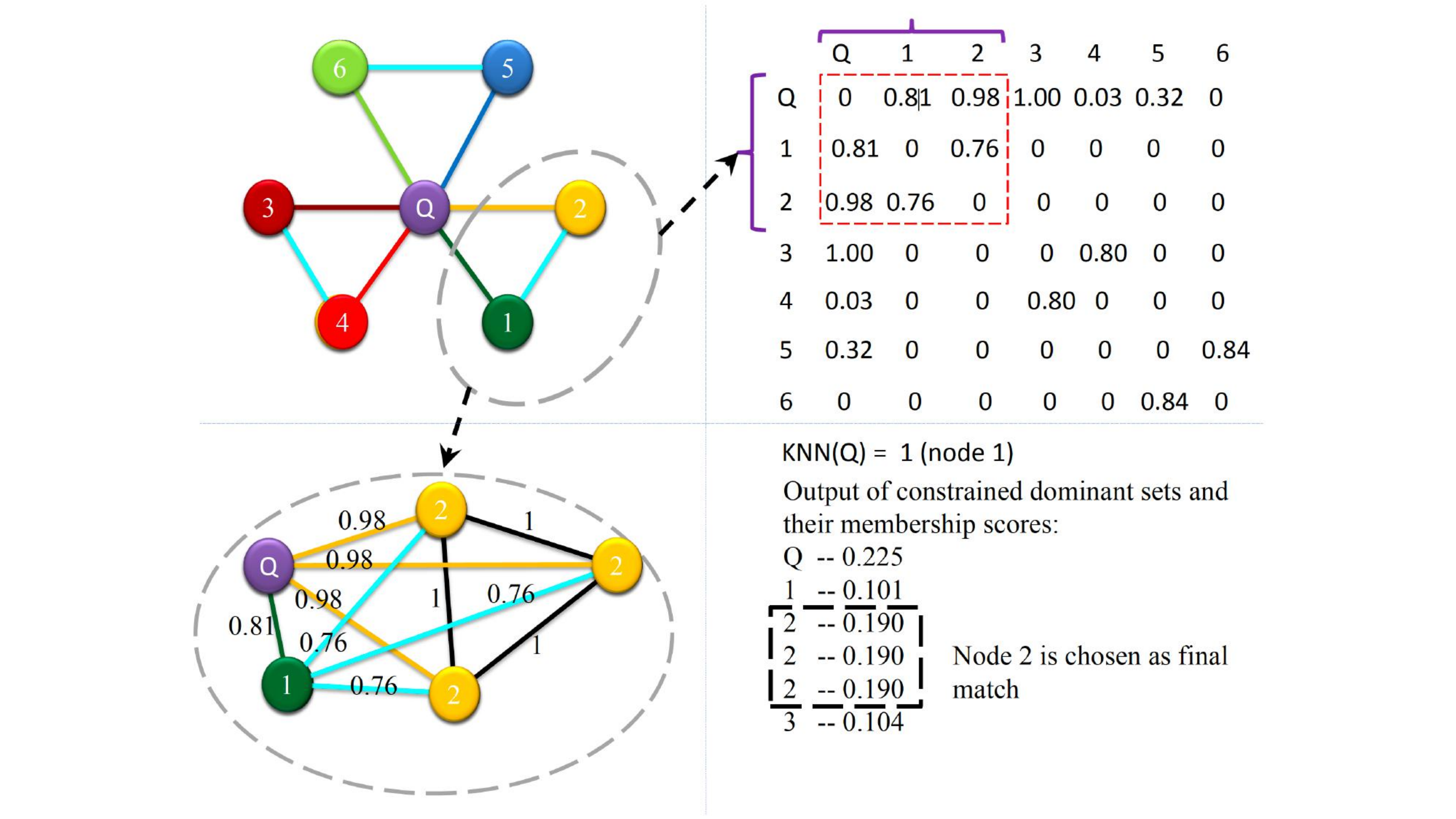}
	\caption{Exemplar graph for post processing. {\bf Top left:} reduced graph for Fig. \ref{fig: Postprocessing1} which contains unique matched reference images. {\bf Bottom left:} Part of the full graph which contains the gray circled nodes of the reduced graph and the query. {\bf Top right:} corresponding affinity of the reduced graph. {\bf Bottom right:} The outputs of nearest neighbor approach, consider only the node's pairwise similarity, (KNN($\mathcal{Q}$)=node 3 which is the dark red node) and constrained dominant sets approach ($\mathcal{CDS(Q)}$ = node 2 which is the yellow node).}
	\label{fig: PostprocessingGraph}
\end{figure}


The advantages of using constrained dominant sets are numerous. First, it provides a unique local (and hence global) solution whose support coincides with the union of all dominant sets of $\hat G $, which contains the query. Such solution contains all the local solutions which have strong relation with the user-selected query. As it can be observed in Fig. \ref{fig: PostprocessingGraph} (bottom right), the Constrained Dominant Set which contains the query $\mathcal{Q}$, $\mathcal{CDS(Q)}$, is the union of all overlapping dominant sets (the query, one green, one dark red and three yellow nodes) containing the query as one of the members. If we assume to have no cyan link between the green and yellow nodes, as long as there is a strong relation between the green node and the query, $\mathcal{CDS(Q)}$ will not be affected. In addition, due to the noise, the strong affinity between the query and the green node may be reduced, while still keeping the strong relation with the cyan link which, as a result, will preserve the final result. Second, in addition to fusing all the local solutions leveraging the notion of centrality, one of the interesting properties of dominant set framework is that it assigns to each image a score corresponding to how similar it is to the rest of the images in the solution. Therefore, not only it helps selecting the best local solution, but also choosing the best final match from the chosen local solution. Third, an interesting property of constrained dominant sets approach is that it not only considers the pairwise similarity of the query and the reference images, but also the similarity among the reference images. This property helps the algorithm avoid assignment of wrong high pairwise affinities. As an example, with reference to Fig. \ref{fig: PostprocessingGraph}, if we consider the nodes pairwise affinities, the best solution will be the dark red node (score 1.00). However, using constrained dominant sets and considering the relation among the neighbors, the solution bounded by the red dotted rectangle can be found, and by choosing the node with the highest membership score, the final best match is the yellow node which is more similar to the query image than the reference image depicted by the dark red node (see Fig. \ref{fig: Postprocessing1}).


\begin{figure*}[h!]
	\centering 
	\includegraphics[width=15cm,height=11cm ]{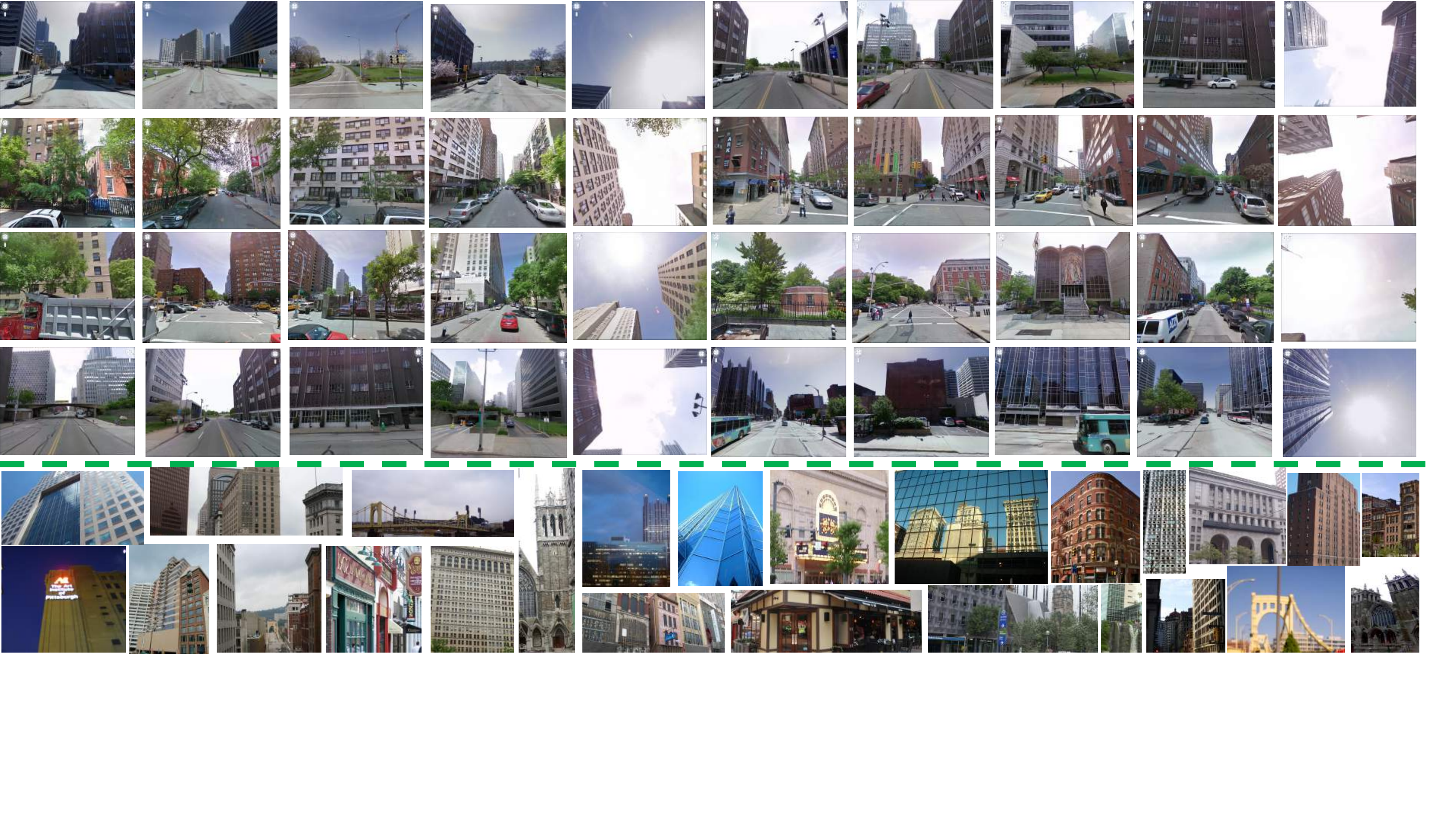}
	\vspace*{-2cm}
	\caption{The top four rows are sample street view images from eight different places of \textit{\textbf{WorldCities}} dataset. The bottom two rows are sample user uploaded images from the test set.}
	\label{sample_image}
\end{figure*}

\section{Experimental Results}\label{secExperiments}


\subsection{Dataset Description}\label{Dataset_discription}
We evaluate the proposed algorithm  using publicly available reference data sets of over 102k Google street view images \cite{amirshahpami2014} and a new dataset, \textit{\textbf{WorldCities}}, of high resolution 300k Google street view images collected for this  work. The 360 degrees view of each place mark is broken down into one top and four side view images. The \textit{\textbf{WorldCities}} dataset is publicly available. \footnote{\url{http://www.cs.ucf.edu/~haroon/UCF-Google-Streetview-II-Data/UCF-Google-Streetview-II-Data.zip}}

The 102k Google street view images dataset covers 204 Km of urban streets and the place marks  are approximately 12 m apart. It covers downtown and the neighboring areas of Orlando, FL; Pittsburgh, PA and partially Manhattan, NY.
The \textit{\textbf{WorldCities}} dataset is a new high resolution reference dataset of 300k street view images that covers 14 different cities from different parts of the world: Europe (Amsterdam, Frankfurt, Rome, Milan and Paris), Australia (Sydney and Melbourne), USA (Vegas, Los Angeles, Phoenix, Houston, San Diego, Dallas, Chicago). Existence of similarity in buildings around the world, which can be in terms  of  their wall designs, edges, shapes, color etc, makes the dataset more challenging than the other. Fig. \ref{sample_image} (top four rows) shows sample reference images taken from different place marks.

For the test set, we use 644 and 500 GPS-tagged user uploaded images downloaded from Picasa, Flickr and Panoramio for the 102k Google street view images and \textit{\textbf{WorldCities}} datasets, respectively. Fig. \ref{sample_image} (last two rows) shows sample test images. Throughout our experiment, we use the all the reference images from around the world to find the best match with the query image, not just with the ground truth city only.




\subsection{Quantitative Comparison With Other Methods}
\subsubsection{Performance on the 102k Google street view images Dataset}
The proposed approach has been then compared with the results obtained by state-of-the-art methods. In Fig. \ref{comparison_plot}, the horizontal axes shows the error threshold in meters and the vertical axes shows the percentage of the test set localized within a particular error threshold. Since the scope of this work is an accurate image localization at a city-scale level, test set images localized above 300 meter are considered a failure.

The black (-*-) curve shows localization result of the approach proposed in \cite{SchindlerBS07} which uses vocabulary tree to localize images. The {\color {red} red (-o-)} curve depicts the results of \cite{zamir2010accurate} where they only consider the first NN for each query feature as best matches which makes the approach very sensitive to the query features they select. Moreover, their approach suffers from lacking global feature information. The {\color{green} green (-o-)} curve illustrates the localization results of \cite{amirshahpami2014} which uses generalized maximum clique problem (GMCP) to solve feature matching problem and follows voting scheme to select the best matching reference image. The black (-o- and $-\lozenge-$) curves show localization results of MAC and RMAC, (regional) maximum activation of convolutions \cite{ToliasICLR2016,ToliasECCV2016}. These approaches build compact feature vectors that encode several image regions without the need to feed multiple inputs to the network. The {\color{cyan} cyan (-o-)} curve represents localization result of NetVLAD \cite{AraGroTorPajSicCVPR2016} which aggregates mid-level (conv5) convolutional features extracted from the entire image into a compact single vector representation amenable to effici,ent indexing. The {\color{cyan} cyan ($-\lozenge-$)} curve depicts localization result of NetVLAD but finetuned on our dataset. The {\color{blue} blue ($-\lozenge-$)} curve show localizaton result of approach proposed in \cite{SatHavSchPolCVPR2016} which exploits geometric relations between different database images retrieved by a query to handle geometric burstness. The {\color{blue} blue (-o-)} curve shows results from our baseline approach, that is,  we use voting scheme to select best match reference image and estimate the location of the query image. We are able to make a 10\% improvement w.r.t the other methods with only our baseline approach (without post processing). The {\color{magenta} magenta (-o-)} curve illustrates geo-localization results of our proposed approach using dominant set clustering based feature matching and constrained dominant set clustering based post processing. As it can be seen, our approach shows about 20\% improvement over the state-of-the-art techniques.

\begin{figure*}[h!]
	\centering
	\includegraphics[width=1\linewidth,trim=5.00cm 0.0cm 3.55cm 0.0cm, clip]{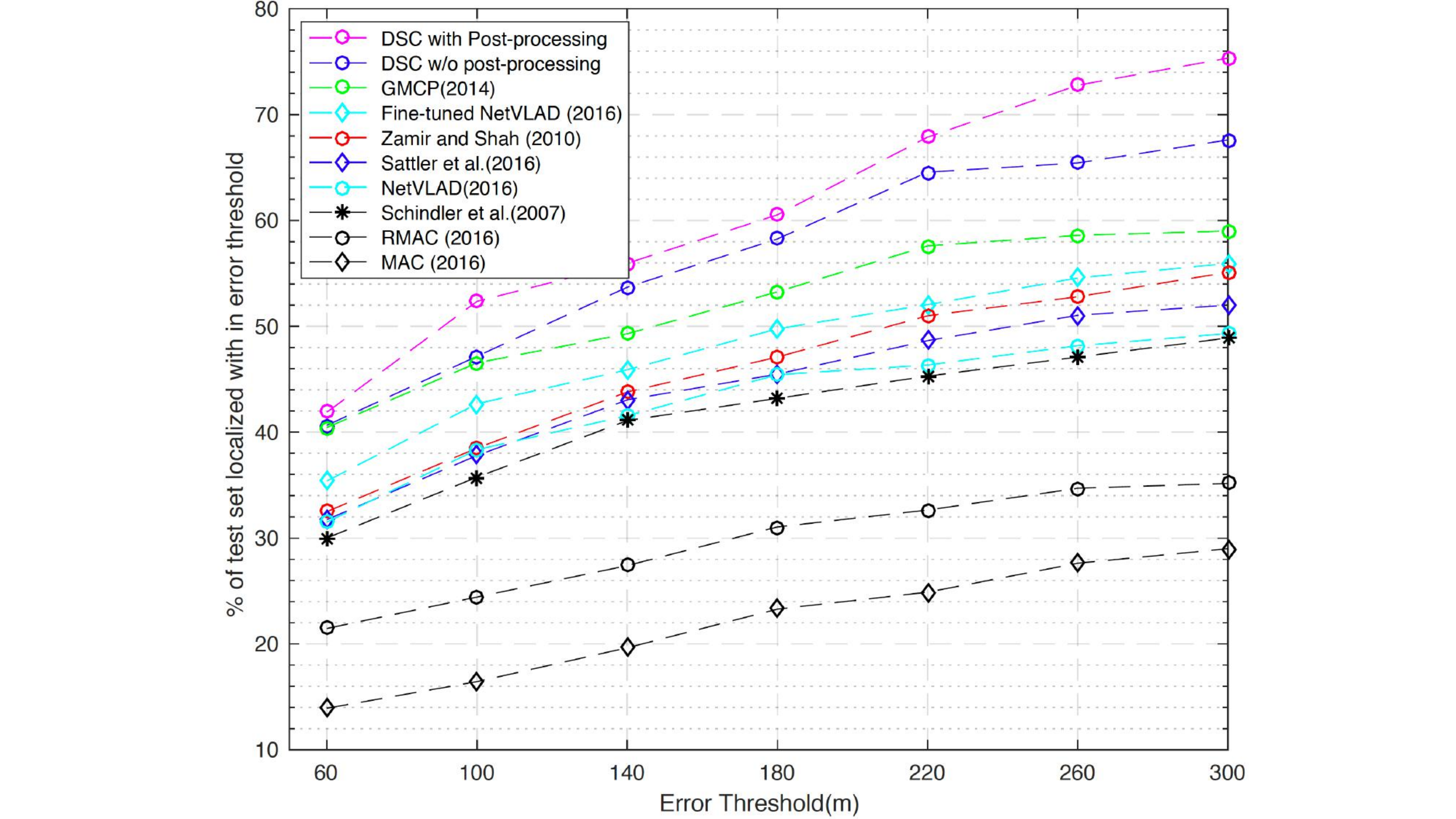}
	\caption{Comparison of our baseline (without post processing) and final method, on overall geo-localization results, with state-of-the-art approaches on the first dataset (102K Google street view images).}
	
	\label{comparison_plot}
\end{figure*}

\textbf{\textit{Computational Time.}} Fig. \ref{plot:cpuTimeRatio}, on the vertical axis, shows the ratio between GMCP (numerator) and our approach (denominator) in terms of CPU time taken for each query images to be localized. As it is evident from the plot, this ratio can range from 200 (our approach 200x faster than GMCP) to a maximum of even 750x faster.


\begin{figure*}[h!]
	\centering
	\includegraphics[width=1\linewidth,trim=3cm 7.5cm 3.75cm 9cm, clip]{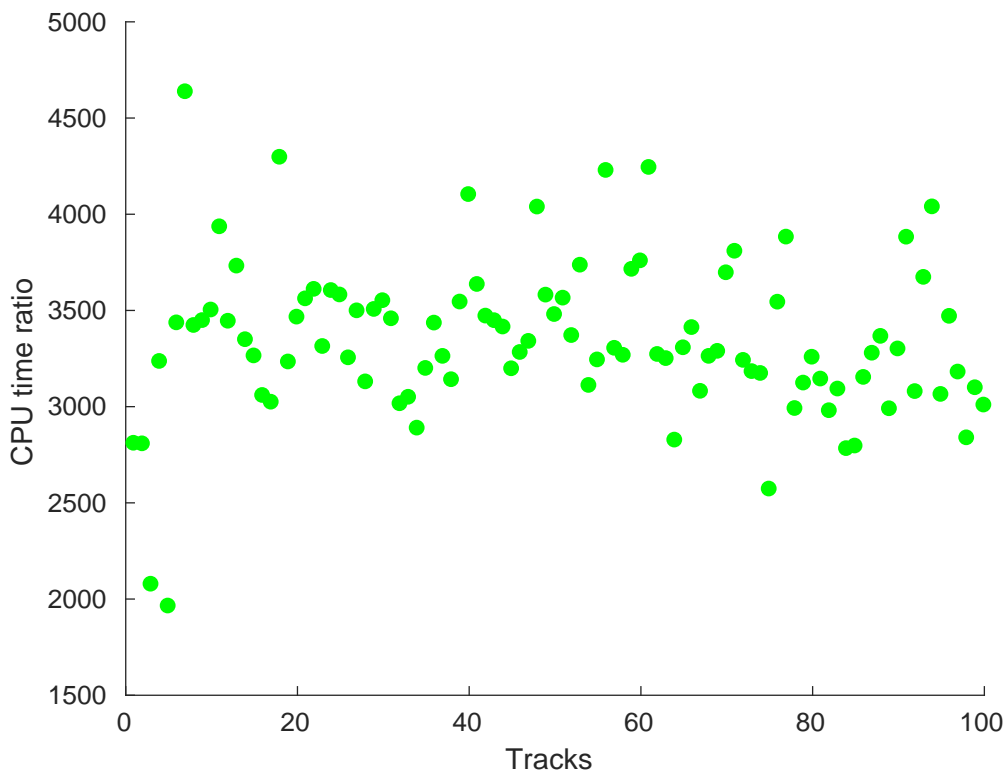}
	\caption{The ratio of CPU time taken between GMCP based geo-localization \cite{amirshahpami2014} and our approach, computed as CPU time for GMCP/CPU time for DSC.}
	\label{plot:cpuTimeRatio}
\end{figure*}

\subsubsection{Performance on the WorldCities Dataset}
We have also compared the performance of different algorithms on the new dataset of 300k Google street view images created by us. Similarly to the previous tests, Fig.~\ref{comparisoin_plot_NewData} reports the percentage of the test set localized within a particular error threshold. Since the new dataset is relatively more challenging, the overall performance achieved by all the methods is lower compared to 102k image dataset.

From bottom to top of the graph in Fig. \ref{comparisoin_plot_NewData} we present the results of \cite{ToliasICLR2016,ToliasECCV2016} black ($-\lozenge-$ and -o-), \cite{SatHavSchPolCVPR2016} {\color{blue} blue ($-\lozenge-$)}, \cite{zamir2010accurate} {\color {red} red (-o-)}, \cite{AraGroTorPajSicCVPR2016}  {\color{cyan} cyan (-o-)}, fine tuned \cite{AraGroTorPajSicCVPR2016} {\color{cyan} cyan ($-\lozenge-$)}, \cite{amirshahpami2014} {\color{green} green (-o-)}, our baseline approach without post processing {\color{blue} blue (-o-)} and our final approach with post processing  {\color{magenta} magenta (-o-)} . The improvements obtained with our method are lower than in the other dataset, but still noticeable (around 2\% for the baseline and 7\% for the final approach).

Some qualitative results for Pittsburgh, PA are presented in Fig. \ref{sample_result_plot_map}.

\begin{figure*}[h!]
	\centering
	\includegraphics[width=1\linewidth,trim=5.00cm 0.0cm 3.55cm 0.0cm, clip]{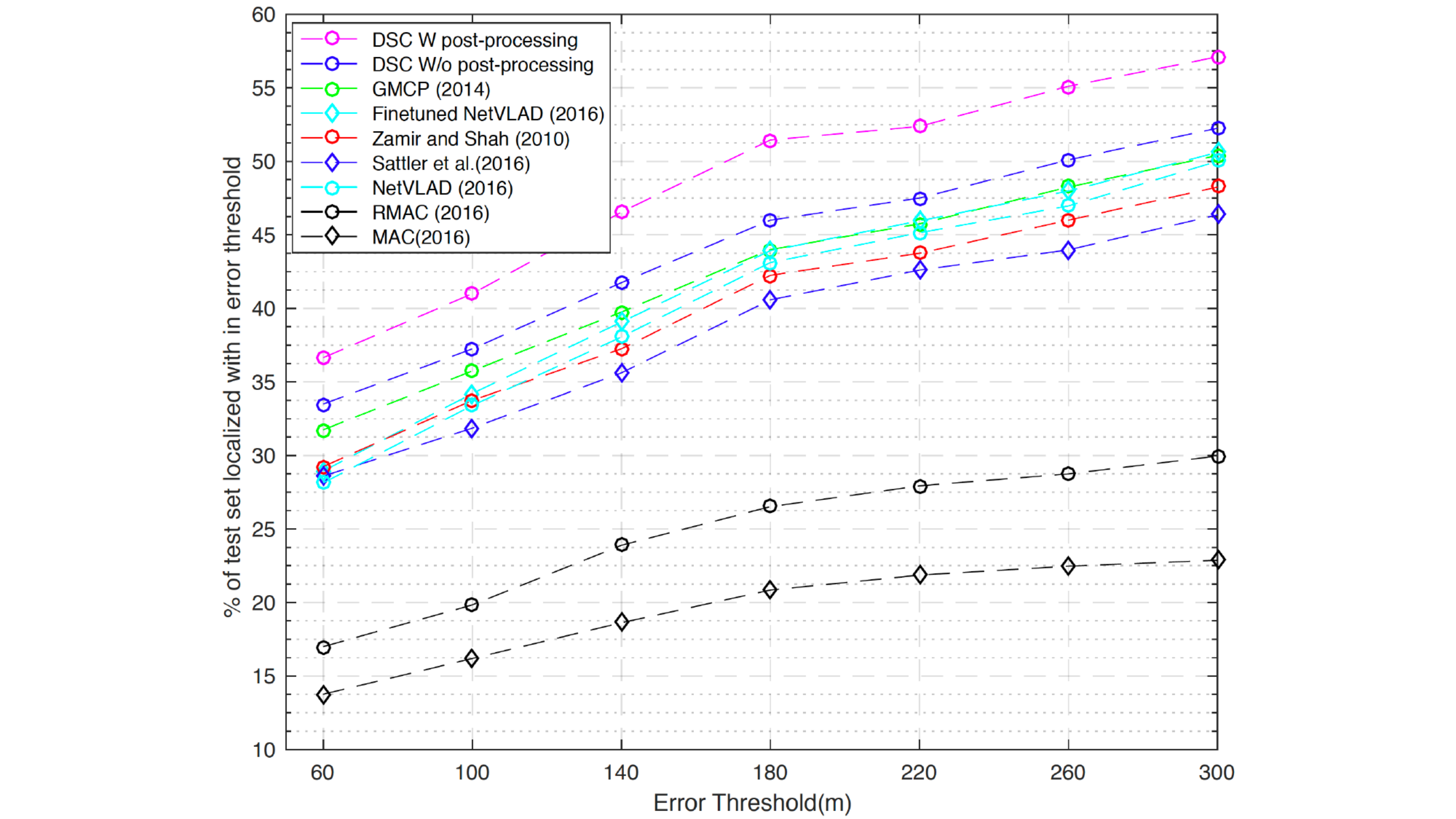}
	\caption{Comparison of overall geo-localization results using DSC with and without post processing  and  state-of-the-art approaches on the \textit{\textbf{WorldCities}} dataset. }
	\label{comparisoin_plot_NewData}
\end{figure*}

\begin{figure*}
	\centering
	\includegraphics[width=1\linewidth,trim=0cm 0.5cm 0cm 0.0cm, clip]{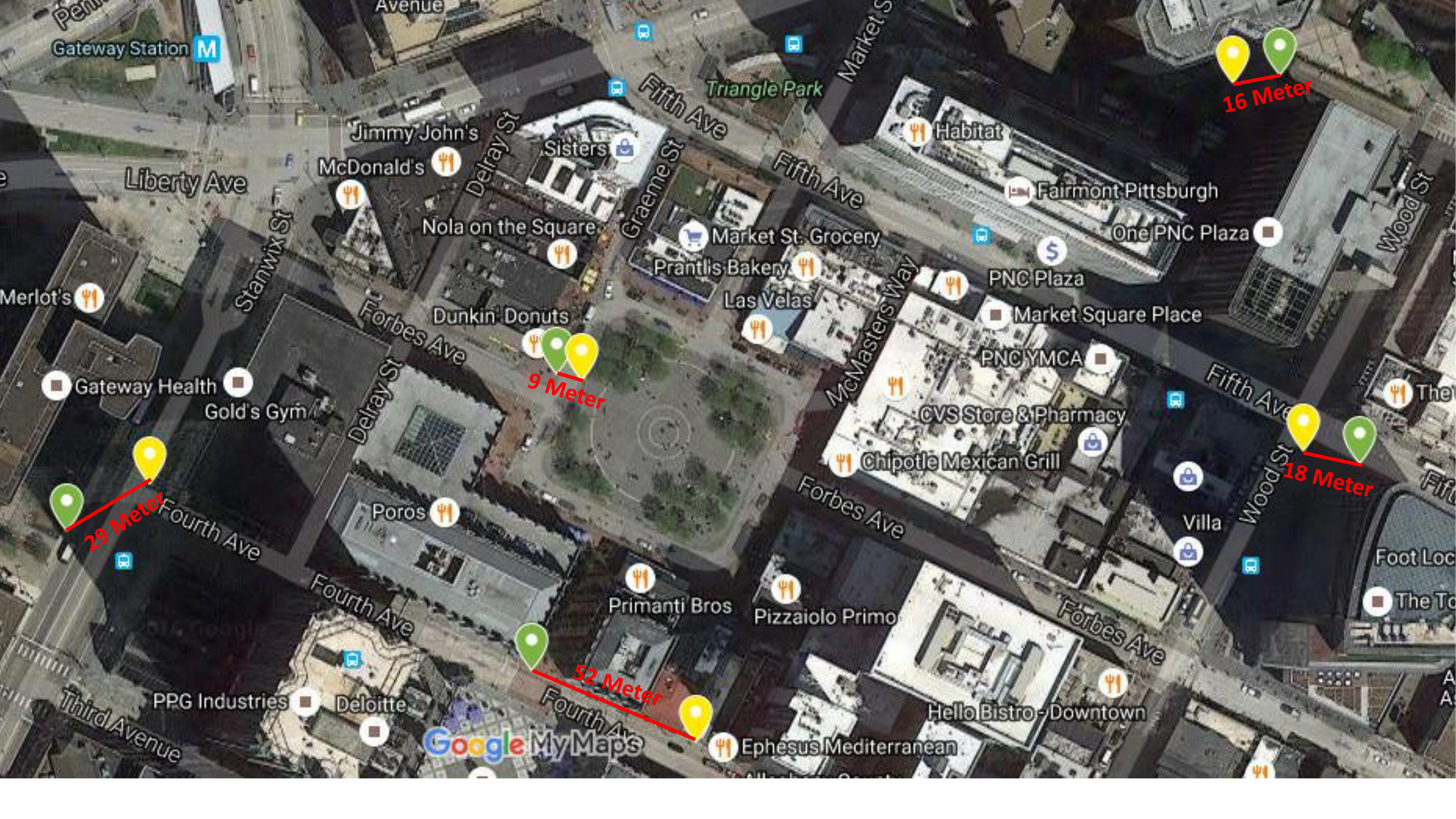}
	\vspace*{-0.5cm}
	\caption{Sample qualitative results taken from Pittsburgh area. The green ones are the ground truth while yellow locations indicate our localization results.}
	\label{sample_result_plot_map}
\end{figure*}
\subsection{Analysis }

\subsubsection{Outlier Handling} In order to show that our dominant set-based feature matching technique is robust in handling outliers, we  conduct  an experiment by fixing the number of NNs (disabling the dynamic selection of NNs) to different numbers. It is obvious that the higher the number of NNs are considered for each query feature, the higher will be the number of outlier NNs in the input graph, besides the increased computational cost and an elevated chance of query features whose NNs do not contain any inliers surviving the pruning stage.

Fig. \ref{plot:outlier_handling} shows the results of geo-localization obtained by using GMCP and dominant set based feature matching on 102K Google street view images \cite{amirshahpami2014}. The graph shows the percentage of the test set localized within the distance of 30 meters as a function of number of NNs. The blue curve shows the results using dominant sets: it is evident that when the number of NNs increases, the performance improves despite the fact that more outliers are introduced in the input graph. This is mainly because our framework takes advantage of the few inliers that are added along with many outliers. The red curve shows the results of GMCP based localization and as the number of NNs increase the results begin to drop. This is mainly due to the fact that their approach imposes hard constraint that at least one matching reference feature should be selected for each query feature whether or not the matching feature is correct.

\begin{figure}[h!]
	\centering
	\includegraphics[width=1\linewidth, trim=2.00cm 1.1cm 3.55cm 0.0cm, clip ]{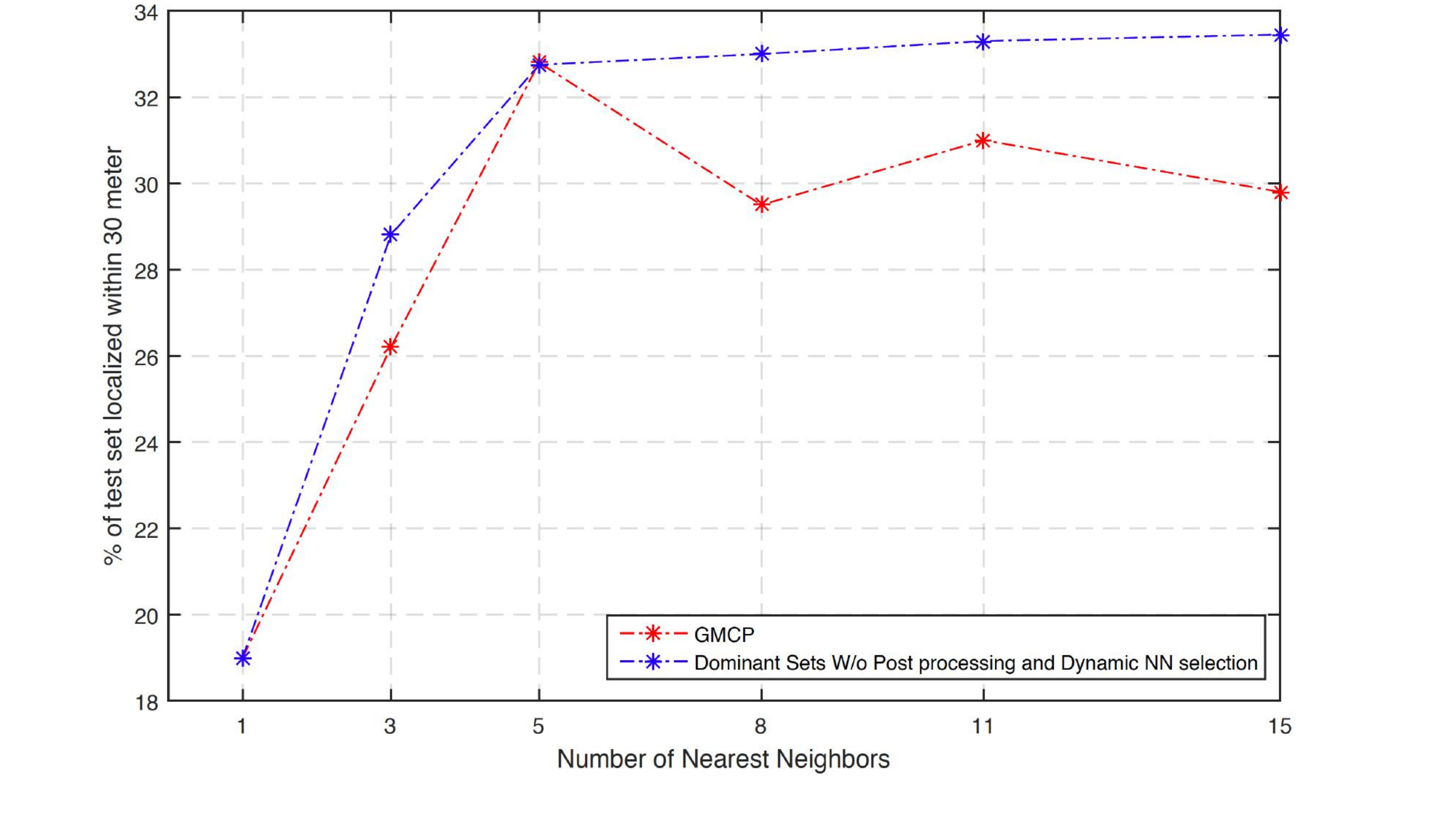}
	\caption{Geo-localization results using different number of NN}
	\label{plot:outlier_handling}
\end{figure}

\subsubsection{Effectiveness of the Proposed Post Processing} In order to show the effectiveness of the post processing step, we perform an experiment comparing our constrained dominant set based post processing with a simple voting scheme to select the best matching reference image. The GPS information of the best matching reference image is used to estimate the location of the query image. In Fig. \ref{plot:post_processing}, the vertical axis shows the percentage of the test set localized within a particular error threshold shown in the horizontal axis (in meters). The blue and magenta curves depict geo-localization results of our approach using a simple voting scheme and constrained dominant sets, respectively. The green curve shows the results from GMCP based geo-localization. As it is clear from the results, our approach with post processing exhibits superior performance compared to both GMCP and our baseline approach.

Since our post-processing algorithm can be easily plugged in to an existing retrieval methods, we perform another experiment to determine how much improvement we can achieve by our post processing. We use \cite{TorSivOkuPajPAMI2015,ToliasICLR2016,ToliasECCV2016} methods to obtain candidate reference images and employ as an edge weight the similarity score generated by the corresponding approaches. Table \ref{table:PostTable} reports, for each dataset, the first row shows rank-1 result obtained from the existing algorithms while the second row (w\_post) shows rank-1 result obtained after adding the proposed post-processing step on top of the retrieved images. For each query, we use the first 20 retrieved reference images. As the results demonstrate, Table \ref{table:PostTable}, we are able to make up to 7\%  and 4\% improvement on  102k Google street view images and \textit{\textbf{WorldCities}} datasets, respectively. We ought to note that, the total additional time required to perform the above post processing, for each approach, is less than 0.003 seconds on average.

%
%
%

%
%
%

\begin{table}[h]
	\centering
	\begin{tabular}{cc|c|c|c|c|c|r}
		
		&	 &{NetVLAD} & {NetVLAD*} &{RMAC}&{MAC} \\ \cline{1-6}
		\cline{1-6}
		\multicolumn{1}{ c  }{\multirow{2}{*}{Dts 1} } &
		\multicolumn{1}{ |c| }{Rank1} & 49.2 & 56.00 &35.16 & 29.00  \\ \cline{2-6}
		\multicolumn{1}{ c  }{}                        &
		\multicolumn{1}{ |c| }{w\_post} & \textbf{51.60} & \textbf{58.05} &\textbf{40.18} & \textbf{36.30}  \\ \cline{1-6}
		\cline{1-6}
		\multicolumn{1}{ c  }{\multirow{2}{*}{Dts 2} } &
		\multicolumn{1}{ |c| }	{Rank1} & 50.00 & 50.61 &29.96 & 22.87  \\ \cline{2-6}
		\multicolumn{1}{ c  }{}                        &
		\multicolumn{1}{ |c| }{ w\_post} & \textbf{53.04} & \textbf{52.23} &\textbf{33.16} & \textbf{26.31}  \\ \cline{1-6}

	\end{tabular}
	
	\caption{Results of the experiment, done on the 102k Google street view images (Dts1) and \textit{\textbf{WorldCities}} (Dts2) datasets, to see the impact of the post-processing step when the candidates of reference images are obtained by other image retrieval algorithms }
	\label{table:PostTable}
\end{table}

The NetVLAD results are obtained from the features generated using the best trained model downloaded from the author's project page \cite{TorSivOkuPajPAMI2015}. It's fine-tuned version (NetVLAD*) is obtained from the model we fine-tuned using images within 24m range as a positive set and images with GPS locations greater than 300m  as a negative set.

The MAC and RMAC results are obtained using MAC and RMAC representations extracted from fine-tuned VGG networks downloaded from the authors webpage \cite{ToliasICLR2016,ToliasECCV2016}.


\begin{figure}[h!]
	\centering
	\includegraphics[width=1\linewidth, trim=5.00cm 0.1cm 3.55cm 0.0cm, clip ]{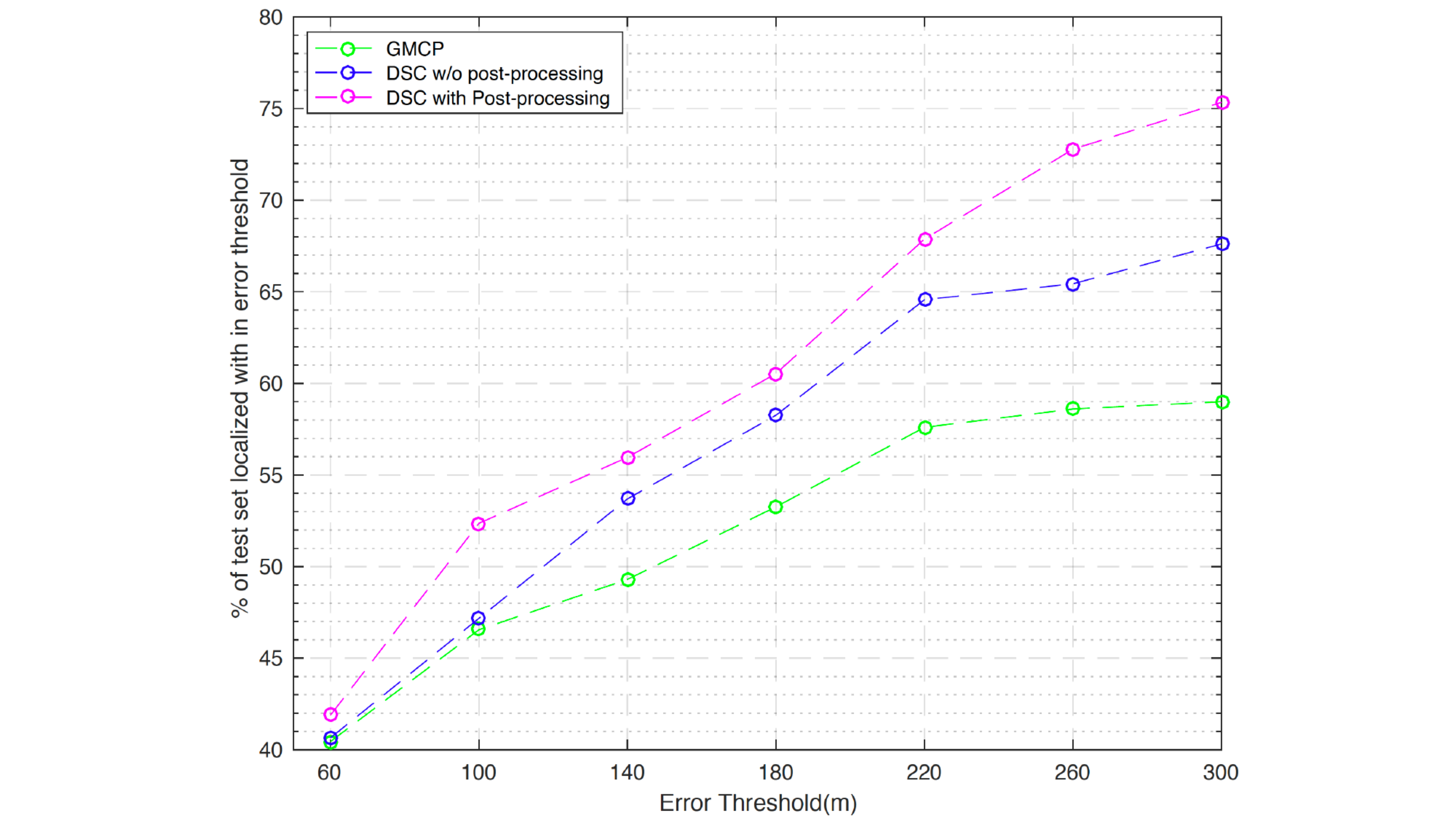}
	\caption{The effectiveness of constrained dominant set based post processing step over simple voting scheme. }
	\label{plot:post_processing}
\end{figure}

\smallskip
\subsubsection{Assessment of Global Features Used in Post Processing Step} 

The input graph for our post processing step utilizes the global similarity between the query and the matched reference images. Wide variety of global features can be used for the proposed technique. In our experiments, the similarity between query and the corresponding matched reference images is computed between their global features, using HSV, GIST, CNN6, CNN7 and fine-tuned NetVLAD. The performance of the proposed post processing technique highly depends on the discriminative ability of the global features used to built the input graph.
\begin{figure}[h]
	\centering
	\includegraphics[width=1.02\linewidth, trim=5.00cm .2cm 4.45cm .1cm, clip ]{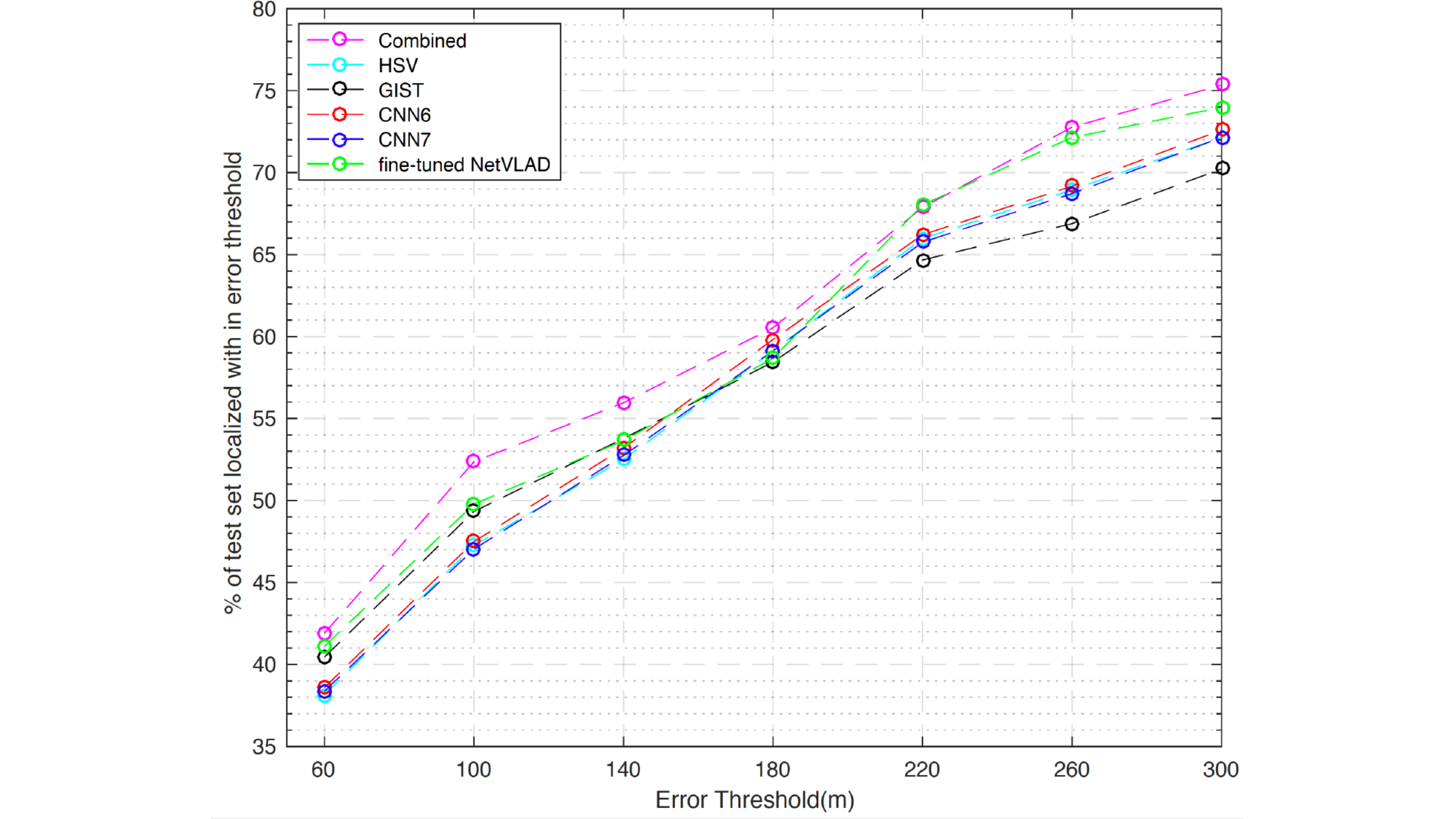}
	\caption{Comparison of geo-localization results using different global features for our post processing step.}
	\label{plot:global_features}
\end{figure}
Depending on how informative the feature is, we dynamically assign a weight for each global feature based on the area under the normalized score between the query and the matched reference images. 
To show the effectiveness of 
this approach, we perform an experiment to find the location of our test set images using both individual and combined global features. Fig. \ref{plot:global_features} shows the results attained by using fine-tuned NetVLAD, CNN7, CNN6, GIST, HSV and by combining them together. The combination of all the global features outperforms the individual feature performance, demonstrating the benefits of fusing the global features based on their discriminative abilities for each query.


\section{Summary}\label{geo_summary}

In this chapter, we proposed a novel framework for city-scale image geo-localization. Specifically, we introduced dominant set clustering-based multiple NN feature matching approach. Both global and local features are used in our matching step in order to improve the matching accuracy. In the experiments, carried out on two large city-scale datasets, we demonstrated the effectiveness of  post processing employing the novel constrained dominant set over a simple voting scheme. Furthermore, we showed that our proposed approach is 200 times, on average, faster than GMCP-based approach \cite{amirshahpami2014}. Finally, the newly-created dataset (\textit{\textbf{WorldCities}}) containing more than 300k Google Street View images used in our experiments is available to the public for research purposes.

As a natural future direction of research, we can extend the results of this work for estimating the geo-spatial trajectory of a video in a city-scale urban environment from a moving camera with unknown intrinsic camera parameters.
 

\chapter{Simultaneous Clustering and Outlier Detection using Dominant sets}
\label{SCOD}
	In this chapter, we present a unified approach for simultaneous clustering and outlier detection in data. We utilize some properties of a family of quadratic optimization problems related to dominant sets, a well-known graph-theoretic notion of a cluster which generalizes the concept of a maximal clique to edge-weighted graphs. Unlike most (all) of the previous techniques, in our framework the number of clusters arises intuitively and outliers are obliterated automatically. The resulting algorithm discovers both parameters from the data. Experiments on real and on large scale synthetic dataset demonstrate the effectiveness of our approach and the utility of carrying out both clustering and outlier detection in a concurrent manner.
	
\section{Introduction} 
\label{introduction_SCOD}
In the literature, clustering and outlier detection are often treated as separate problems. However, it is natural to consider them simultaneously. The problem of outlier detection is deeply studied in both communities of data mining and statistics \cite{ChaBanKumACM2009,Hawbook1980}, with different perspectives.

A classical statistical approach for finding outliers in multivariate data is Minimum Covariance Determinant (MCD). The main objective of this approach is to identify a subset of points which minimizes the determinant of the variance-covariance matrix over all subsets of size $n-l$, where $n$ is the number of multivariate data points and $l$ is the number of outliers. The resulting variance-covariance matrix can be integrated into the Mahalanobis distance and used as part of a chi-square test to identify multivariate outliers \cite{RouDriTechnometrics1998}. However, the high computational complexity makes it impractical for high-dimensional datasets.  

A distance-based outlier detection is introduced by authors in \cite{EdwRayVLDB1998}, which does not depend on any distributional assumptions and can easily be generalized to multidimensional datasets. Intuitively, in this approach data points which are far away from their nearest neighbors are considered as an outlier. However, outliers detected by these approaches are global outliers, that is, the outlierness is with respect to the whole dataset. 

In \cite{MarKriHanRaySanSIGMOD2000}, the authors introduced a new concept that is local outlier factor (LOF), which shows how isolated an object is with respect to its surrounding neighborhood. In this method, they claim that in some situations local outliers are more important than global outliers which are not easily detected by distance-based techniques. The concept of local outliers has subsequently been extended in several directions \cite{ChaBanKumACM2009,ChaSunKnowl2006}. Authors in \cite{CheKeSODA2008} studied a similar problem in the context of facility location and clustering. Given a set of points in a metric space and parameters $k$ and $m$, the goal is to remove $m$ outliers, such that the cost of the optimal $k-median$ clustering of the remaining points is minimized. In \cite{ChaGiosdm2013} authors have proposed $k$-means{-}{-} which generalizes $k-means$ with the aim of simultaneously clustering  data and discovering outliers. However, the algorithm inherits the weaknesses of the classical $k-means$ algorithm: requiring the prior knowledge of cluster numbers $k$; and, sensitivity to initialization of the centroids, which leads to unwanted solutions.

More recently, authors in \cite{OttXiaTozChawaNips2014} modelled clustering and outlier detection as an integer programming optimization task and then proposed a Lagrangian relaxation to design a scalable subgradient-based algorithm. The resulting algorithm discovers the number of clusters from the data however it requires the cost of creating a new cluster and the number of outliers in advance.

In this chapter, we propose a modified dominant set clustering problem for simultaneous clustering and outlier detection from data (SCOD). Unlike most of the above approaches our method requires no prior knowledge on both the number of clusters and outliers, which makes our approach more convenient for real applications.

A naive approach to apply dominant set clustering is to set a threshold, say cluster size, and label clusters with smaller cluster size than the threshold as outliers. However, in the presence of many cluttered noises (outliers) with a uniformly distributed similarity (with very small internal coherency), the dominant set framework extracts the set as one big cluster. That is, cluster size threshold approaches are handicapped in dealing with such cases. Thus what is required is a more robust technique that can gracefully handle outlier clusters of different size and cohesivenesss. 

Dominant set framework naturally provide a principled measure of a cluster's cohesiveness as well as a measure of vertex participation to each group (cluster). On the virtue of this nice feature of the framework, we propose a technique which  simultaneously discover clusters and outlier in a computationally efficient manner. 

The main contributions of this chapter are:
\begin{itemize}
	\item we propose a method which is able to identify number of outliers automatically from the data. 
	\item it requires no prior knowledge of the number of clusters since the approach discovers the number of clusters from the data.	
	\item the effectiveness of the SCOD is tested on both synthetic and real datasets.
\end{itemize}

The rest of the chapter is organised as follows:  Section \ref{Enumerate Dominant Sets Obliterating Outliers} details our approach on enumerating dominant sets while obliterating outliers. Experimental  results are shown in Section , followed by conclusions in Section \ref{Experiments_SCOD}.

\section{Enumerate Dominant Sets Obliterating Outliers}
\label{Enumerate Dominant Sets Obliterating Outliers}
In the dominant set framework, discussed in section \ref{subsect:DominantSetClustering}, a hard partition of the input data is achieved using a \dquote{peel-off} strategy described as follows. 

Initializing the dynamics defined in \eqref{replicatore} to a point near the barycenter of the simplex, say it converges to a point $\x^*$, which is a strict local solution of \eqref{eq2}. Let us determine the dominant set $\mathcal{DS}$ as the support of $\x^*$, $\mathcal{DS}$ = $\sigma(\x^*)$. Then, all the vertices corresponding to the extracted dominant set are removed from the similarity graph. This process is repeated on the remaining graph until all data have been covered, but in applications involving large and noisy data sets this makes little sense. In these cases, a better strategy used in \cite{PavPelPAMI07} is to stop the algorithm using a predefined threshold based on the size of the given data and assign the unprocessed ones to the "nearest" extracted dominant set according to some distance criterion.

This approach has proven to be effective in dealing with situations involving the presence of cluttered backgrounds \cite{EyaMarICIAP2015}. However, it lacks an intuitive way to terminate. In fact, either a manual decision on the number of clusters to be extracted stops the \dquote{peel-off} process or all points will be covered in one of the above two ways. 
It is this limitation which makes the dominant set framework not able to deal with the problem of automated simultaneous clustering and outlier detection.

In this work, we took into account two features which make the dominant set framework able to deal with simultaneous clustering and outlier detection problem, in which the number of clusters arises intuitively while outliers are automatically obliterated: the first one deals with cluster cohesiveness and the second one deals with clusters of different size. 

A nice feature of the dominant set framework is that it naturally provides a principled measure of a
cluster's cohesiveness as well as a measure of a vertex participation to each group. The degree of membership to a cluster is expressed by the components of the characteristic vector $\x^*$ of the strict local solution of \eqref{eq2}: if a component has a small value, then the corresponding node has a small contribution to the cluster, whereas if it has a large value, the node is strongly associated with the cluster. Components corresponding to nodes
not participating in the cluster are zero. A good cluster is one where elements that are strongly associated with it also have large values connecting one another in the similarity matrix. 

The cohesiveness $\mathcal{C}$ of a cluster is measured by the value of equation \eqref{eq2} at its corresponding characteristic vector, $\mathcal{C}$ = $f(\x_c)$: 
\[ \mathcal{C} = f(\x_c) = \x_c' A \x_c\]

A good cluster has high  $\mathcal{C}$ value. The average global cohesiveness $\mathcal{GC}$ of a given similarity matrix can be computed by fixing the vector $\x$ to the barycenter of the simplex, specifically $x_i=1/N$ where $N$ is the size of the data and $i=1 \dots N$.

If the payoff matrix A is symmetric, then $f(\x) = \x' A \x$ is a strictly increasing function along any non-constant trajectory of any payoff-monotonic dynamics of which replicator dynamics are a special instance. This property  together with cohesiveness measure allowed us modify the dominant set framework for SCOD. 

Initializing the dynamics to the barycenter, say $\x_t$ at a time $t=1$, a step at each iteration implies an increase in $f(\x_{t+i})$ for any $i>1$, which again entails that at convergence at time $t=c$, the cohesiveness of the cluster, extracted as the support of $\x_{c}$, is greater than the average global cohesiveness ($\mathcal{GC}$), i.e \[ \mathcal{GC} = \x_1' A \x_1 < \x_c' A \x_c \].

In the dominant set framework, there are situations where false positive clusters arise.

First, large number of very loosely coupled objects with similar affinities may arise as a big cluster. This can be handled using the cohesiveness measure as there will not be any big step of the point that initializes the dynamics.

Secondly, a small compact set of outliers form a cluster whose cohesiveness is greater than the average global cohesiveness of the original similarity. 
In our automated framework, to address these issues, a new affinity ($\mathcal{S}$) is built from the original pairwise similarity (A) based on M-estimation from robust statistics.

To every candidate $i$ a weight which intuitively measures its average relationship with the local neighbors is assigned: 

\begin{equation}
\mathcal{S}(i,j) = w(i)w(j)A(i,j)
\label{eq: LearnAffinity}
\end{equation}

\noindent where $w(i)$ = $\frac{1}{|\mathcal{N}_i|}\sum\limits_{j\in \mathcal{N}_i}A(i,j)$ and $\mathcal{N}_i$ is the set of top $\mathcal{N}$ similar items, based on the pairwise similarity (A), of object $i$. The framework is not sensitive to the setting of the parameter $(\mathcal{N})$. In all the experiments we fixed it as 10\% of the size of the data. One may also choose it based on the average distances among the objects. A similar study, though with a different intention, has been done in \cite{ChaYeuPR2008} and illustrated that this approach makes the system less sensitive to the parameter sigma to built the similarity.

The newly built affinity ($\mathcal{S}$) can be used in different ways: first, we can use it to recheck if the extracted sets are strict local solution of (\ref{eq2}) setting (A) = $(\mathcal{S})$. Another simpler and efficient way is using it for the cohesiveness measure i.e, an extracted set, to be a cluster, should satisfy the cohesiveness criteria in both affinities A and $\mathcal{S}$.

Figure \ref{fig:sampleData} illustrates the second case. The red points in the middle are a compact outlier sets which forms a dominant set whose cohesiveness ($\mathcal{C}$ = 0.952) is greater than the average global cohesiveness ($\mathcal{GC}$ = 0.580). However, its cohesiveness in the newly built affinity ($\mathcal{CL}$ = 0.447) is less than that of the average global cohesiveness. Observe that the two true positive cluster sets (green and blue) have a cohesiveness measures (in both affinities $\mathcal{C}$ and $\mathcal{CL}$) which are greater than the average global cohesiveness. Algorithm \ref{alg:Algorithm1} summarizes the detail.
%
%

\begin{algorithm}[H]
	\caption{Cluster Obliterating Outliers}
	{\bf INPUT:} Affinity $\mat{A}$\\
	
	\begin{algorithmic}[1]
		
		\State \textit{Outliers} $\leftarrow \emptyset$
		\State \textit{Clusters} $\leftarrow \emptyset$ \\	
		$\mathcal{S}$ $\leftarrow$ Build new affinity from $\mat{A}$ using \eqref{eq: LearnAffinity}\\
		Initialize $\x$ to the barycenter and $i$ and $j$ to 1\\
		$\mathcal{GC}$ $\leftarrow$ $\vct x\T \mat{A} \x$
		
		\While{size of $\mat{A}$ $\ne$ 0}
		$\x_c$ $\leftarrow$ Find local solution of \eqref{eq2}
		\If {$\vct x\T_c \mathcal{S} \x_c$ $<$ $\mathcal{GC}$ or $\vct x\T_c \mat{A} \x_c$ $<$ $\mathcal{GC}$} \\
		
		\hspace*{1cm}$\mathcal{O}_j$ $\leftarrow$ $\sigma(x_c)$, find the $j^{th}$ outlier set\\
		\hspace*{1cm}$j$ $\leftarrow$ $j+1$\\
		\hspace*{0.6cm}\textbf{else} \\
		\hspace*{1cm}$\mathcal{DS}_i$ = $\sigma(\x_c)$, find the $i^{th}$ dominant set\\
		\hspace*{1cm}$i$ $\leftarrow$ $i+1$			   
		\EndIf \\
		Remove $\sigma(x_c)$ from the affinity matrices $\mathcal{S}$ and $\mat{A}$ 
		
		\EndWhile \\
		
		\textit{Clusters} = $\bigcup\limits_{k=1}^i\mathcal{DS}_k$\\
		\textit{Outliers} = $\bigcup\limits_{k=1}^j\mathcal{O}_k$
	\end{algorithmic}
	{\bf OUTPUT:} \{\textit{Clusters}, \textit{Outliers}\}
	\label{alg:Algorithm1}	
\end{algorithm}

\begin{figure}[t!]
	\centering
	\includegraphics[width=1.02\linewidth,trim=1.25cm 1.5cm 1.5cm 1.5cm,clip]{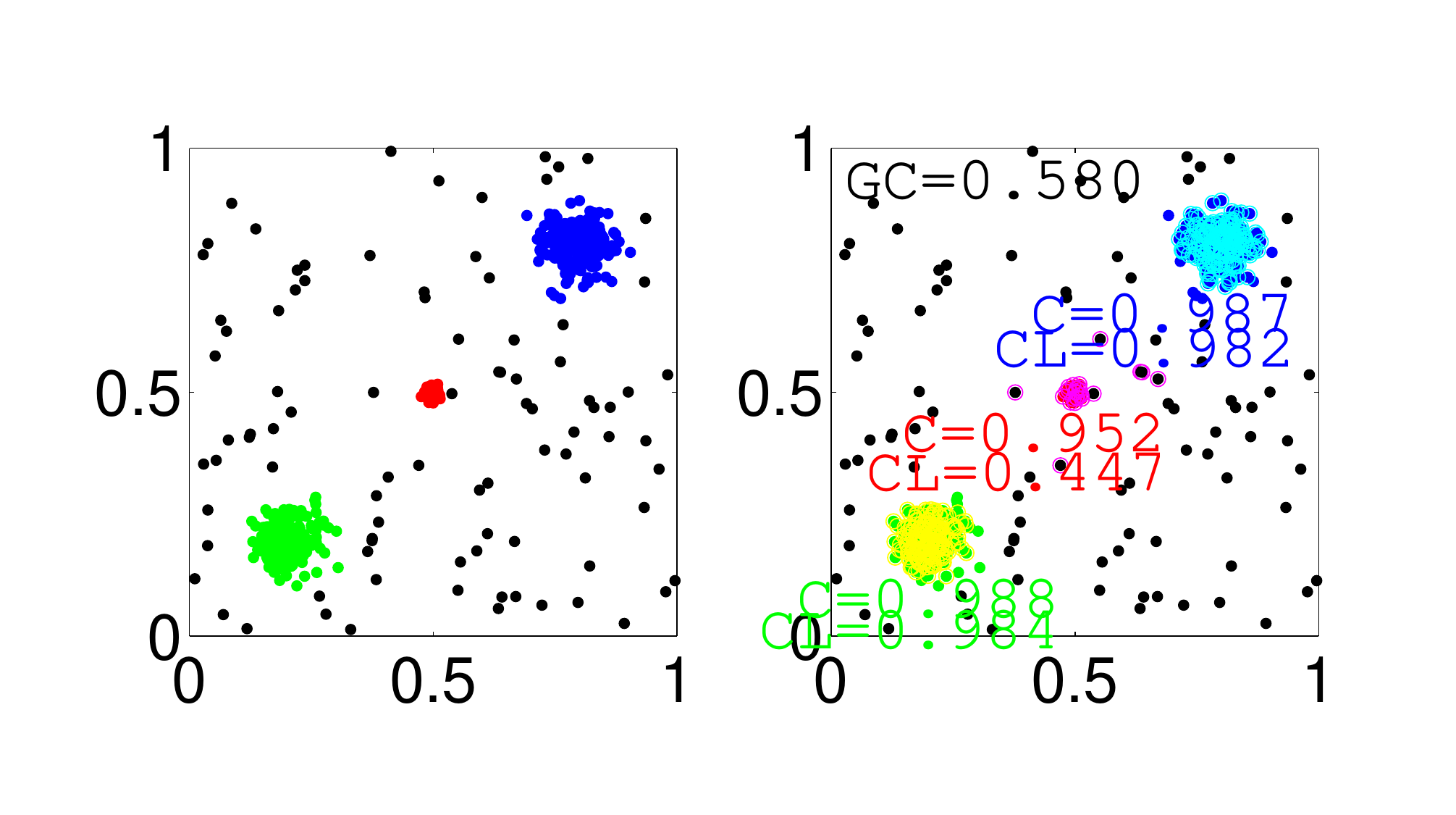}\\
	\caption{\small{ Examplar plots: {\bf Left:} Original data points with different colors which show possible clusters. {\bf Right:} Extracted clusters and their cohesiveness measures, $\mathcal{C}$ with affinity $(\mathcal{A})$ and $(\mathcal{CL})$ with the learned affinity $(\mathcal{S})$}}
	\label{fig:sampleData}
\end{figure}

\section {Experiments}
\label{Experiments_SCOD}
In this section we evaluate the proposed approach on both synthetic and real datasets. First we evaluate our method on a synthetic datasets and present quantitative analysis. Then, we present experimental results on real datasets KDD-cup and SHUTTLE. 

A pairwise distance $\mathcal{D}$ among individuals is transformed to similarity (edge-weight) using a standard Gaussian kernel $$\mat A^\sigma_{ij}=\ind{i\neq j}exp(-\mathcal{D}/{2\sigma^2})$$ where $\sigma$ is choosen as the median distance among the items, and $\ind{P}=1$ if $P$ is true, $0$ otherwise. We compare our Dominant set clustering based approach result with $k$-means{-}{-} \cite{ChaGiosdm2013}.

\subsection{Synthetic data}
The synthetic datasets are used to see the performance of our approach in a controlled environment and evaluate between different methods. Similar to \cite{ChaGiosdm2013}, we generated  synthetic data as follows: $K$ cluster center points are sampled randomly from the space $[0,1 ]^d$ and then $m$ cluster member points are generated for each $k$ clusters by sampling each coordinate from the normal distribution $\mathcal{N}(0,\sigma)$. Finally, $l$ outliers are sampled uniformly at random from the space $[0,1]^d$, where $d$ is the dimensionality of the data.

To assess the performance of our algorithm we use the following  metrics:
\begin{itemize}
	\item The Jaccard coefficient $J$ between the outliers found by our algorithm and the ground truth outliers. It measures how accurately a method selects the ground truth outliers. Computed as :$$J(O,O^{*})=\frac{|O\cap O^{*}|}{|O\cup O^{*}|}$$ where $O$ is the set of outliers returned by the algorithm while $O^*$ are the ground truth outliers. The optimal value is 1.
	\item V-Measure \cite{RosHirEMNLPCoNLL2007} indicates the quality of the overall clustering solution. The outliers are 	considered as an additional class for this measure. Similar to the first measure, also in this case the optimal value is 1.	
	
	\item The \textit{purity} of the results is computed as the fraction of the majority class of each cluster with respect to the size of the cluster. Again, the optimal value is 1.
	
\end{itemize}

We evaluate the performance of the algorithm by varying different parameters of the data-generation process. Our objective is to create increasingly difficult settings so that the outliers eventually become indistinguishable from the points that belong to clusters. The result of our experiments are shown in Figures \ref{fig:Outliers}, \ref{fig:Dimension} and \ref{fig:Sigma} in which we vary the parameters $l$, $d$ and $\sigma$ respectively. For each cases, the rest of the parameters will be kept fixed. In each Figure we show the three measures described above Jaccard Index, V-Measure and Purity. Each box-plots indicate results after running each experiment 30 times. To be fair on the comparison, in each case of the experiment we run $k$-means{-}{-} 10 times (with different initialization) and report the best result, since their algorithm depends on the initialization of the centroids.

As we can see from the Figures, the performance of our algorithm is extremely good. In Figure \ref{fig:Outliers}, were we vary the number of outliers, our approach scored almost 1 in all measurements. This is mainly because, we introduced a robust criteria, that takes in to account the cohesiveness of each extracted clusters, which enables our approach to obliterate outliers in an efficient way. While the results of $k$-means{-}{-} decreases as the number of outliers increases. In Figure \ref{fig:Dimension}, the case were we vary the dimension, our approach scores relatively low in most of the measurements when the dimension is set to 2. But we gate excellent result as the dimension increases. In the last case, in Figure \ref{fig:Sigma}, we can see that our method is invariant to the value of standard deviation and it gates almost close to 1 in most of the measurements.

\begin{figure}[th!]
	\begin{center}
		\includegraphics[width=1\linewidth]{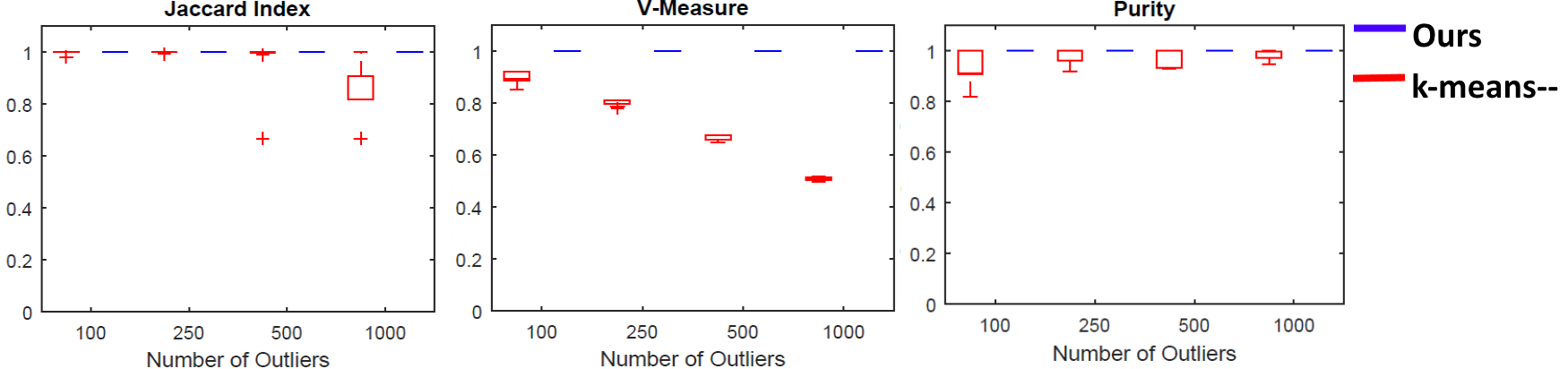}
	\end{center}
	
	\caption{Results of the algorithm on synthetic datasets with respect to increasing number of outliers ($l$). While fixing parameters as $k$ = 10, $m$ = 100, $d$ = 32, and $\sigma$ = 0.2}
	\label{fig:Outliers}
\end{figure}

\begin{figure}[h!]
	\begin{center}
		\includegraphics[width=1\linewidth]{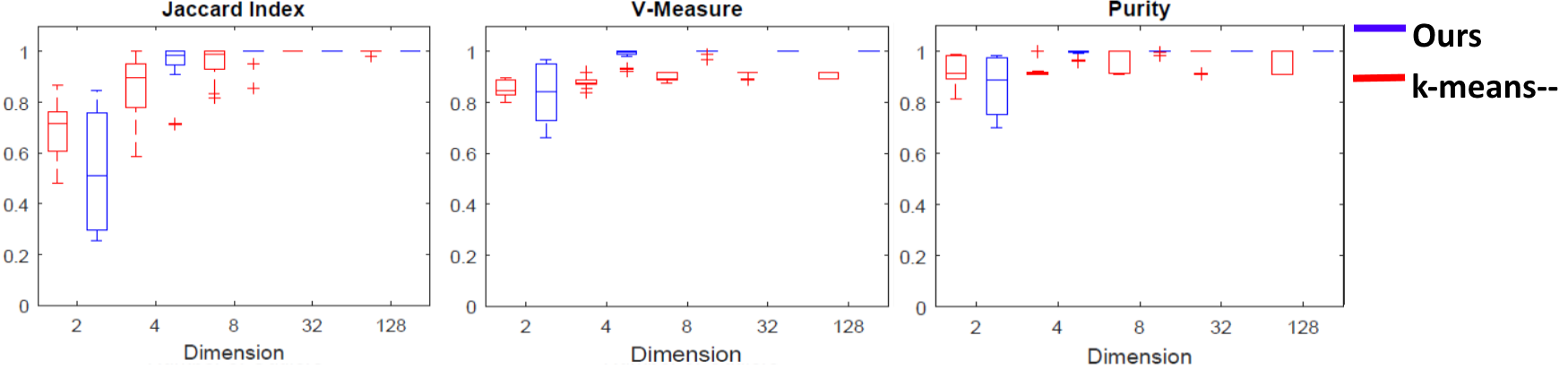}
	\end{center}
	
	\caption{Results of the algorithm on synthetic datasets with respect to increasing dimension ($d$). While fixing parameters as $k$ = 10, $m$ = 100, $l$ = 100, and $\sigma$ = 0.1}
	\label{fig:Dimension}
\end{figure}
\begin{figure}[h!]
	\begin{center}
		\includegraphics[width=1\linewidth]{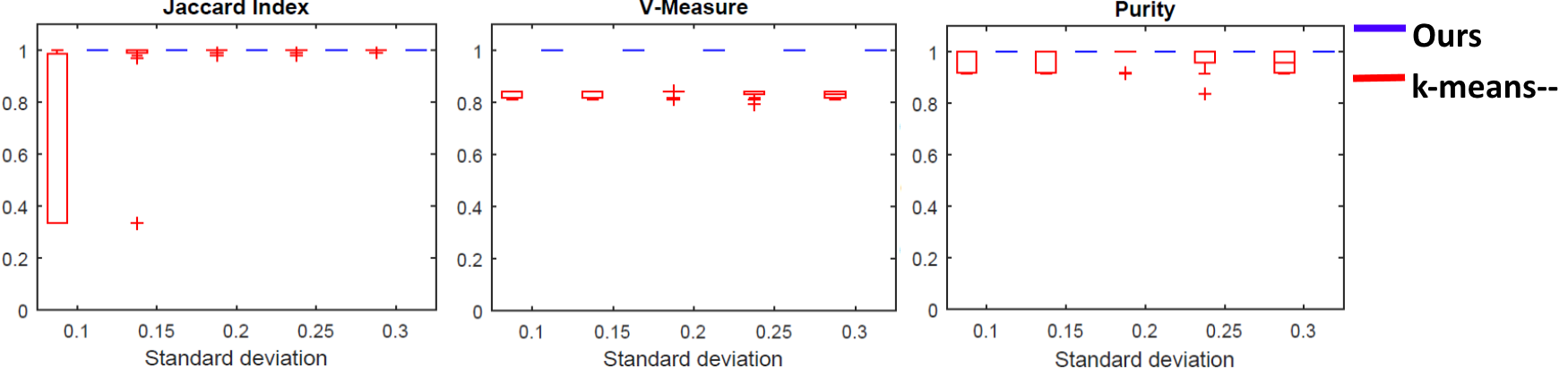}
	\end{center}
	
	\caption{Results of the algorithm on synthetic datasets with respect to the standard deviation used to generate the data ($\sigma$). While fixing parameters as $k$ = 10, $m$ = 100, $l$ = 100, and $d$ = 32}
	\label{fig:Sigma}
\end{figure}

\subsection{Real datasets}
In this section we will discuss the performance of our approach on real datasets. 
\subsubsection{SHUTTLE}
We first consider the \dquote{SHUTTLE} dataset which is publicly available on UCI Machine Learning repository \cite{FraAsuUCI2010}. This dataset contains 7 class labels while the main three classes account $99.6\%$ of the dataset, each with $78.4\%$, $15.5\%$, and $5.6\%$ of frequency. We took these three classes as non-outliers while the rest $(0.4\%)$ are considered as outliers. The dataset contains one categorical attribute, which is taken as class label, and 9 numerical attribute. We use the training part of the dataset, which consists of 43500 instances.

The results of our algorithm on this dataset is shown on Table \ref{table:Tableshuttle1} . The \textit{precision} is computed with respect to the outliers found by the algorithm to the ground truth outliers. Since the number of outliers $l$ and cluster $k$, required by $k$-mean{-}{-}, is typically not known exactly we explore how its misspecification affects their results.

To investigate the influence of number of cluster ($k$) on $k$-means{-}{-}, we run the experiments varying values of $k$ while fixing number of outliers to $l = 0.4\%$ (the correct  value). As it can be seen from Table \ref{table:Tableshuttle1} miss-specification of number of clusters has a negative effect on $k$-means{-}{-}. The approach performs worst in all measurements as the number of clusters decreases. Our approach has the best result in all measurements. We can observed how providing  $k$-means{-}{-} with different numbers of $k$ results in worst performance which highlights the advantage of our method which is capable of automatically selecting the number of clusters and outliers from the data.
\begin{table}[!h]
	\caption{Results on SHUTTLE dataset with fixed $l$ and varying $K$}
	\label{table:Tableshuttle1}
	\begin{center}
		\begin{tabular}{lllllllllllllll} 
			Method   &  $K$  & $l$ & precision & Purity & 	 V-measure\\
			\noalign{\smallskip}
		\cline{1-6}
			\noalign{\smallskip}
			& 10 &   0.4\% &0.155  & 		0.945 	 & 		0.39\\ 
			$k$-means{-}{-}   & 15 &   0.4\% &0.160  & 		0.957 	 & 		0.35\\ 
			& 20 &   0.4\% &0.172  & 		0.974 	 & 		0.33\\ 		
			\cline{1-6}
			
			\textbf{Ours}     & n.a. &   n.a.&\textbf{0.29}  & 	\textbf{0.977} 	 & 		\textbf{0.41}\\
		\cline{1-6}
		\end{tabular}
	\end{center}
\end{table}

We made further investigation on the sensitivity of  $k$-means{-}{-} on the number of outliers ($l$) by varying the values from $0.2\%$ to $0.8\%$, while fixing $k = 20$. As the results on table \ref{table:Tableshuttle2} shows, as the number of outliers increase the precision of $k$-means{-}{-} decreases, means their algorithm suffers as more outliers are asked to be retrieved the more difficult it will become to separate them from the rest of the data. As we can see from Table \ref{table:Tableshuttle2} our approach has stable and prevailing results over $k$-means{-}{-} in all experiments. Our method is prone to such variations in the parameters, from the fact that it is able to automatically identify both the number of cluster and outliers from the data. 

\begin{table}[bth]
	\caption{Results on SHUTTLE dataset with fixed $k$ varying $l$}
	\label{table:Tableshuttle2}
	\begin{center}
		\begin{tabular}{lllllllllllllll} 
			Method   &  $k$  & $l$ & Precision & Purity & 	 V-measure\\ 		
		\cline{1-6}
		\noalign{\smallskip}
			\noalign{\smallskip}
			& 20 &   0.2\% &0.207  & 		0.945 	 & 		0.310\\ 
			$k$-means{-}{-}   & 20 &   0.4\% &0.172  & 		0.957 	 & 		0.305\\ 
			& 20 &   0.8\% &0.137  & 		0.974 	 & 		0.292\\ 		
		\cline{1-6}
			\textbf{Ours}     & n.a. &   n.a.&\textbf{0.29}  & 	\textbf{0.977} 	 & 		\textbf{0.41}\\
		\cline{1-6}
		\end{tabular}
	\end{center}
\end{table}

\subsubsection{KDD-CUP}
We further evaluate our approach on 1999 KDD-CUP dataset which contains instances describing connections of tcp packet sequences. Every row is labeled as \textit{intrusion} or \textit{normal} along with their intrusion types. Since the dataset has both categorical and numerical attributes, for simplicity, we consider only 38 numerical attributes after having normalized each one of them so that they have 0 mean and standard deviation equal to 1. Similar to \cite{ChaGiosdm2013}, we used $10\%$ of the dataset for our experiment, that is, around 494,021 instances. There are 23 classes while $98.3\%$ of the instances belong to only 3 classes, namely the class \textit{smurf} $56.8\%$, the class \textit{neptune} $21.6\%$ and the class \textit{normal} $19.6\%$. We took these three classes as non-outliers while the rest $(1.7\%)$ are considered as outliers.

The result of our algorithm on KDD-CUP dataset is reported in Table \ref{table:kdd}. Here also we compared our result with $k$-means{-}{-} while taking different values of $k$.
We see that both techniques perform quit well in purity, that is, they are able to extract clusters which best matches the ground truth labels. While our algorithm better performs in identifying outliers with relatively good precision.

\begin{table}[!h]
	\caption{Results on KDD-CUP dataset with fixed number of outliers while varying cluster number}
	\label{table:kdd}
	\begin{center}
		\begin{tabular}{lllllllllllllll} 
			Method   &  $k$  & $l$ & Precision & Purity \\
			\cline{1-5}
			\noalign{\smallskip}
			& 5 &   1.7\% &0.564  & 		0.987 	 \\ 
			$k$-means{-}{-}     & 7 &   1.7\% &0.568  & 	\textbf{0.990} \\ 
			& 13&   1.7\% &0.593  & 		0.981	 \\ 		
		\cline{1-5}
			
			\textbf{Ours}        & n.a. &  n.a. &\textbf{0.616}  & 	\textbf{0.990} 	\\
		\cline{1-5}
		\end{tabular}
	\end{center}
\end{table}

\section{Summary}
\label{Summary}
In this chapter, we propose a modified dominant set clustering problem for simultaneous clustering and outlier detection from the data. Unlike most of the previous approaches our method requires no prior knowledge on both the number of clusters and outliers, which makes our approach more convenient for real application. Moreover, our proposed algorithm is simple to implement and highly scalable. We tested the performance of SCOD on both large scale synthetic and real datasets, and showed prevailing result.
 

\chapter{Conlusion}

In this thesis, we proposed several algorithms to solve problems in computer vision and pattern recognition, using the same underlining clustering framework, i.e., dominant set clustering and its extensions. We formalized multi-target tracking, image geo-localization and outlier detection as finding cluster (clique) from the constructed graph. The intrinsic properties of the proposed underlining clustering approach make it suitable to solve the above-mentioned problems. In compassion with several other clustering approaches, it has several advantages like: it doesn't need any a prior knowledge on the number of clusters, it does clustering while obliterating outliers in simultaneous fashion, able to deal with compact clusters and with situations involving arbitrarily-shaped clusters in a context of heavy background noise, does not have any assumptions with the structure of the affinity matrix, and it is fast and scalable to large scale problems.

In Chapter 2, we briefly introduce dominant set clustering framework and its extension, constrained dominant sets. We then present a new fast approach to extract constrained dominant sets from the graph. 

In Chapter 3 and 4, we explore the fundamental problem of multi-target tracking in surveillance videos. Chapter 3 presents a dominant set clustering (DSC) based tracker, which formulates the tracking task as finding dominant set (cluster) on the constructed undirected edge weighted graph. We utilized both appearance and position information for data association in a \textit{global} manner, avoiding the locally-limited approach typically present in previous works. Experimental results compared with the state-of-the-art tracking approaches show the superiority of our tracker. However, since we followed a "peel-off" strategy to enumerate dominant sets from the graph, that is, at each iteration, we remove clusters (dominant sets) from the graph, which causes change in the scale of the problem.

Chapter 4 presents a new tracking approach which is able to overcome the limitations of our DSC tracker, and also extend the problem to multiple cameras with non-overlapping field of view. In this chapter, we presented a constrained dominant set clustering (CDSC) based framework for solving multi-target tracking problem in multiple non-overlapping cameras. The proposed method utilizes a three-layer hierarchical approach, where within-camera tracking is solved using first two layers of our framework resulting in tracks for each person, and later in the third layer the proposed across-camera tracker merges tracks of the same person across different cameras. Experiments on a challenging real-world dataset (MOTchallenge DukeMTMCT) validate the effectiveness of our model. We further perform additional experiments to show effectiveness of the proposed across-camera tracking on one of the largest video-based people re-identification datasets (MARS). Here each query is treated as a constraint set and its corresponding members in the resulting constrained dominant set cluster are considered as possible candidate matches to their corresponding query.

In chapter 5, we proposed a novel framework for city-scale image geo-localization. Specifically, we introduced dominant set clustering-based multiple NN feature matching approach. Both global and local features are used in our matching step in order to improve the matching accuracy. In the experiments, carried out on two large city-scale datasets, we demonstrated the effectiveness of post processing employing the constrained dominant set over a simple voting scheme. 
We evaluate the proposed framework on an existing dataset as well as a new larger dataset, and show that it outperforms the state-of-the-art by 20\% and 7\%, respectively, on the two datasets. Furthermore, we showed that our proposed approach is 200 times, on average, faster than GMCP-based approach \cite{amirshahpami2014}. Moreover, the newly-created dataset (WorldCities) containing more than 300k Google Street View images used in our experiments is available to the public for research purposes.

Finally, in chapter 6, we present a modified dominant set clustering problem for simultaneous clustering and outlier detection from the data. Unlike most of the previous approaches our method requires no prior knowledge on both the number of clusters and outliers, which makes our approach more convenient for real application. Moreover, our proposed algorithm is simple to implement and highly scalable. We first test the performance of SCOD on large scale of synthetic datasets which confirms that in a controlled set up, the algorithm is able to achieve excellent result in an efficient manner. We conduct further evaluation on real datasets and attain a significant improvement.

\cleardoublepage
\phantomsection
\addcontentsline{toc}{chapter}{\bibname}
\small
\bibliographystyle{plain}
\bibliography{thesis}

\end{document}